\pdfoutput=1
\documentclass[11pt]{article}

\usepackage[preprint]{acl}

\usepackage{times}
\usepackage{latexsym}

\usepackage[T1]{fontenc}
\usepackage[utf8]{inputenc}

\usepackage{microtype}

\usepackage{inconsolata}

\usepackage{graphicx}
\usepackage{booktabs}       % professional-quality tables
\usepackage{multicol}
\usepackage{rotating}
\usepackage{changepage}
\usepackage{colortbl}
\usepackage{multirow}
\usepackage{tabularx}
\usepackage{amsmath}
\usepackage{amsfonts}
\usepackage{hyperref}
\newcommand\blfootnote[1]{%
  \begingroup
  \renewcommand\thefootnote{}%
  \footnote{#1}%
  \addtocounter{footnote}{-1}%
  \endgroup
}

\title{Induction Heads as an Essential Mechanism for Pattern Matching in In-context Learning}

\author{Joy Crosbie$^{\star}$\\
  ILLC \\
  University of Amsterdam \\
  \texttt{joy.m.crosbie@gmail.com} \\\And
  Ekaterina Shutova \\
  ILLC \\
  University of Amsterdam \\
  \texttt{e.shutova@uva.nl} \\}

\begin{document}
\maketitle
\blfootnote{$^{\star}$ Corresponding author.}
\blfootnote{Code: \href{https://github.com/JoyC177/Induction_Heads_ICL}{github.com/JoyC177/Induction\_Heads\_ICL}}
\begin{abstract}
Large language models (LLMs) have shown a remarkable ability to learn and perform complex tasks through in-context learning (ICL). However, a comprehensive understanding of its internal mechanisms is still lacking.
This paper explores the role of \textit{induction heads} in a few-shot ICL setting. We analyse two state-of-the-art models, Llama-3-8B and InternLM2-20B on abstract pattern recognition and NLP tasks. 
Our results show that even a minimal ablation of induction heads leads to ICL performance decreases of up to \textasciitilde32\% for abstract pattern recognition tasks, bringing the performance close to random. For NLP tasks, this ablation substantially decreases the model's ability to benefit from examples, bringing few-shot ICL performance close to that of zero-shot prompts. We further use \textit{attention knockout} to disable specific induction patterns, and present fine-grained evidence for the role that the induction mechanism plays in ICL. 
\end{abstract}

\section{Introduction}
Large language models have shown a remarkable ability to learn and perform complex tasks through in-context learning (ICL)
\citep{brown2020language, touvron2023llama2}. In ICL, the model receives a demonstration context and a query question as a prompt for prediction. Unlike supervised learning, ICL utilises the pretrained model's capabilities to recognise and replicate patterns within the demonstration context, thereby enabling accurate predictions for the query without the use of gradient updates.

Given the success and wide applicability of ICL, much recent research was dedicated to better understanding its properties. 
Several works explored its connections to gradient descent \citep{dai-etal-2023-gpt, von2023transformers}, suggesting that ICL functions as an implicit form of fine-tuning at inference time. Other works investigated factors influencing ICL, showing that it is driven by the distributions of the training data \citep{chan2022data} and scales with model size, revealing new abilities at certain parameter thresholds \citep{brown2020language, wei2022emergent}. During inference, the properties of demonstration samples also affect ICL performance , with aspects such as the label space, input distribution and input-label pairing playing a crucial role \citep{min-etal-2022-rethinking, webson-pavlick-2022-prompt}. While this work identified interesting properties of ICL and effective ICL prompting strategies, a comprehensive understanding of its operational mechanisms within the models is still lacking.  

Our paper aims to fill this gap, by directly investigating the internal model computations that enable ICL. Our work is inspired by recent research in the field of mechanistic interpretability, which aims to reverse engineer the "algorithm" by which Transformer models process information \citep{geva-etal-2023-dissecting, olah2020zoom, wang2022interpretability}.
As a significant milestone in this area, \citet{elhage2021mathematical} demonstrated the existence of \textit{induction heads} in Transformer LMs. These heads scan the context for previous instances of the current token using a \textit{prefix matching} mechanism, which identifies if and where a token has appeared before. If a matching token is found, the head employs a \textit{copying} mechanism to increase the probability of the subsequent token, facilitating exact or approximate repetition of sequences and embodying the algorithm "[A][B]...[A] → [B]". 

Building on this foundation, \citet{olsson2022context} hypothesised that induction heads are capable of abstract pattern matching and conducted a qualitative analysis of attention patterns observed in an example from an abstract classification task. They observed that these heads, when focused on the final token, tended to attend to previous instances of the correct label. However, they did not conduct any experiments to verify this.
In a related study, \citet{bansal-etal-2023-rethinking} investigated which heads were important for NLP tasks and found some degree of overlap between these heads and those that had high scores associated with induction. However, they did not directly assess the contribution of induction heads to ICL performance of the models in these tasks.

To the best of our knowledge, our paper is the first one to directly, empirically link the specific computations performed by induction heads to measurable improvements in ICL performance. 
We investigate the role of induction heads as an underlying mechanism of few-shot ICL across a range of tasks: (1) abstract pattern recognition tasks; (2) popular NLP tasks. We focus on two state-of-the-art models: Llama-3-8B \citep{dubey2024llama} and InternLM2-20B \citep{cai2024internlm2}. We first compute prefix matching scores for all of their attention heads, and then perform head ablation experiments by removing 1\% and 3\% of the heads with the highest scores. For the NLP tasks, we additionally conduct experiments with semantically unrelated labels to encourage the models to rely on ICL for task completion. To confirm our hypothesis regarding the specific role of the induction mechanisms in these heads, we further perform \textit{attention knockout} experiments, where we selectively inhibit each token's ability to attend to any tokens that directly followed tokens similar to the current token. This effectively simulates a loss of function in the prefix matching and copying mechanisms.

We find that the ablation of induction heads results in a substantial decrease in ICL performance, much more so than when an equivalent percentage of random heads are ablated. Additionally, when we block the induction attention pattern in these heads, performance drops to levels comparable to or worse than those seen with full head ablations. The latter confirms not merely the importance of induction heads for ICL, but also the specific role of the prefix matching and copying mechanism, retracing its application at the token level. 
Our work is thus the first to provide comprehensive experimental evidence, demonstrating (1) the essential role of induction heads in few-shot ICL in large-scale, widely-used LMs and real-world tasks and (2) that induction heads utilise a "fuzzy" version of the prefix matching and copying mechanisms to enable pattern matching.
\begin{figure*}[t!]
\centering
  \includegraphics[width=\textwidth]{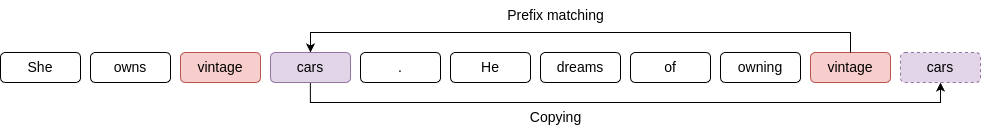}
  \caption{In the sequence ``...vintage cars ... vintage'', an induction head identifies the initial occurrence of ``vintage'', attends to the subsequent word ``cars'' for prefix matching, and predicts ``cars'' as the next word through the copying mechanism.}
\label{fig:pfx_copy}
\end{figure*}

\section{Related Work}
\subsection{Factors that influence ICL performance}
ICL performance is influenced by various factors, both during the pretraining and inference stage. During pretraining, ICL abilities emerge when training data displays specific distributional properties, such as items frequently appearing in clusters and the presence of many infrequently occurring classes \citep{chan2022data}. Additionally, the scalability of these abilities correlates directly with model size \citep{brown2020language}, and emergent capabilities become apparent at particular scales of pretraining or beyond certain parameter thresholds \citep{wei2022emergent, lu2023emergent}. In the inference stage, the characteristics of demonstration examples, such as label space, input distribution, and input-label pairing format, significantly impact ICL performance \citep{min-etal-2022-rethinking}. Additionally, models are more affected by label choice than by instruction semantics \citep{webson-pavlick-2022-prompt}, and larger models can adapt to input-label mappings even with flipped or unrelated labels \citep{wei2023larger}. Further, fine-tuning on semantically unrelated in-context input-label pairs enhances performance on new ICL tasks \citep{wei-etal-2023-symbol}.

\subsection{Mechanistic Interpretability}
Research in mechanistic interpretability seeks to explain model behaviours in terms of their internal components \cite{olah2020zoom}.
\citet{elhage2021mathematical} demonstrated that small, attention-only Transformers can be deconstructed into \textit{circuits}—specific sub-graphs within the model's computational graph that perform distinct tasks \cite{wang2022interpretability}. By analysing these circuits, \citeauthor{elhage2021mathematical} demonstrated the existence of induction heads. Circuit discovery has since led to the localisation of other behaviours within Transformer models \citep{meng2022locating,NEURIPS2020_92650b2e, wang2022interpretability}.
A common technique is \textit{mean ablation}, where components' activations are replaced with their average activation value across a reference distribution to determine whether disrupting this specific component hinders the model's ability to perform the task. \citet{wang2022interpretability} studied indirect object identification in GPT-2 small \citep{radford2019language}, by obscuring the indirect object with random names, averaging these activations, and replacing the target activations during inference. This identified specific attention heads crucial for the task.

 Another recent technique---\textit{attention knockout}---systematically disables specific attention weights to assess their impact on model outputs \cite{geva-etal-2023-dissecting}. \citet{geva-etal-2023-dissecting} explored how factual knowledge is extracted from LLMs during inference, by blocking the final position from attending to both subject and non-subject positions across specific layers. They revealed that essential information from non-subject positions is accessed first.
 Additionally, \citet{wang-etal-2023-label} blocked label tokens from accessing prior demonstration text in shallow layers, and demonstrated that label words gather information early in forward propagation.

\subsection{Induction heads \& ICL}
Building on the work of \citet{elhage2021mathematical}, \citet{olsson2022context} provided initial evidence that induction heads might be an underlying mechanism of ICL, defined as the gradual reduction of loss with increasing token indices. They observed that induction heads emerge early in training, coinciding with significant improvements in the model’s ability to learn from context. They also found that modifying or removing these heads impairs ICL performance, assessed by a heuristic measure of loss reduction. In contrast, we employ a few-shot prompting ICL setting and evaluate ICL performance directly through the model's accuracy on specific tasks.

Using head importance pruning, \citet{bansal-etal-2023-rethinking} demonstrated that up to 70\% of attention heads and 20\% of feed forward layers can be removed from the OPT-66B model \citep{zhang2022opt} with minimal performance decline across various downstream tasks. They uncovered a notable consistency in the relevance of certain attention heads for ICL, irrespective of the task or the number of few-shot examples. They investigated whether these crucial heads were induction heads, finding some degree of overlap with those that had high prefix matching scores. Our research extends these findings by not only identifying induction heads but also conducting ablation studies to directly evaluate their impact on ICL performance in diverse tasks.

\section{Background}
Following the framework established by \citet{elhage2021mathematical} and further discussed by \citet{ren2024identifying}, the operations within a multi-head attention (MHA) layer, comprising $H$ attention heads, can be reformulated as follows:
\begin{equation}
\begin{aligned}
& \sum_{h=1}^{H} \text{softmax}\left(\frac{\mathbf{x}W^{h}_{q}(\mathbf{x}W^{h}_{k})^T}{\sqrt{d_{h}}} + M^h\right)\mathbf{x}W^{h}_{v}W^{h}_{o}\\
= & \sum_{h=1}^{H} \text{softmax}\left(\frac{\mathbf{x}W^{h}_{QK}\mathbf{x}^{T}}{\sqrt{d_{h}}} + M^h\right)\mathbf{x}W^{h}_{OV}
\label{eq:attention}
\end{aligned}
\end{equation}
where $\mathbf{x} = [\mathbf{x}_1^T, \mathbf{x}_2^T, ..., \mathbf{x}_N^T]^T \in \mathbb{R}^{N \times d}$ represents the sequence of embeddings, with each $\mathbf{x}_i = \mathbf{t}_i\mathbf{W}_e \in \mathbb{R}^{1 \times d}$ denoting the embedding of the $i$-th input token $t_i$. $\mathbf{W}_e \in \mathbb{R}^{|\mathcal{V}| \times d}$ denotes the embedding matrix over the vocabulary $\mathcal{V}$. $\mathbf{W}^{h}_q, \mathbf{W}^{h}_k, \mathbf{W}^{h}_v \in \mathbb{R}^{d \times d_h}$ denote the query, key, and value parameter matrices of the $h$-th attention head. $\mathbf{W}_o \in \mathbb{R}^{d \times d}$ represents the output transformation of the MHA layer, which can be deconstructed as $\mathbf{W}_o = [(\mathbf{W}_o^{1})^T (\mathbf{W}_o^{2})^T ... (\mathbf{W}_o^{H})^T]$, where $\mathbf{W}_o^h \in \mathbb{R}^{d_h \times d}$. $M^h$ is a casual attention mask, which ensures each position can only attend to preceding positions (i.e. $M_{rc}^h = -\infty \space\,\forall c > r$ and zero otherwise).

In Equation (\ref{eq:attention}), $\mathbf{W}^{h}_{QK} = \mathbf{W}^{h}_{q} (\mathbf{W}^{h}_{k})^T$, termed the Query-Key (QK) circuit, calculates the attention pattern of the $h$-th head. The matrix $\mathbf{W}^{h}_{OV} = \mathbf{W}^{h}_{v} (\mathbf{W}^{h}_{o})^T$, termed the Output-Value (OV) circuit, determines each head's independent output for the current token. Leveraging this decomposition, \citet{elhage2021mathematical} discovered a distinct behaviour in certain attention heads, which they named \textit{induction heads}. This behaviour emerges when these heads process sequences of the form "[A] [B] ... [A] → ".  In these heads, the QK circuit directs attention towards [B], which appears directly after the previous occurrence of the current token [A]. This behaviour is termed \textit{prefix matching}. The OV circuit subsequently increases the output logit of the [B] token, termed \textit{copying}. An overview of this mechanism is shown in Figure \ref{fig:pfx_copy}. The authors showed that these heads can operate on different distributions as long as the abstract property of repeated sequences' likelihood holds true.

\section{Methods}
\subsection{Models}
We utilise two recently developed open-source models, namely Llama-3-8B \citep{dubey2024llama} and InternLM2-20B \citep{cai2024internlm2}, both of which are based on the original Llama \citep{touvron2023llama1} architecture. Llama-3-8B, comprises 32 layers, each with 32 attention heads. It has shown superior performance compared to its predecessors, even the larger Llama-2 models.

InternLM2-20B features 48 layers with 48 attention heads each. We selected InternLM2-20B for its exemplary performance on the Needle-in-the-Haystack\footnote{\href{https://github.com/gkamradt/LLMTest_NeedleInAHaystack}{github.com/gkamradt/LLMTest\_NeedleInAHaystack}} 
task, which assesses LLMs' ability to retrieve a single critical piece of information embedded within a lengthy text. This mirrors the functionality of induction heads, which scan the context for prior occurrences of a token to extract relevant subsequent information.

\subsection{Identifying Induction Heads}
To identify induction heads within models, we measure the ability of all attention heads to perform prefix matching on random input sequences.\footnote{In this work, the term "induction heads" refers to behavioural induction heads. A true induction head must be verified mechanistically; however, our analysis employs prefix-matching scores as a proxy. We will continue to use the term "induction heads" for simplicity.} We follow the task-agnostic approach to computing prefix matching scores outlined by \citet{bansal-etal-2023-rethinking}.
We argue that focusing solely on prefix matching scores is sufficient for our analysis, as high prefix matching cores specifically indicate induction heads, while less relevant heads tend to show high copying capabilities \citep{bansal-etal-2023-rethinking}. We generate a sequence of 50 random tokens, excluding the 4\% most common and least common tokens. This sequence is repeated four times to form the input to the model. The prefix matching score is calculated by averaging the attention values from each token to the tokens that directly followed the same token in earlier repeats. The final prefix matching scores are averaged over five random sequences.

The prefix matching scores for Llama-3-8B are shown in Figure \ref{fig:pfx_llama}. For IntermLM2-20B, we refer to Figure \ref{fig:pfx_internlm} in Appendix \ref{sec:pfx_intern}. Both models exhibit heads with notably high prefix matching scores, distributed across various layers. In the Llama-3-8B model, \textasciitilde 3\% of the heads have a prefix matching score of 0.3 or higher, indicating a degree of specialisation in prefix matching, and some heads have high scores of up to 0.98.

\begin{figure}[t!]
  \centering
    \includegraphics[scale=0.17]{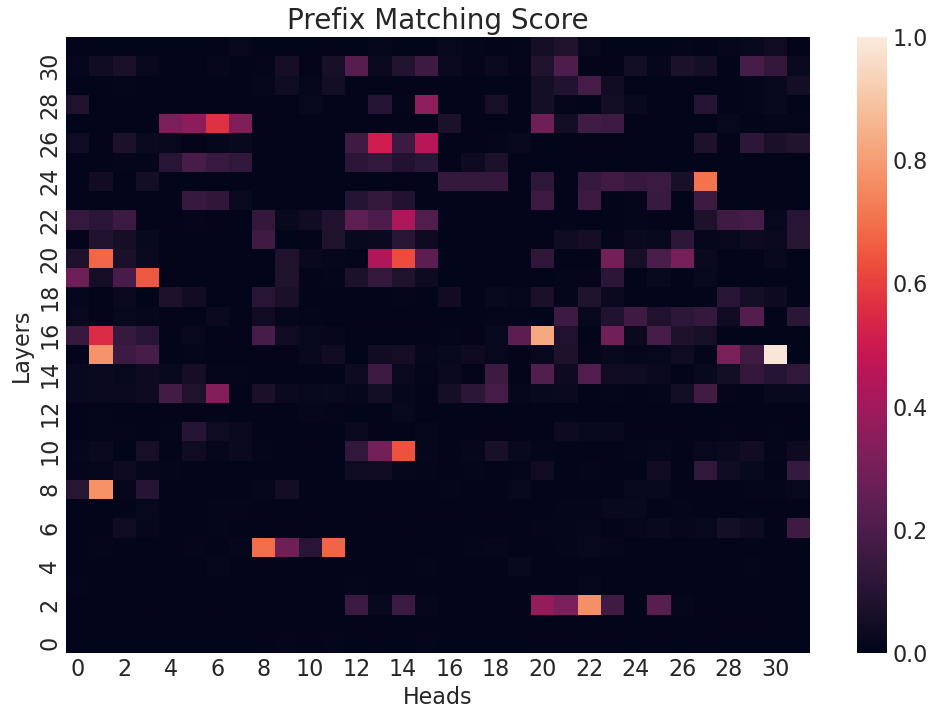}
    \caption{Prefix matching scores for Llama-3-8B.}
    \label{fig:pfx_llama}
\end{figure}

\subsection{Head Ablations}
To investigate the significance of induction heads for a specific ICL task, we conduct mean-ablations of 1\% and 3\% of the heads with the highest prefix matching scores, aiming to introduce noise and disrupt their function. We utilise premises from the MNLI \citep{williams-etal-2018-broad} training set, which encompasses data from ten diverse genres of written and spoken English, thereby capturing a wide spectrum of language complexity. For each model, premises are concatenated to create 500 samples, each comprising of 3000 tokens, exceeding the longest prompt used in our experiments. Each sample is passed through the model, during which head activations are recorded with the Pyvene \citep{wu-etal-2024-pyvene} library. These activations are then averaged to generate a mean embedding vector for each head. During inference, the activations of the ablated heads are replaced with the mean activations by truncating them to the length of the current prompt.

As a control condition, we also randomly ablate 1\% and 3\% of the model's heads, which are selected following the layer distribution of the previously ablated induction heads. For example, if three induction heads were ablated in layer 24, we randomly select three heads from the same layer for ablation. This method allows us to control for layer-specific effects, ensuring that any observed differences in performance are attributable to the function of the induction heads themselves, rather than differences in the computational roles or importance of the various layers.

\subsection{Attention Knockout}
To better understand the function of induction heads in few-shot ICL, we conduct \textit{attention knockout} experiments.
We hypothesise that when induction heads process a structured dataset, they generate a specific "induction pattern" by consistently attending back to tokens that previously followed similar ones. Drawing on the methodology of \citet{geva-etal-2023-dissecting}, we empirically test whether induction heads exhibit this pattern by blocking tokens from attending to tokens that previously followed similar ones. We define two positions $r,c \in [1, N]$ where $r < c$. To inhibit attention, we prevent $x_c^{h}$ from attending to $x_r^{h}$ in head $h$ by modifying the attention weights in that layer (Eq. \ref{eq:attention}) as follows: $M_{rc}^{h}=-\infty$. 
This restricts the current position from obtaining information from the blocked positions for that particular head.
By comparing the effects of completely removing a head with merely blocking its pattern-directed attention, we can gather empirical evidence suggesting whether these heads predominantly implement this specific induction pattern. 

\section{Ablation Experiments: Abstract Pattern Recognition Tasks}

We use a modified version of Eleuther AI’s Language Model Evaluation Harness \citep{eval-harness} for our experiments. While the default framework randomly samples different in-context examples for each query, we adapted it to ensure a balanced sampling approach, in terms of both the number of examples and classes in the dataset. Aside from this modification, all default settings were maintained. 

For predictions, we utilise the harness’s \textit{multiple-choice} method, which identifies the target word assigned the largest log probability among all target words. We report accuracy averaged over three random seeds for example selection. For random head ablation, we report accuracy over three ablation seeds, totalling nine runs.

\paragraph{Datasets and experimental setup}
We first conduct ablation experiments on abstract pattern recognition tasks. In these tasks, we expect the model to rely on ICL more so than leveraging information from its training data or weights. The first set of tasks focus on recognising predefined patterns in sequences of letters. In the \textsc{repetition} task, the model must determine whether a sequence of four letters consists of repeating two tokens (e.g.  A B A B). The \textsc{recursion} task involves detecting whether four-letter-sequences contain recursion (e.g. A B B B). In the \textsc{centre-embedding} task, whether a sequence of five letters contains a centre-embedding (e.g. G L O L G). If the specific pattern is not present in any task, the sequence is a random sequence of letters.

 In the second set of tasks, the model is required to label sequences of words that contain a pattern rarely seen in natural language data (i.e. not a common semantic relation). We pair specific semantic categories of words---such as <fruit, animal>, <furniture, profession> (\textsc{WordSeq 1}) and <vegetable, vehicle>, <body part, instrument> (\textsc{WordSeq 2})---and conduct a binary or four-way classification of these pairs. Figure~\ref{fig:data_examples} presents examples from the letter-sequence tasks and one of the word-sequence tasks.

\begin{figure}[t!]
\centering
\small
\resizebox{0.5\textwidth}{!}{%
\renewcommand{\arraystretch}{1.25}
\begin{tabular}{|lll|} 
\hline
\textbf{Task} & \textbf{Pattern (\textbf{Foo})} & \textbf{No pattern (\textbf{Bar})} \\
\hline
\multirow{2}{*}{Repetition} & X H X H & Z E W F \\
                            & Q A Q A & F I O E \\
\hline
\multirow{2}{*}{Recursion}  & V D D D & N O W T \\
                            & F L L L & P Z X F \\
\hline
\multirow{2}{*}{Centre-embedding} & X B V B X & L Q I F P \\
                                  & V G M G V & H M T B A \\
\hline
\multirow{2}{*}{WordSeq 1 (binary)} & grape shark & couch soldier \\
                                                   & fig panda & nightstand writer \\
\hline
\end{tabular}
}
\caption{Each letter-sequence dataset features examples following the respective patterns labelled "Foo" and random sequences labelled "Bar". For word sequence tasks, examples include pairs of semantically categorised words.}
\label{fig:data_examples}
\end{figure}
\setlength{\tabcolsep}{5pt}
\begin{table*}[ht]
\centering
\footnotesize

\begin{tabularx}{\textwidth}{ccccccccc}
\toprule
\multicolumn{9}{c}{\textbf{Llama-3-8B}}\\
\midrule
Task & Shot & Full & \multicolumn{1}{c}{1\% ind.\vspace{-0.25pt}} & \multicolumn{1}{c}{1\% ind.\vspace{-0.25pt}} & \multicolumn{1}{c}{1\% rnd.\vspace{-0.25pt}} & \multicolumn{1}{c}{3\% ind.\vspace{-0.25pt}} & \multicolumn{1}{c}{3\% ind.\vspace{-0.25pt}} & \multicolumn{1}{c}{3\% rnd.\vspace{-0.25pt}}\\
 &  & model & \multicolumn{1}{c}{heads} & \multicolumn{1}{c}{pattern} & \multicolumn{1}{c}{heads} & \multicolumn{1}{c}{heads} & \multicolumn{1}{c}{pattern} & \multicolumn{1}{c}{heads} \\
\midrule
Repetition & 5 & 67.3 & 61.9 (-6.4) & 62.5 (-4.8) & 64.9 (-2.4) & 50.0 (-17.3) & 53.2 (-14.1) & 62.6 (-4.7)\\
           & 10 & 91.3 & 59.7 (-31.6) & 65.3 (-26.0) & 85.9 (-5.4) & 51.5 (-39.8) & 58.7 (-32.6) & 79.7 (-11.6)\\
\midrule
Recursion & 5 & 67.8 & 63.0 (-4.8) & 61.5 (-6.3) & 66.1 (-1.7) & 52.0 (-15.8) & 55.1 (-12.7) & 68.0 (+0.2)\\
          & 10 & 91.5 & 67.9 (-23.6) & 68.7 (-22.8) & 86.5 (-5.0) & 52.9 (-38.6) & 58.9 (-32.6) & 84.6 (-6.9)\\
\midrule
Centre-embedding & 5 & 58.8 & 54.9 (-3.9) & 55.0 (-3.8) & 57.2 (-1.6) & 49.1 (-9.7) & 50.5 (-8.3) & 56.4 (-2.4)\\
                 & 10 & 80.4 & 53.0 (-27.4) & 56.5 (-23.9) & 74.6 (-5.8) & 50.7 (-29.7) & 52.3 (-28.1) & 71.5 (-8.9)\\
\midrule
WordSeq 1 (binary) & 5 & 83.1 & 72.1 (-11.0) & 71.8 (-11.3) & 82.1 (-1.0) & 51.6 (-31.5) & 56.9 (-26.2) & 78.8 (-4.3)\\
                   & 10 & 99.4 & 96.2 (-3.2) & 97.3 (-2.1) & 99.3 (-0.1) & 69.4 (-30.0) & 82.2 (-17.2) & 98.3 (-1.1)\\
\midrule
WordSeq 2 (binary) & 5 & 77.9 & 65.4 (-12.5) & 65.2 (-12.7) & 76.8 (-1.1) & 52.0 (-25.9) & 55.7 (-22.2) & 72.4 (-5.5)\\
                   & 10 & 99.4 & 94.9 (-4.5) & 96.4 (-3.0) & 98.8 (-0.6) & 67.2 (-32.2) & 81.1 (-18.3) & 97.7 (-1.7)\\
\midrule
WordSeq 1 (4-way) & 20 & 78.3 & 55.2 (-23.1) & 59.8 (-18.5) & 76.5 (-1.8) & 40.8 (-37.5) & 45.2 (-33.1) & 71.4 (-6.9)\\
\midrule
WordSeq 2 (4-way) & 20 & 81.3 & 55.9 (-25.4) & 59.8 (-21.5) & 76.0 (-5.3) & 42.3 (-39.0) & 47.5 (-33.8) & 68.6 (-12.7)\\
\toprule
\end{tabularx}
\footnotesize
\textbf{InternLM2-20B}
\begin{tabularx}{\textwidth}{cccccccccc}
\midrule
Task & Shot & Full & \multicolumn{1}{c}{1\% ind.\vspace{-0.25pt}} & \multicolumn{1}{c}{1\% ind.\vspace{-0.25pt}} & \multicolumn{1}{c}{1\% rnd.\vspace{-0.25pt}} & \multicolumn{1}{c}{3\% ind.\vspace{-0.25pt}} & \multicolumn{1}{c}{3\% ind.\vspace{-0.25pt}} & \multicolumn{1}{c}{3\% rnd.\vspace{-0.25pt}}\\
 &  & model & \multicolumn{1}{c}{heads} & \multicolumn{1}{c}{pattern} & \multicolumn{1}{c}{heads} & \multicolumn{1}{c}{heads} & \multicolumn{1}{c}{pattern} & \multicolumn{1}{c}{heads} \\
\midrule
Repetition & 5 & 68.7 & 63.3 (-5.4) & 62.0 (-6.7) & 67.6 (-1.1) & 62.4 (-6.3) & 58.5 (-10.2) & 64.4 (-4.3)\\
           & 10 & 88.1 & 73.4 (-14.7) & 72.1 (-16.0) & 86.9 (-1.2) & 72.5 (-15.6) & 61.2 (-26.9) & 84.1 (-4.0)\\
\midrule
Recursion & 5 & 68.1 & 62.1 (-6.0) & 59.9 (-8.2) & 67.3 (-0.8) & 59.5 (-8.6) & 55.3 (-12.8) & 65.3 (-2.8)\\
          & 10 & 88.5 & 70.3 (-18.2) & 69.9 (-18.6) & 87.1 (-1.4) & 67.5 (-21.0) & 57.1 (-31.4) & 85.1 (-3.4)\\
\midrule
Centre-embedding & 5 & 59.5 & 56.9 (-2.6) & 56.1 (-3.4) & 58.6 (-0.9) & 55.3 (-4.2) & 54.1 (-5.4) & 56.7 (-2.8)\\
                 & 10 & 75.3 & 60.1 (-15.2) & 58.5 (-16.8) & 74.3 (-1.0) & 60.2 (-15.1) & 54.3 (-21.0) & 70.7 (-4.6)\\
\midrule
WordSeq 1 (binary) & 5 & 76.8 & 66.5 (-10.3) & 64.9 (-11.9) & 76.1 (-0.7) & 61.5 (-15.3) & 56.8 (-20.0) & 77.6 (+0.8)\\
                   & 10 & 95.6 & 90.0 (-5.6) & 88.2 (-7.4) & 94.6 (-1.0) & 89.1 (-6.5) & 83.3 (-12.3) & 96.3 (+0.7)\\
\midrule
WordSeq 2 (binary) & 5 & 83.3 & 74.2 (-9.1) & 70.1 (-13.2) & 82.5 (-0.8) & 68.3 (-15.0) & 64.1 (-19.2) & 83.2 (-0.1)\\
                   & 10 & 99.0 & 98.1 (-0.9) & 96.8 (-2.2) & 99.0 & 94.1 (-4.9) & 86.7 (-12.3) & 98.7 (-0.3)\\
\midrule
WordSeq 1 (4-way) & 20 & 77.3 & 67.9 (-9.4) & 65.4 (-11.9) & 76.8 (-0.5) & 41.3 (-36.0) & 37.4 (-39.9) & 73.1 (-4.2)\\
\midrule
WordSeq 2 (4-way) & 20 & 80.4 & 78.3 (-2.1) & 75.5 (-4.9) & 79.4 (-1.0) & 54.8 (-25.6) & 47.5 (-32.9) & 74.6 (-5.8)\\
\bottomrule
\end{tabularx}

\caption{Llama-3-8B (top) and InternLM2-20B (bottom) ablation experiments on the abstract pattern recognition tasks. For both models, zero-shot performance of the full model is \textasciitilde$50\%$ in all tasks. Columns labelled "1\% ind. heads" and "3\% ind. heads" show results from fully ablating 1\% and 3\% of heads with the highest prefix scores, respectively. "1\% ind. pattern" and "3\% ind. pattern" columns depict outcomes from blocking induction attention patterns in 1\% and 3\% of these heads (Sec. \ref{sec:Knockout}). Columns "1\% rnd. heads" and "3\% rnd. heads" illustrate the effects of randomly ablating 1\% and 3\% of all heads in the model. Performance differences due to the ablation, compared to the full model, are indicated in parentheses.}
\label{tab:Abstr_Results_Mean}
\end{table*}
We make use of the semantically unrelated labels "Foo" and "Bar" for all binary tasks. For the four-way tasks, we additionally use "Mur" and "Res". For the letter-sequence datasets, we randomly generate sequences of the specified length, ensuring each dataset contains 500 examples with class balance. For the binary word-sequence datasets, we adapted the sampler to ensure it never samples examples with the same instantiation of categories as the query. For instance, "mango monkey: Foo" would not be sampled when the query is "mango shark:". This adjustment tests whether the induction heads are capable of the fuzzy pattern matching necessary for ICL, rather than just copying. Each binary word-sequence dataset contains 512 examples with class balance and each four-way word-sequence dataset 1024. Detailed prompt information can be found in Appendix \ref{sec:abstr_prompts}. We report five- and ten-shot ICL performance for the binary classification tasks, and 20-shot performance for the four-way tasks.

\paragraph{Llama-3-8B results} The results for Llama-3-8B are presented in Table \ref{tab:Abstr_Results_Mean}.
Few-shot prompting outperforms a zero-shot baseline whose performance is close to random across all tasks (\textasciitilde$50\%$ in the binary case), showing evidence of ICL. Ablating 1\% of induction heads leads to a substantial decrease in performance of up to 31.6\% across all tasks and settings. Increasing this to 3\% causes a further decline, with the letter-sequence tasks dropping to near-random performance. A similar trend is noted for the binary word-sequence tasks in the five-shot setting. Though the model achieves near-perfect performance in the ten-shot binary word-sequence tasks, induction head ablations still result in performance declines of up to 18.3\%. These substantial performance decreases strongly indicate that induction heads play a crucial role in few-shot ICL.\\
In contrast, a random ablation of 1\% has a much milder impact, reducing performance by no more than 5.8\% for the binary tasks, and 5.3\% for the four-way tasks. Although increasing this ablation to 3\% consistently leads to further performance decreases, these declines remain substantially milder compared to those observed with induction head ablation for all tasks and settings. This further confirms the importance of specifically induction heads in ICL. 

\paragraph{InternLM2-20B results}
The results for InternLM2-20B are presented in Table \ref{tab:Abstr_Results_Mean} (bottom). Few-shot prompting consistently outperforms a (close to random) zero-shot baseline across all tasks, providing clear evidence of ICL. The observed patterns are consistent with those seen in Llama-3-8B: Ablating 1\% of induction heads leads to a substantial decrease in performance, though these declines are not as pronounced as those seen in the Llama-3-8B model. The trend persists with the ablation of 3\% of induction heads, which results in performance decreases of up to 36\%. Ablating 3\% of randomly selected heads only results in a performance decrease of up to 5.8\%. These findings further affirm that induction heads play a crucial role in few-shot ICL, showing that these results generalise across models.

\section{Ablation Experiments: NLP Tasks}
\paragraph{Datasets and experimental setup}
To assess the impact of induction heads on ICL performance in real-world tasks, we conduct ablation experiments on a range of NLP classification tasks. We make use of the following datasets from the SuperGLUE benchmark \citep{wang2019superglue}: BoolQ (question answering), RTE (natural language inference), WIC (word-sense disambiguation) and WSC (coreference resolution). Additionally, we include ETHOS \citep{mollas2020ethos}\footnote{We employ a 80/20\% training-validation split.} (hate speech detection), SST-2 \citep{socher2013recursive} (sentiment analysis) and SUBJ \citep{conneau-kiela-2018-senteval} (subjectivity). All are binary classification tasks.

For all datasets except SST-2, we map the label space to target words "Yes"/"No", as this has been shown to improve ICL performance \citep{webson-pavlick-2022-prompt}.\footnote{For SST-2, we found this to actually lead to poor zero-shot performance, so we use the labels "Positive"/"Negative".} For BoolQ, ETHOS and SUBJ, we adopt the "Input/Output" template proposed by \citet{wei2023larger}. For the other datasets, we use informative prefixes for each element of the prompt (e.g. "Premise:"). Our prompts are shown in App. \ref{sec:NLP_prompts}. For all tasks, demonstration examples are selected from the training set and evaluations are performed on the validation set. We report ten-shot prompting performance.

We additionally conduct experiments with semantically-unrelated labels (SUL). Specifically, we map the label space from "Yes"/"No" to "Foo"/"Bar". This removes the semantic priors typically provided by labels, forcing the model to rely solely on input-label mappings for ICL \citep{wei2023larger}. Thus, we expect the model to strongly rely on induction heads to recognise patterns.

We define \textit{ICL benefit} as the metric quantifying the model's advantage from demonstration examples by measuring the performance difference between the ten-shot and zero-shot settings. For each model, we evaluate the full model's ICL benefit for each task and then quantify the percentage change in these benefits due to each ablation.\footnote{A full overview of model accuracy is shown in App. \ref{sec:NLP_Tables_Mean}.}

\begin{figure}[t!]
    \centering
    \includegraphics[width=0.75\columnwidth, height=7.5cm]{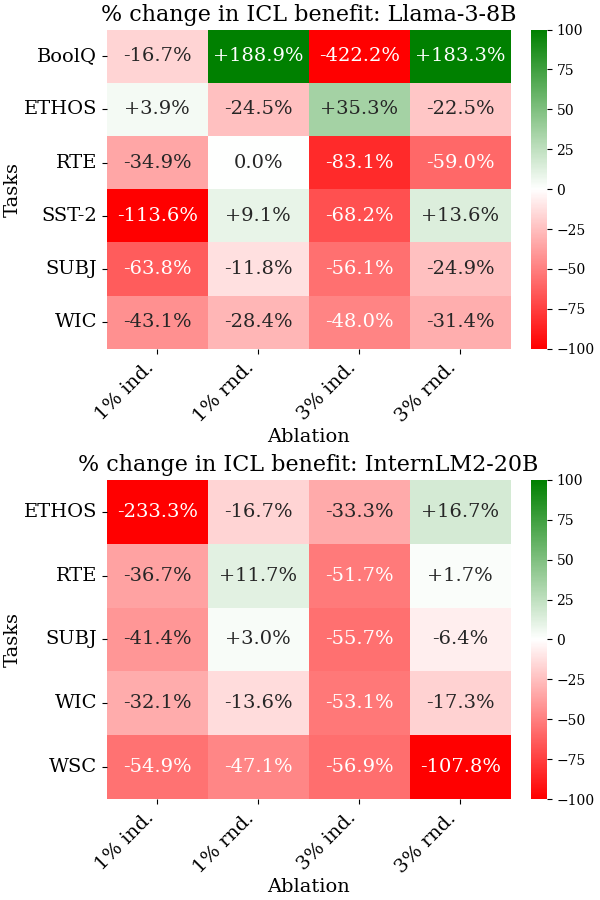}
        \caption{Change in ICL benefit for Llama-3-8B (top) and InternLM2-20B (bottom), due to head ablations when compared to that of the full model. "1\% ind." and "3\% ind." denote ablating the top respective percentage of induction heads. "1\% rnd." and "3\% rnd." denote randomly ablating the respective percentage of all heads in the model.}
    \label{fig:NLP_Mean}
\end{figure}

\paragraph{Llama-3-8B results} We observe ICL benefits for all tasks except WSC, which we exclude from further analysis. The impact of ablations on these benefits are detailed in Figure \ref{fig:NLP_Mean} (top).
Ablating 1\% of induction heads reduces the ICL benefit considerably more than a 1\% random ablation across all tasks except ETHOS. Specifically, in the SUBJ task, this targeted ablation results in a 63.8\% reduction, whereas the random ablation leads to only a 11.8\% decrease. For SST-2, we observe a decline of 113\% with targeted ablation, indicating that ten-shot performance has dropped below zero-shot level. This suggests that the model may no longer leverage demonstration examples effectively for this task. Increasing the induction head ablation to 3\% further decreases the ICL benefit for some tasks.

We further find that for the induction head ablations, the decrease in ICL benefit is due to ten-shot performance being affected more than zero-shot performance for all tasks except ETHOS (App. \ref{sec:NLP_Tables_Mean}). Such a pattern is not present in random head ablations. This supports the importance of induction heads for ICL specifically, as opposed to the general functioning of the LLM.
\begin{figure}[t!]
    \centering
    \footnotesize
    \textbf{SUL Ablation Experiments: Llama-3-8B}
    \includegraphics[width=\columnwidth, height=3cm]{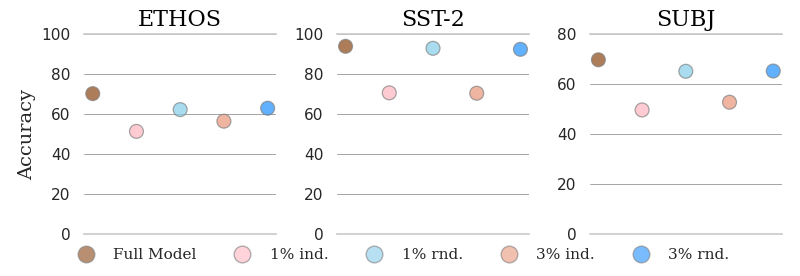}
    \footnotesize
    \textbf{SUL Ablation Experiments: InternLM2-20B}
    \includegraphics[width=\columnwidth, height=5.3cm]{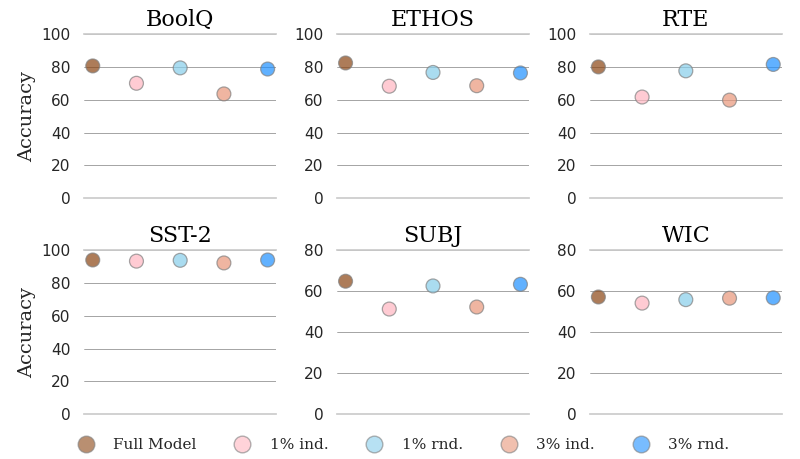}
    \caption{SUL ablation experiments for Llama-3-8B (top) and InternLM2-20B (bottom) on NLP tasks. "1\% ind." and "3\% ind." denote ablating the top respective percentage of induction heads. "1\% rnd." and "3\% rnd." denote randomly ablating the respective percentage of all heads in the model.}
    \label{fig:SUL_Mean}
\end{figure}

In the SUL setting, the model achieves above-random performance only for the ETHOS, SST-2 and SUBJ tasks, which are presented in Figure \ref{fig:SUL_Mean} (top). 
Ablating 1\% of induction heads leads to substantial performance declines of around 20\% for all tasks. For random ablations at the same level, these declines are considerably smaller. These findings suggest that the model's capacity to reference previous examples for learning input-label mappings may be compromised. Increasing the induction head ablation to 3\% did not reduce performance further, indicating that the heads responsible are amongst those with the highest prefix-scores.

\paragraph{InternLM2-20B results} We observe ICL benefits on all tasks with the exceptions of BoolQ and SST-2, which we therefore exclude from further analysis. The results for the remaining tasks are shown in Figure \ref{fig:NLP_Mean} (bottom). Consistent with observations from the Llama-3-8B model, we note that: (1) targeted ablations of induction heads negatively affect ten-shot performance more than zero-shot performance across all tasks, whereas random ablations do not follow this pattern (App. \ref{sec:NLP_Tables_Mean}); (2) a 1\% targeted ablation of induction heads diminishes the ICL benefit considerably more than a 1\% random ablation. Unlike in Llama-3-8B, increasing the targeted ablation to 3\% further reduces the ICL benefit in all tasks other than ETHOS. For WSC, the 3\% random ablation appears to decrease the ICL benefit more than the targeted ablation. However, this effect is partly attributed to an increase in zero-shot performance from the random ablation (App. \ref{sec:NLP_Tables_Mean}).

 In the SUL setting, the model achieves above-random performance in all tasks except for WSC and the results are presented in Fig. \ref{fig:SUL_Mean} (bottom). Induction head ablations consistently lead to performance declines of up to 20\%, which are generally substantially larger than those caused by random ablations across tasks, although the differences are smaller for SST-2 and WIC.

We observed similar trends for the five-shot setting (App. \ref{sec:NLP_Tables_Mean}). These consistent findings across different models and settings reaffirm our conclusions on the critical role of induction heads in ICL.

\section{Attention Knockout Experiments}
\label{sec:Knockout}

 \paragraph{Attention pattern analysis} To further explore the functional significance of induction heads, we leverage the structure of the word-sequence datasets. For instance, in the binary \textsc{WordSeq 1} task (classifying <fruit-animal> pairs vs. <furniture-profession> ones), if an induction head focuses on a \textit{furniture} token, it should typically attend to \textit{profession} tokens that have historically followed \textit{furniture}. To validate this, we conduct a qualitative analysis of the attention patterns exhibited by the top five induction heads with the highest prefix matching scores from Llama-3-8B on the binary and four-way versions of the \textsc{WordSeq 1} task (App. \ref{sec:att_pattern}). 
 
\begin{figure}
 \centering
 \begin{minipage}{0.48\columnwidth}
 \centering
   \includegraphics[scale=0.33]{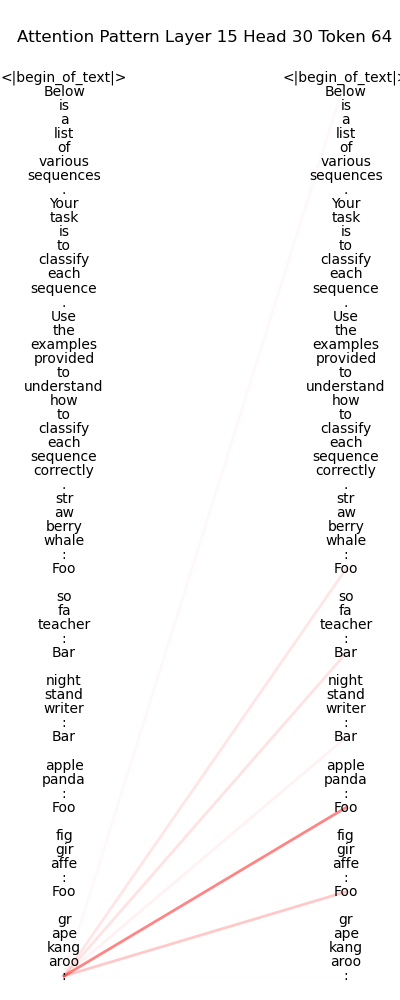}
   \caption{Attention pattern for the final ":" token in head 30 of layer 15 in the Llama-3-8B model.}
 \label{fig:att_pattern_final}
 \end{minipage}\hfill
 \begin{minipage}{0.48\columnwidth}
 \centering
   \includegraphics[scale=0.33]{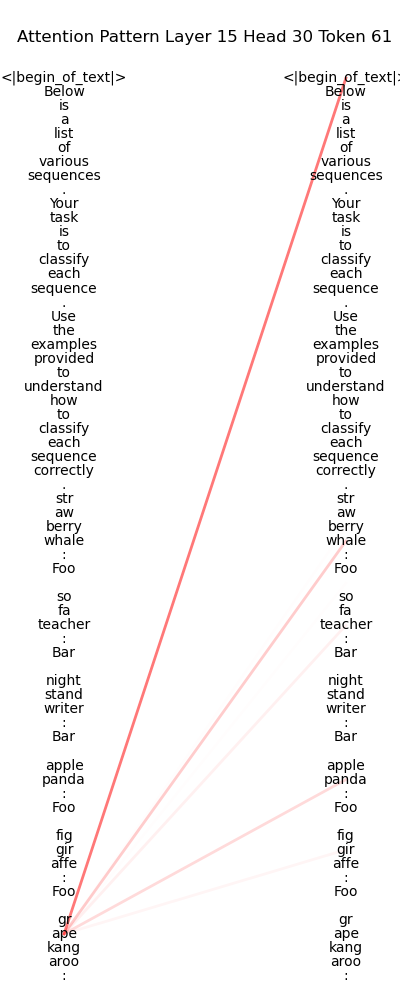}
   \caption{Attention pattern for the "ape" token in head 30 of layer 15 in the Llama-3-8B model.}
 \label{fig:att_pattern_ape}
 \end{minipage}
 \end{figure}

 Figure \ref{fig:att_pattern_final} illustrates the attention pattern of the highest scoring induction head for the final ":" token on a sample from the binary task. This token predominantly attends to label tokens, particularly focusing on "Foo," the target label in this instance. Such patterns, while not universally consistent, are particularly evident in induction heads located in later layers, suggesting that induction mechanisms may facilitate the model's ability to selectively focus on relevant labels in certain contexts.

Figure \ref{fig:att_pattern_ape} displays the attention pattern for the "ape" token (from 'grape'). We observe that the head attends to tokens that previously followed fruit tokens. This behaviour was consistent across token categories and observed in various heads and samples, suggesting that induction heads not only learn input-label mappings, but may also grasp the underlying pattern itself. 

\paragraph{Experimental setup} To disrupt this induction pattern, we block attention from each token to all tokens that directly followed tokens of the same category, including word categories, labels, newlines, and colons. In contrast, the letter-sequence tasks require recognising abstract patterns rather than categorical sequences. Thus, our blocking strategy is exclusively focused on labels, newlines, and colons.

\paragraph{Llama-3-8B results} 
Table~\ref{tab:Abstr_Results} (top) shows that blocking the induction pattern for 1\% of the induction heads results in performance declines comparable to those observed with complete head ablations. When the block is increased to 3\% of the induction heads, performance declines are similarly consistent in most settings. Though in the ten-shot binary word sequence tasks the declines are notably lower, they are still substantial, exceeding 17\% for both tasks. These persistent declines across both full and pattern-specific ablations provide strong empirical evidence that induction heads predominantly rely on this attention pattern.

\paragraph{InternLM2-20B results}
Table~\ref{tab:Abstr_Results} (bottom) shows that blocking the induction pattern in 1\% of induction heads results in performance declines within 4.1\% of full head ablations. Unlike in Llama-3-8B, this approach consistently causes greater performance decreases than full ablations. This gap widens further when the knockout is applied to 3\% of the induction heads. These observations indicate that for this model, blocking the induction pattern has a more detrimental impact on head functionality than eliminating the head entirely, suggesting that the induction heads are dependent on their specific operational patterns for performance.

\section{Conclusion}
In this paper, we explored the extent to which the prefix matching and copying capabilities of induction heads play a role in few-shot ICL. Our findings highlight a general trend where the ablation of induction heads affects few-shot performance more severely than zero-shot performance. This observation underscores the significance of induction heads in enabling the model to learn effectively from a limited number of examples. Moreover, 
even a minimal ablation of just 1\% of these heads results in a substantial decrease in ICL, suggesting that induction heads are crucial for the model’s few-shot learning capabilities. 
Additionally, when the induction pattern is blocked for a percentage of these heads, performance drops to levels comparable to full head ablations. This provides further empirical evidence that induction heads predominantly utilise this induction pattern and are dependent on it for optimal performance.
Overall, our findings indicate that induction heads are a fundamental mechanism underlying ICL. 

\section*{Limitations}
The limitations of this study are twofold. First, we lack a mechanistic basis for definitively identifying induction heads; instead, we rely on prefix-matching scores as a proxy. Consequently, it remains unclear how the grouped-query attention mechanism influences the circuits and whether induction heads form differently as a result. Second, our investigation of induction attention patterns (attention knockout) is confined to abstract-pattern recognition tasks. Further research is required to verify these findings across NLP tasks.

Furthermore, it is worth noting that our head ablation and attention knockout experiments demonstrate that induction heads play an important role in ICL. However, what we do not test for is whether they are the only computational process in the Transformer LM that plays such a role (this can not in principle be tested by a targeted ablation or blocking experiment where only a particular part of the computational process is disabled). Therefore, it is possible that additional mechanisms and/or circuits will be discovered in the future that contribute to ICL, alongside induction heads.
\bibliography{references}

\begin{thebibliography}{33}
\providecommand{\natexlab}[1]{#1}

\bibitem[{Bansal et~al.(2023)Bansal, Gopalakrishnan, Dingliwal, Bodapati, Kirchhoff, and Roth}]{bansal-etal-2023-rethinking}
Hritik Bansal, Karthik Gopalakrishnan, Saket Dingliwal, Sravan Bodapati, Katrin Kirchhoff, and Dan Roth. 2023.
\newblock \href {https://doi.org/10.18653/v1/2023.acl-long.660} {Rethinking the role of scale for in-context learning: An interpretability-based case study at 66 billion scale}.
\newblock In \emph{Proceedings of the 61st Annual Meeting of the Association for Computational Linguistics (Volume 1: Long Papers)}, pages 11833--11856, Toronto, Canada. Association for Computational Linguistics.

\bibitem[{Brown et~al.(2020)Brown, Mann, Ryder, Subbiah, Kaplan, Dhariwal, Neelakantan, Shyam, Sastry, Askell et~al.}]{brown2020language}
Tom Brown, Benjamin Mann, Nick Ryder, Melanie Subbiah, Jared~D Kaplan, Prafulla Dhariwal, Arvind Neelakantan, Pranav Shyam, Girish Sastry, Amanda Askell, et~al. 2020.
\newblock Language models are few-shot learners.
\newblock \emph{Advances in neural information processing systems}, 33:1877--1901.

\bibitem[{Cai et~al.(2024)Cai, Cao, Chen, Chen, Chen, Chen, Chen, Chen, Chen, Chu et~al.}]{cai2024internlm2}
Zheng Cai, Maosong Cao, Haojiong Chen, Kai Chen, Keyu Chen, Xin Chen, Xun Chen, Zehui Chen, Zhi Chen, Pei Chu, et~al. 2024.
\newblock Internlm2 technical report.
\newblock \emph{arXiv preprint arXiv:2403.17297}.

\bibitem[{Chan et~al.(2022)Chan, Santoro, Lampinen, Wang, Singh, Richemond, McClelland, and Hill}]{chan2022data}
Stephanie Chan, Adam Santoro, Andrew Lampinen, Jane Wang, Aaditya Singh, Pierre Richemond, James McClelland, and Felix Hill. 2022.
\newblock Data distributional properties drive emergent in-context learning in transformers.
\newblock \emph{Advances in Neural Information Processing Systems}, 35:18878--18891.

\bibitem[{Conneau and Kiela(2018)}]{conneau-kiela-2018-senteval}
Alexis Conneau and Douwe Kiela. 2018.
\newblock \href {https://aclanthology.org/L18-1269} {{S}ent{E}val: An evaluation toolkit for universal sentence representations}.
\newblock In \emph{Proceedings of the Eleventh International Conference on Language Resources and Evaluation ({LREC} 2018)}, Miyazaki, Japan. European Language Resources Association (ELRA).

\bibitem[{Dai et~al.(2023)Dai, Sun, Dong, Hao, Ma, Sui, and Wei}]{dai-etal-2023-gpt}
Damai Dai, Yutao Sun, Li~Dong, Yaru Hao, Shuming Ma, Zhifang Sui, and Furu Wei. 2023.
\newblock \href {https://doi.org/10.18653/v1/2023.findings-acl.247} {Why can {GPT} learn in-context? language models secretly perform gradient descent as meta-optimizers}.
\newblock In \emph{Findings of the Association for Computational Linguistics: ACL 2023}, pages 4005--4019, Toronto, Canada. Association for Computational Linguistics.

\bibitem[{Dubey et~al.(2024)Dubey, Jauhri, Pandey, Kadian, Al-Dahle, Letman, Mathur, Schelten, Yang, Fan et~al.}]{dubey2024llama}
Abhimanyu Dubey, Abhinav Jauhri, Abhinav Pandey, Abhishek Kadian, Ahmad Al-Dahle, Aiesha Letman, Akhil Mathur, Alan Schelten, Amy Yang, Angela Fan, et~al. 2024.
\newblock The llama 3 herd of models.
\newblock \emph{arXiv preprint arXiv:2407.21783}.

\bibitem[{Elhage et~al.(2021)Elhage, Nanda, Olsson, Henighan, Joseph, Mann, Askell, Bai, Chen, Conerly et~al.}]{elhage2021mathematical}
Nelson Elhage, Neel Nanda, Catherine Olsson, Tom Henighan, Nicholas Joseph, Ben Mann, Amanda Askell, Yuntao Bai, Anna Chen, Tom Conerly, et~al. 2021.
\newblock A mathematical framework for transformer circuits.
\newblock \emph{Transformer Circuits Thread}, 1.

\bibitem[{Gao et~al.(2023)Gao, Tow, Abbasi, Biderman, Black, DiPofi, Foster, Golding, Hsu, Le~Noac'h, Li, McDonell, Muennighoff, Ociepa, Phang, Reynolds, Schoelkopf, Skowron, Sutawika, Tang, Thite, Wang, Wang, and Zou}]{eval-harness}
Leo Gao, Jonathan Tow, Baber Abbasi, Stella Biderman, Sid Black, Anthony DiPofi, Charles Foster, Laurence Golding, Jeffrey Hsu, Alain Le~Noac'h, Haonan Li, Kyle McDonell, Niklas Muennighoff, Chris Ociepa, Jason Phang, Laria Reynolds, Hailey Schoelkopf, Aviya Skowron, Lintang Sutawika, Eric Tang, Anish Thite, Ben Wang, Kevin Wang, and Andy Zou. 2023.
\newblock \href {https://doi.org/10.5281/zenodo.10256836} {A framework for few-shot language model evaluation}.

\bibitem[{Geva et~al.(2023)Geva, Bastings, Filippova, and Globerson}]{geva-etal-2023-dissecting}
Mor Geva, Jasmijn Bastings, Katja Filippova, and Amir Globerson. 2023.
\newblock \href {https://doi.org/10.18653/v1/2023.emnlp-main.751} {Dissecting recall of factual associations in auto-regressive language models}.
\newblock In \emph{Proceedings of the 2023 Conference on Empirical Methods in Natural Language Processing}, pages 12216--12235, Singapore. Association for Computational Linguistics.

\bibitem[{Lu et~al.(2023)Lu, Bigoulaeva, Sachdeva, Madabushi, and Gurevych}]{lu2023emergent}
Sheng Lu, Irina Bigoulaeva, Rachneet Sachdeva, Harish~Tayyar Madabushi, and Iryna Gurevych. 2023.
\newblock Are emergent abilities in large language models just in-context learning?
\newblock \emph{arXiv preprint arXiv:2309.01809}.

\bibitem[{Meng et~al.(2022)Meng, Bau, Andonian, and Belinkov}]{meng2022locating}
Kevin Meng, David Bau, Alex Andonian, and Yonatan Belinkov. 2022.
\newblock Locating and editing factual associations in gpt.
\newblock \emph{Advances in Neural Information Processing Systems}, 35:17359--17372.

\bibitem[{Min et~al.(2022)Min, Lyu, Holtzman, Artetxe, Lewis, Hajishirzi, and Zettlemoyer}]{min-etal-2022-rethinking}
Sewon Min, Xinxi Lyu, Ari Holtzman, Mikel Artetxe, Mike Lewis, Hannaneh Hajishirzi, and Luke Zettlemoyer. 2022.
\newblock \href {https://doi.org/10.18653/v1/2022.emnl} {Rethinking the role of demonstrations: What makes in-context learning work?}
\newblock In \emph{Proceedings of the 2022 Conference on Empirical Methods in Natural Language Processing}, Abu Dhabi, United Arab Emirates. Association for Computational Linguistics.

\bibitem[{Mollas et~al.(2020)Mollas, Chrysopoulou, Karlos, and Tsoumakas}]{mollas2020ethos}
Ioannis Mollas, Zoe Chrysopoulou, Stamatis Karlos, and Grigorios Tsoumakas. 2020.
\newblock Ethos: an online hate speech detection dataset.
\newblock \emph{arXiv preprint arXiv:2006.08328}.

\bibitem[{Olah et~al.(2020)Olah, Cammarata, Schubert, Goh, Petrov, and Carter}]{olah2020zoom}
Chris Olah, Nick Cammarata, Ludwig Schubert, Gabriel Goh, Michael Petrov, and Shan Carter. 2020.
\newblock Zoom in: An introduction to circuits.
\newblock \emph{Distill}, 5(3):e00024--001.

\bibitem[{Olsson et~al.(2022)Olsson, Elhage, Nanda, Joseph, DasSarma, Henighan, Mann, Askell, Bai, Chen et~al.}]{olsson2022context}
Catherine Olsson, Nelson Elhage, Neel Nanda, Nicholas Joseph, Nova DasSarma, Tom Henighan, Ben Mann, Amanda Askell, Yuntao Bai, Anna Chen, et~al. 2022.
\newblock In-context learning and induction heads.
\newblock \emph{arXiv preprint arXiv:2209.11895}.

\bibitem[{Radford et~al.(2019)Radford, Wu, Child, Luan, Amodei, Sutskever et~al.}]{radford2019language}
Alec Radford, Jeffrey Wu, Rewon Child, David Luan, Dario Amodei, Ilya Sutskever, et~al. 2019.
\newblock Language models are unsupervised multitask learners.
\newblock \emph{OpenAI blog}, 1(8):9.

\bibitem[{Ren et~al.(2024)Ren, Guo, Yan, Liu, Qiu, and Lin}]{ren2024identifying}
Jie Ren, Qipeng Guo, Hang Yan, Dongrui Liu, Xipeng Qiu, and Dahua Lin. 2024.
\newblock Identifying semantic induction heads to understand in-context learning.
\newblock \emph{arXiv preprint arXiv:2402.13055}.

\bibitem[{Socher et~al.(2013)Socher, Perelygin, Wu, Chuang, Manning, Ng, and Potts}]{socher2013recursive}
Richard Socher, Alex Perelygin, Jean Wu, Jason Chuang, Christopher~D Manning, Andrew~Y Ng, and Christopher Potts. 2013.
\newblock Recursive deep models for semantic compositionality over a sentiment treebank.
\newblock In \emph{Proceedings of the 2013 conference on empirical methods in natural language processing}, pages 1631--1642.

\bibitem[{Touvron et~al.(2023{\natexlab{a}})Touvron, Lavril, Izacard, Martinet, Lachaux, Lacroix, Rozi{\`e}re, Goyal, Hambro, Azhar et~al.}]{touvron2023llama1}
Hugo Touvron, Thibaut Lavril, Gautier Izacard, Xavier Martinet, Marie-Anne Lachaux, Timoth{\'e}e Lacroix, Baptiste Rozi{\`e}re, Naman Goyal, Eric Hambro, Faisal Azhar, et~al. 2023{\natexlab{a}}.
\newblock Llama: Open and efficient foundation language models.
\newblock \emph{arXiv preprint arXiv:2302.13971}.

\bibitem[{Touvron et~al.(2023{\natexlab{b}})Touvron, Martin, Stone, Albert, Almahairi, Babaei, Bashlykov, Batra, Bhargava, Bhosale et~al.}]{touvron2023llama2}
Hugo Touvron, Louis Martin, Kevin Stone, Peter Albert, Amjad Almahairi, Yasmine Babaei, Nikolay Bashlykov, Soumya Batra, Prajjwal Bhargava, Shruti Bhosale, et~al. 2023{\natexlab{b}}.
\newblock Llama 2: Open foundation and fine-tuned chat models.
\newblock \emph{arXiv preprint arXiv:2307.09288}.

\bibitem[{Vig et~al.(2020)Vig, Gehrmann, Belinkov, Qian, Nevo, Singer, and Shieber}]{NEURIPS2020_92650b2e}
Jesse Vig, Sebastian Gehrmann, Yonatan Belinkov, Sharon Qian, Daniel Nevo, Yaron Singer, and Stuart Shieber. 2020.
\newblock \href {https://proceedings.neurips.cc/paper_files/paper/2020/file/92650b2e92217715fe312e6fa7b90d82-Paper.pdf} {Investigating gender bias in language models using causal mediation analysis}.
\newblock In \emph{Advances in Neural Information Processing Systems}, volume~33, pages 12388--12401. Curran Associates, Inc.

\bibitem[{Von~Oswald et~al.(2023)Von~Oswald, Niklasson, Randazzo, Sacramento, Mordvintsev, Zhmoginov, and Vladymyrov}]{von2023transformers}
Johannes Von~Oswald, Eyvind Niklasson, Ettore Randazzo, Jo{\~a}o Sacramento, Alexander Mordvintsev, Andrey Zhmoginov, and Max Vladymyrov. 2023.
\newblock Transformers learn in-context by gradient descent.
\newblock In \emph{International Conference on Machine Learning}, pages 35151--35174. PMLR.

\bibitem[{Wang et~al.(2019)Wang, Pruksachatkun, Nangia, Singh, Michael, Hill, Levy, and Bowman}]{wang2019superglue}
Alex Wang, Yada Pruksachatkun, Nikita Nangia, Amanpreet Singh, Julian Michael, Felix Hill, Omer Levy, and Samuel Bowman. 2019.
\newblock Superglue: A stickier benchmark for general-purpose language understanding systems.
\newblock \emph{Advances in neural information processing systems}, 32.

\bibitem[{Wang et~al.(2022)Wang, Variengien, Conmy, Shlegeris, and Steinhardt}]{wang2022interpretability}
Kevin Wang, Alexandre Variengien, Arthur Conmy, Buck Shlegeris, and Jacob Steinhardt. 2022.
\newblock Interpretability in the wild: a circuit for indirect object identification in gpt-2 small.
\newblock \emph{arXiv preprint arXiv:2211.00593}.

\bibitem[{Wang et~al.(2023)Wang, Li, Dai, Chen, Zhou, Meng, Zhou, and Sun}]{wang-etal-2023-label}
Lean Wang, Lei Li, Damai Dai, Deli Chen, Hao Zhou, Fandong Meng, Jie Zhou, and Xu~Sun. 2023.
\newblock \href {https://doi.org/10.18653/v1/2023.emnlp-main.609} {Label words are anchors: An information flow perspective for understanding in-context learning}.
\newblock In \emph{Proceedings of the 2023 Conference on Empirical Methods in Natural Language Processing}, pages 9840--9855, Singapore. Association for Computational Linguistics.

\bibitem[{Webson and Pavlick(2022)}]{webson-pavlick-2022-prompt}
Albert Webson and Ellie Pavlick. 2022.
\newblock \href {https://doi.org/10.18653/v1/2022.naacl-main.167} {Do prompt-based models really understand the meaning of their prompts?}
\newblock In \emph{Proceedings of the 2022 Conference of the North American Chapter of the Association for Computational Linguistics: Human Language Technologies}, pages 2300--2344, Seattle, United States. Association for Computational Linguistics.

\bibitem[{Wei et~al.(2022)Wei, Tay, Bommasani, Raffel, Zoph, Borgeaud, Yogatama, Bosma, Zhou, Metzler et~al.}]{wei2022emergent}
Jason Wei, Yi~Tay, Rishi Bommasani, Colin Raffel, Barret Zoph, Sebastian Borgeaud, Dani Yogatama, Maarten Bosma, Denny Zhou, Donald Metzler, et~al. 2022.
\newblock Emergent abilities of large language models.
\newblock \emph{arXiv preprint arXiv:2206.07682}.

\bibitem[{Wei et~al.(2023{\natexlab{a}})Wei, Hou, Lampinen, Chen, Huang, Tay, Chen, Lu, Zhou, Ma, and Le}]{wei-etal-2023-symbol}
Jerry Wei, Le~Hou, Andrew Lampinen, Xiangning Chen, Da~Huang, Yi~Tay, Xinyun Chen, Yifeng Lu, Denny Zhou, Tengyu Ma, and Quoc Le. 2023{\natexlab{a}}.
\newblock \href {https://doi.org/10.18653/v1/2023.emnlp-main.61} {Symbol tuning improves in-context learning in language models}.
\newblock In \emph{Proceedings of the 2023 Conference on Empirical Methods in Natural Language Processing}, pages 968--979, Singapore. Association for Computational Linguistics.

\bibitem[{Wei et~al.(2023{\natexlab{b}})Wei, Wei, Tay, Tran, Webson, Lu, Chen, Liu, Huang, Zhou et~al.}]{wei2023larger}
Jerry Wei, Jason Wei, Yi~Tay, Dustin Tran, Albert Webson, Yifeng Lu, Xinyun Chen, Hanxiao Liu, Da~Huang, Denny Zhou, et~al. 2023{\natexlab{b}}.
\newblock Larger language models do in-context learning differently.
\newblock \emph{arXiv preprint arXiv:2303.03846}.

\bibitem[{Williams et~al.(2018)Williams, Nangia, and Bowman}]{williams-etal-2018-broad}
Adina Williams, Nikita Nangia, and Samuel Bowman. 2018.
\newblock \href {https://doi.org/10.18653/v1/N18-1101} {A broad-coverage challenge corpus for sentence understanding through inference}.
\newblock In \emph{Proceedings of the 2018 Conference of the North {A}merican Chapter of the Association for Computational Linguistics: Human Language Technologies, Volume 1 (Long Papers)}, pages 1112--1122, New Orleans, Louisiana. Association for Computational Linguistics.

\bibitem[{Wu et~al.(2024)Wu, Geiger, Arora, Huang, Wang, Goodman, Manning, and Potts}]{wu-etal-2024-pyvene}
Zhengxuan Wu, Atticus Geiger, Aryaman Arora, Jing Huang, Zheng Wang, Noah Goodman, Christopher Manning, and Christopher Potts. 2024.
\newblock \href {https://aclanthology.org/2024.naacl-demo.16} {pyvene: A library for understanding and improving {P}y{T}orch models via interventions}.
\newblock In \emph{Proceedings of the 2024 Conference of the North American Chapter of the Association for Computational Linguistics: Human Language Technologies (Volume 3: System Demonstrations)}, pages 158--165, Mexico City, Mexico. Association for Computational Linguistics.

\bibitem[{Zhang et~al.(2022)Zhang, Roller, Goyal, Artetxe, Chen, Chen, Dewan, Diab, Li, Lin et~al.}]{zhang2022opt}
Susan Zhang, Stephen Roller, Naman Goyal, Mikel Artetxe, Moya Chen, Shuohui Chen, Christopher Dewan, Mona Diab, Xian Li, Xi~Victoria Lin, et~al. 2022.
\newblock Opt: Open pre-trained transformer language models.
\newblock \emph{arXiv preprint arXiv:2205.01068}.

\end{thebibliography}
\clearpage
\appendix
\onecolumn
\section{Appendix}

\label{sec:appendix}

\subsection{Prefix Matching Scores IntermLM-20B}
\label{sec:pfx_intern}
\begin{figure}[ht]
    \centering
    \includegraphics[scale=0.25]{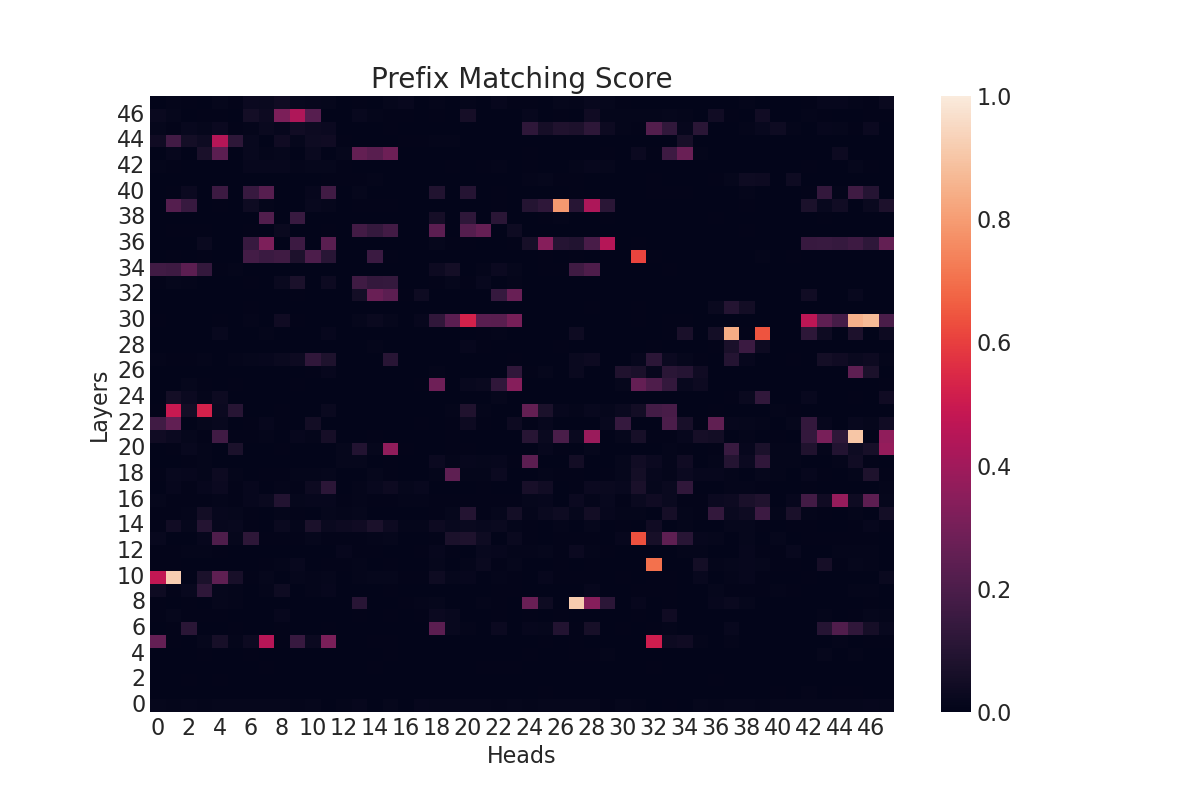}
    \caption{Prefix matching scores for InternLM2-20B}
    \label{fig:pfx_internlm}
\end{figure}

\subsection{Seeds}
\label{sec:seeds}
\begin{table*}[h!]
  \centering
  \begin{tabular}{ll}
    \hline
    \textbf{Task}           & \textbf{Seeds} \\
    \hline
    Prefix Matching Sequences           & 0, 1, 2, 3, 4\\
    Demonstration Example Selection     & 42, 43, 44 \\
    Random Head Ablation                & 42, 43, 44 \\             
    \hline
  \end{tabular}
  \caption{Seed details.}
\end{table*}

\subsection{Prompt Examples}
We provide five-shot prompting examples for all binary tasks and 20-shot prompting examples for all four-way tasks. 
\label{sec:prompts}
\subsubsection{Abstract Pattern Recognition Tasks}
\label{sec:abstr_prompts}
\textbf{Repetition:}\\
\texttt{Below is a list of various sequences. Your task is to classify
each sequence. Use the examples provided to understand how to
classify each sequence correctly.\\\\
D Y P A:Bar\\\\
M B K O:Bar\\\\
P L P L:Foo\\\\
V S V S:Foo\\\\
S I X P:Bar\\\\
J E J E:}\\\\
\textbf{Recursion:}\\
\texttt{Below is a list of various sequences. Your task is to classify each sequence. Use the examples provided to understand how to classify each sequence correctly.\\\\
H U V J:Bar\\\\
Q R T K:Bar\\\\
X X X X:Foo\\\\
U Q Q Q:Foo\\\\
O R P L:Bar\\\\
V H H H:}\\\\
\textbf{Centre-embedding:}\\
\texttt{Below is a list of various sequences. Your task is to classify each sequence. Use the examples provided to understand how to classify each sequence correctly.\\\\
Q T J D G:Bar\\\\
R Y P O N:Bar\\\\
R Y I Y R:Foo\\\\
C B V B C:Foo\\\\
W A Y S M:Bar\\\\
N F I F N:}\\\\
\textbf{WordSeq 1 (binary):}\\
\texttt{Below is a list of various sequences. Your task is to classify each sequence. Use the examples provided to understand how to classify each sequence correctly.\\\\
strawberry cat:Foo\\\\
cherry rabbit:Foo\\\\
nightstand nurse:Bar\\\\
grape rabbit:Foo\\\\
couch artist:Bar\\\\
cabinet scientist:}\\\\
\textbf{WordSeq 1 (four-way):}\\
\texttt{Below is a list of various sequences. Your task is to classify each sequence. Use the examples provided to understand how to classify each sequence correctly.\\\\
papaya writer:Res\\\\
ottoman zebra:Mur\\\\
desk chef:Bar\\\\
mango chef:Res\\\\
shelf lion:Mur\\\\
nectarine cat:Foo\\\\
chair pilot:Bar\\\\
fig doctor:Res\\\\
bench scientist:Bar\\\\
mango horse:Foo\\\\
chair lawyer:Bar\\\\
mango lion:Foo\\\\
chair scientist:Bar\\\\
couch shark:Mur\\\\
kiwi farmer:Res\\\\
cabinet tiger:Mur\\\\
papaya actor:Res\\\\
recliner monkey:Mur\\\\
lemon elephant:Foo\\\\
desk lawyer:Bar\\\\
bookcase electrician:}\\\\
\textbf{WordSeq 2 (binary):}\\
\texttt{Below is a list of various sequences. Your task is to classify each sequence. Use the examples provided to understand how to classify each sequence correctly.\\\\
potato yacht:Foo\\\\
garlic minivan:Foo\\\\
neck banjo:Bar\\\\
spinach trolley:Foo\\\\
ankle drum:Bar\\\\
arm oboe:}\\\\
\textbf{WordSeq 2 (four-way):}\\
\texttt{Below is a list of various sequences. Your task is to classify each sequence. Use the examples provided to understand how to classify each sequence correctly.\\\\
celery helicopter:Foo\\\\
lettuce violin:Mur\\\\
mouth train:Res\\\\
broccoli bus:Foo\\\\
pumpkin accordion:Mur\\\\
arm accordion:Bar\\\\
foot helicopter:Res\\\\
lettuce trolley:Foo\\\\
eye boat:Res\\\\
mouth piano:Bar\\\\
arm plane:Res\\\\
knee guitar:Bar\\\\
arm plane:Res\\\\
carrot saxophone:Mur\\\\
cucumber scooter:Foo\\\\
eggplant harmonica:Mur\\\\
garlic scooter:Foo\\\\
onion trombone:Mur\\\\
toe cello:Bar\\\\
arm bicycle:Res\\\\
finger submarine:}\\

\subsubsection{NLP Tasks}
\label{sec:NLP_prompts}
\textbf{BoolQ}\\
\texttt{Read the passage and answer the question.\\\\
Input: Natural-born-citizen clause -- Status as a natural-born citizen of the United States is one of the eligibility requirements established in the United States Constitution for holding the office of President or Vice President. This requirement was intended to protect the nation from foreign influence.\\\\
Can a canadian be president of the united states?\\\\
Output: No\\\\
Input: Pepsi Zero Sugar -- Pepsi Zero Sugar (sold under the names Diet Pepsi Max until early 2009 and then Pepsi Max until August 2016), is a zero-calorie, sugar-free, carbohydrate-free, ginseng-infused cola sweetened with aspartame, marketed by PepsiCo. In Fall 2016, PepsiCo renamed the drink Pepsi Zero Sugar from Pepsi Max. It has nearly twice the caffeine of Pepsi's other cola beverages. Pepsi Zero Sugar contains 69 milligrams of caffeine per 355ml (12 fl oz), versus 36 milligrams in Diet Pepsi.\\\\
Is pepsi zero sugar the same as diet pepsi?\\\\
Output: No\\\\
Input: Phone hacking -- Phone hacking, being a form of surveillance, is illegal in many countries unless it is carried out as lawful interception by a government agency. In the News International phone hacking scandal, private investigator Glenn Mulcaire was found to have violated the Regulation of Investigatory Powers Act 2000. He was sentenced to six months in prison in January 2007. Renewed controversy over the phone hacking claims led to the closure of the News of the World in July 2011.\\\\
Is it illegal to hack into someones phone?\\\\
Output: Yes\\\\
Input: Legal threat -- A legal threat is a statement by a party that it intends to take legal action on another party, generally accompanied by a demand that the other party take an action demanded by the first party or refrain from taking or continuing actions objected to by the demanding party.\\\\
Can you threaten to take someone to court?\\\\
Output: Yes\\\\
Input: Devil's food cake -- Devil's food cake is a moist, airy, rich chocolate layer cake. It is considered a counterpart to the white or yellow angel food cake. Because of differing recipes and changing ingredient availability over the course of the 20th century, it is difficult to precisely qualify what distinguishes devil's food from the more standard chocolate cake, though it traditionally has more chocolate than a regular chocolate cake, making it darker. The cake is usually paired with a rich chocolate frosting.\\\\
Is devil's food cake the same as chocolate?\\\\
Output: No\\\\
Input: Ethanol fuel -- All biomass goes through at least some of these steps: it needs to be grown, collected, dried, fermented, distilled, and burned. All of these steps require resources and an infrastructure. The total amount of energy input into the process compared to the energy released by burning the resulting ethanol fuel is known as the energy balance (or ``energy returned on energy invested''). Figures compiled in a 2007 report by National Geographic Magazine point to modest results for corn ethanol produced in the US: one unit of fossil-fuel energy is required to create 1.3 energy units from the resulting ethanol. The energy balance for sugarcane ethanol produced in Brazil is more favorable, with one unit of fossil-fuel energy required to create 8 from the ethanol. Energy balance estimates are not easily produced, thus numerous such reports have been generated that are contradictory. For instance, a separate survey reports that production of ethanol from sugarcane, which requires a tropical climate to grow productively, returns from 8 to 9 units of energy for each unit expended, as compared to corn, which only returns about 1.34 units of fuel energy for each unit of energy expended. A 2006 University of California Berkeley study, after analyzing six separate studies, concluded that producing ethanol from corn uses much less petroleum than producing gasoline.\\\\
Does ethanol take more energy make that produces?\\\\
Output:}\\\\
\textbf{ETHOS}\\
\texttt{Does the input contain hate speech?\\\\
Input: My God, the cancer is spreading\\\\
Output: No\\\\
Input: I’m searching for a hitman right now to kill this whore and every other like her. H A T E W O M E N\\\\
Output: Yes\\\\
Input: why is it censored, why does todays generation get butt hurf and offended so easily, if u dont like it, keep walking then! this use to be a free country but its going down the toilet now.\\\\
Output: No\\\\
Input: Stand the fuck up and take them out all you fucking lazy white males! fuck being called a racist protect your woman and children. If you find it in the street kill it and bury it right there these animals will wipe you out if your not careful\\\\
Output: Yes\\\\
Input: Ugly white girls and white dudes would not get away with that.\\\\
Output: Yes\\\\
Input: Watching this kid trying to step up and walk with that crutches is the best thing in the world. So idiot\\\\
Output:}\\\\
\textbf{RTE}\\
\texttt{You are given a premise and a hypothesis. Given the premise, tell me if the hypothesis is correct.\\\\
Premise: Les Paul, who continues to perform weekly at New York Iridium Jazz Club, has finished recording "Les Paul \& Friends."\\
Hypothesis: Iridium Jazz Club is located in New York.\\\\
Answer: Yes\\\\
Premise: The discovery of the body of a warrior - thought to have died in battle more than 2,000 years ago - could help archaeologists to pinpoint the site of an ancient holy site. The young warrior, aged about 30, with his spear, a sword, his belt and scabbard, stunned archaeologists who found his stone coffin.\\
Hypothesis: Altai ice maiden triggers major dispute.\\\\
Answer: No\\\\
Premise: Switzerland has ratified bilateral agreements with the members of the European Union in March 2004, but the new members (Cyprus , Czech Republic , Estonia, Hungary, Latvia, Lithuania, Malta, Poland, Slovakia and Slovenia) were not included in the deal.\\
Hypothesis: Lithuania intends to introduce the use of the Euro as an official currency on January 1, 2007.\\\\
Answer: No\\\\
Premise: Jill Pilgrim, general counsel of USA Track and Field, brought up the issue during a panel on women's sports at the sports lawyers conference. Pilgrim said the law regarding who is legally considered a woman is changing as sex-change operations become more common.\\
Hypothesis: Sex-change operations become more common.\\\\
Answer: Yes\\\\
Premise: Lastly, the author uses the precedent of marijuana legalization in other countries as evidence that legalization does not solve any social problems, but instead creates them.\\
Hypothesis: Drug legalization has benefits.\\\\
Answer: No\\\\
Premise: Dana Reeve, the widow of the actor Christopher Reeve, has died of lung cancer at age 44, according to the Christopher Reeve Foundation.\\
Hypothesis: Christopher Reeve had an accident.\\\\
Answer:}\\\\
\textbf{SST-2}\\
\texttt{Classify the review according to its sentiment.\\\\
Review: -- and especially williams , an american actress who becomes fully english \\\\
Sentiment: Positive\\\\
Review: each other so intensely , but with restraint\\\\
Sentiment: Positive\\\\
Review: tear your eyes away \\\\
Sentiment: Negative\\\\
Review: fascinating and playful \\\\
Sentiment: Positive\\\\
Review: been discovered , indulged in and rejected as boring before i see this piece of crap again \\\\
Sentiment: Negative\\\\
Review: it 's a charming and often affecting journey . \\\\
Sentiment:}\\\\
\textbf{SUBJ}\\
\texttt{Does the input contain personal opinions, feelings, or beliefs?\\\\
Input: by taking entertainment tonight subject matter and giving it humor and poignancy , auto focus becomes both gut-bustingly funny and crushingly depressing .\\\\
Output: Yes\\\\
Input: if you open yourself up to mr . reggio 's theory of this imagery as the movie 's set . . . it can impart an almost visceral sense of dislocation and change .\\\\
Output: Yes\\\\
Input: but for one celebrant this holy week is different .\\\\
Output: No\\\\
Input: with grit and determination molly guides the girls on an epic journey , one step ahead of the authorities , over 1 , 500 miles of australia 's outback in search of the rabbit-proof fence that bisects the continent and will lead them home .\\\\
Output: No\\\\
Input: just about all of the film is confusing on one level or another , making ararat far more demanding than it needs to be .\\\\
Output: Yes\\\\
Input: pirates of the caribbean is a sweeping action-adventure story set in an era when villainous pirates scavenged the caribbean seas .\\\\
Output:}\\\\
\textbf{WIC}\\
\texttt{You are given two sentences and a word. Tell me whether the word has the same meaning in both sentences.\\\\
Word: right\\
Sentence 1: He stood on the right.\\
Sentence 2: The pharmacy is just on the right past the bookshop.\\\\
Answer: Yes\\\\
Word: act\\
Sentence 1: She wants to act Lady Macbeth, but she is too young for the role.\\
Sentence 2: The dog acts ferocious, but he is really afraid of people.\\\\
Answer: No\\\\
Word: fall\\
Sentence 1: The hills around here fall towards the ocean.\\
Sentence 2: Her weight fell to under a hundred pounds.\\\\
Answer: No\\\\
Word: Round\\
Sentence 1: Round off the amount.\\
Sentence 2: The total is \$25,715 but to keep the figures simple, I'll round it down to \$25,000.\\\\
Answer: Yes\\\\
Word: have\\
Sentence 1: What do we have here?\\
Sentence 2: I have two years left.\\\\
Answer: No\\\\
Word: class\\
Sentence 1: An emerging professional class.\\
Sentence 2: Apologizing for losing your temper, even though you were badly provoked, showed real class.\\\\
Answer:}\\\\
\textbf{WSC}\\
\texttt{You are given a sentence, a prounoun and a noun. Tell me whether the specified pronoun and the noun phrase refer to the same entity in the sentence.\\\\
Sentence: Jane knocked on Susan 's door but she did not answer.\\
Pronoun: she\\
Noun: Susan\\\\
Answer: Yes\\\\
Sentence: The mothers of Arthur and Celeste have come to the town to fetch them. They are very happy to have them back, but they scold them just the same because they ran away.\\
Pronoun: they\\
Noun: Arthur and Celeste\\\\
Answer: No\\\\
Sentence: The scientists are studying three species of fish that have recently been found living in the Indian Ocean. They appeared two years ago.\\
Pronoun: They\\
Noun: The scientists\\\\
Answer: No\\\\
Sentence: Sergeant Holmes asked the girls to describe the intruder . Nancy not only provided the policeman with an excellent description of the heavyset thirty-year-old prowler, but drew a rough sketch of his face.\\
Pronoun: his\\
Noun: the intruder\\\\
Answer: Yes\\\\
Sentence: The boy continued to whip the pony , and eventually the pony threw him over. John laughed out quite loud. "Served him right," he said.\\
Pronoun: him\\
Noun: pony\\\\
Answer: No\\\\
Sentence: Bernard , who had not told the government official that he was less than 21 when he filed for a homestead claim, did not consider that he had done anything dishonest. Still, anyone who knew that he was 19 years old could take his claim away from him .\\
Pronoun: him\\
Noun: anyone\\\\
Answer:}

\newpage
\subsection{Mean Ablations: Full NLP ablation experiments}
\label{sec:NLP_Tables_Mean}
\begin{table*}[ht!]
    \centering
    \scriptsize
    \begin{tabularx}{0.55\textwidth}{ccccccc}
    \toprule
        \multicolumn{7}{c}{\textbf{Llama-3-8B}}\\
        \midrule
        Task & Shot & Full & \multicolumn{1}{c}{1\% ind.\vspace{-0.25pt}} & \multicolumn{1}{c}{1\% rnd.\vspace{-0.25pt}} & \multicolumn{1}{c}{3\% ind.\vspace{-0.25pt}} & \multicolumn{1}{c}{3\% rnd.\vspace{-0.25pt}}\\
 &  & model & \multicolumn{1}{c}{heads} & \multicolumn{1}{c}{heads} & \multicolumn{1}{c}{heads} & \multicolumn{1}{c}{heads} \\
    \midrule
BoolQ & 0 & 77.8 & 76.7 (-1.1) & 71.9 (-5.9) & 75.7 (-2.1) & 70.5 (-7.3)\\
      & 5 & 79.8 & 77.8 (-2.0) & 78.1 (-1.7) & 73.6 (-6.2) & 76.5 (-3.3) \\
      & 10 & 79.6 & 78.2 (-1.4) & 77.1 (-2.5) & 69.9 (-9.7) & 75.6 (-4.0)\\
\textit{SUL} & 10 & 46.3 & - & - & - & - \\
\midrule
ETHOS & 0 & 70.0 & 69.2 (-0.8) & 70.0  & 66.8 (-3.2) & 70.0 \\
      & 5 & 82.5 & 79.3 (-3.2) & 76.7 (-5.8) & 80.4 (-2.1) & 72.3 (-10.2) \\
      & 10 & 80.2 & 79.8 (-0.4) & 77.7 (-2.5) & 80.6 (+0.4) & 77.9 (-2.3)\\
\textit{SUL} & 10 & 70.3 & 51.4 (-18.9) & 62.3 (-8.0) & 56.5 (-13.8) & 63.0 (-7.3)\\
\midrule
RTE & 0 & 71.1 & 67.2 (-3.9) & 69.0 (-2.1) & 66.4 (-4.7) & 71.5 (+0.4)\\
    & 5 & 79.5 & 74.5 (-5.0) & 77.8 (-1.7) & 71.1 (-8.4) & 77.2 (-2.3) \\
    & 10 & 79.4 & 72.6 (-6.8) & 77.3 (-2.1) & 67.8 (-11.6) & 74.9 (-4.5)\\
\textit{SUL} & 10 & 52.4 & - & - & - & -\\
\midrule
SST-2 & 0 & 92.4 & 92.8 (+0.4) & 92.2 (-0.2) & 93.0 (+0.6) & 91.9 (-0.5)\\
      & 5 & 93.9 & 92.2 (-1.7) & 94.2 (+0.3) & 92.6 (-1.3) & 93.6 (-0.3) \\
      & 10 & 94.6 & 92.5 (-2.1) & 94.6  & 93.7 (-0.9) & 94.4 (-0.2)\\
\textit{SUL} & 10 & 94.0 & 70.7 (-23.3) & 93.0 (-1.0) & 70.5 (-23.5) & 92.5 (-1.5) \\
\midrule
SUBJ & 0 & 52.6 & 52.7 (+0.1) & 53.7 (+1.1) & 50.6 (-2.0) & 53.3 (+0.7)\\
     & 5 & 58.1 & 54.8 (-3.3) & 56.9 (-1.2) & 54.8 (-3.3) & 56.8 (-1.3) \\
     & 10 & 74.7 & 60.7 (-14.0) & 73.2 (-1.5) & 60.3 (-14.4) & 69.9 (-4.8)\\
\textit{SUL} & 10 & 69.8 & 49.7 (-20.1) & 65.2 (-4.6) & 52.8 (-17.0) & 65.3 (-4.5) \\
\midrule
WIC & 0 & 51.4 & 51.3 (-0.1) & 52.0 (+0.6) & 50.5 (-0.9) & 51.2 (-0.2)\\
    & 5 & 59.6 & 57.4 (-2.2) & 58.5 (-1.1) & 56.6 (-3.0) & 57.2 (-2.4) \\
    & 10 & 61.6 & 57.1 (-4.5) & 59.3 (-2.3) & 55.8 (-5.8) & 58.2 (-3.4)\\
\textit{SUL}& 10 & 52.6 & - & - & - & - \\
\end{tabularx}
    \centering
    \scriptsize
    \begin{tabularx}{0.55\textwidth}{ccccccc}
    \toprule
        \multicolumn{7}{c}{\textbf{InternLM2-20B}}\\
        \midrule
        Task & Shot & Full & \multicolumn{1}{c}{1\% ind.\vspace{-0.25pt}} & \multicolumn{1}{c}{1\% rnd.\vspace{-0.25pt}} & \multicolumn{1}{c}{3\% ind.\vspace{-0.25pt}} & \multicolumn{1}{c}{3\% rnd.\vspace{-0.25pt}}\\
 &  & model & \multicolumn{1}{c}{heads} & \multicolumn{1}{c}{heads} & \multicolumn{1}{c}{heads} & \multicolumn{1}{c}{heads} \\
    \midrule
BoolQ & 0 & 88.6 & - & - & - & - \\
      & 5 & 88.5 & - & - & - & -\\
      & 10 & 88.3 & - & - & - & - \\
\textit{SUL} & 10 & 80.7 & 70.1 (-10.6) & 79.5 (-1.2) & 63.6 (-17.1) & 78.8 (-1.9)\\
\midrule
ETHOS & 0 & 82.0 & 82.4 (+0.4) & 81.6 (-0.4) & 82.0  & 81.7 (-0.3)\\
      & 5 & 84.7 & 80.5 (-4.2) & 81.8 (-2.9) & 82.4 (-2.3) & 82.3 (-2.4)\\
      & 10 & 82.6 & 81.6 (-1.0) & 82.1 (-0.5) & 82.4 (-0.2) & 82.4 (-0.2)\\
\textit{SUL} & 10 & 82.5 & 68.3 (-14.2) & 76.7 (-5.8) & 68.6 (-13.9) & 76.4 (-6.1)\\
\midrule
RTE & 0 & 81.2 & 79.8 (-1.4) & 79.8 (-1.4) & 79.1 (-2.1) & 79.5 (-1.7)\\
     & 5 & 84.7 & 81.6 (-3.1) & 84.3 (-0.4) & 80.4 (-4.3) & 83.3 (-1.4)\\
     & 10 & 87.2 & 83.6 (-3.6) & 86.5 (-0.7) & 82.0 (-5.2) & 85.6 (-1.6)\\
    \textit{SUL} & 10 & 80.1 & 61.7 (-18.4) & 77.7 (-2.4) & 59.8 (-20.3) & 81.6 (+1.5)\\
\midrule
SST-2 & 0 & 96.0 & - & - & - & -\\
      & 5 & 95.1 & - & - & - & -\\
      & 10 & 95.3 & - & - & - & -\\
\textit{SUL} & 10 & 94.1 & 93.4 (-0.7) & 93.9 (-0.2) & 92.3 (-1.8) & 94.1 \\
\midrule
SUBJ & 0 & 61.1 & 61.6 (+0.5) & 60.3 (-0.8) & 61.5 (+0.4) & 59.8 (-1.3)\\
     & 5 & 75.4 & 69.2 (-6.2) & 74.8 (-0.6) & 70.5 (-4.9) & 74.1 (-1.3)\\
     & 10 & 81.4 & 73.5 (-7.9) & 81.2 (-0.2) & 70.5 (-10.9) & 78.8 (-2.6)\\
\textit{SUL} & 10 & 64.9 & 51.3 (-13.6) & 62.6 (-2.3) & 52.3 (-12.6) & 63.4 (-1.5)\\
\midrule
WIC & 0 & 60.8 & 61.0 (+0.2) & 61.1 (+0.3) & 59.7 (-1.1) & 61.1 (+0.3)\\
    & 5 & 68.6 & 67.4 (-1.2) & 69.1 (+0.5) & 62.0 (-6.6) & 67.9 (-0.7)\\
    & 10 & 68.9 & 66.5 (-2.4) & 68.1 (-0.8) & 63.5 (-5.4) & 67.8 (-1.1)\\
\textit{SUL}& 10 & 57.2 & 54.2 (-3.0) & 55.9 (-1.3) & 56.6 (-0.6) & 56.8 (-0.4)\\
\midrule
WSC & 0 & 70.2 & 67.3 (-2.9) & 68.9 (-1.3) & 66.4 (-3.8) & 72.1 (+1.9)\\
    & 5 & 73.1 & 67.6 (-5.5) & 72.7 (-0.4) & 69.9 (-3.2) & 73.2 (+0.1)\\
    & 10 & 75.3 & 69.6 (-5.7) & 71.6 (-3.7) & 68.6 (-6.7) & 71.7 (-3.6)\\
\textit{SUL} & 10 & 43.3 & - & - & - & -\\
    \bottomrule
\end{tabularx}
\caption{LLama-3-8B and InternLM2-20B ablation experiments on the NLP tasks. Columns labelled "1\% ind. heads" and "3\% ind. heads" show the results from fully ablating 1\% and 3\% of heads with the highest prefix scores, respectively. Columns "1\% rnd. heads" and "3\% rnd. heads" illustrate the effects of randomly ablating 1\% and 3\% of all heads in the model. The "SUL" row denotes settings using semantically unrelated labels. Performance differences due to the ablation, compared to the full model, are indicated in parentheses.}
\label{tab:NLP_Tables_Mean}
\end{table*}
\newpage

\subsection{Zero Ablations}
To investigate the significance of induction heads for a specific ICL task, we initially conducted zero-ablations of 1\% and 3\% of the heads with the highest prefix matching scores. This ablation process involved masking the corresponding partition of the output matrix, denoted as $\mathbf{W}_o^h$ in Eq. \ref{eq:attention}, by setting it to zero. This effectively renders the heads inactive and thereby prevents their contributions. As a control condition, we also randomly ablated 1\% and 3\% of the model's heads, which were selected from any layer except the final layer.
\subsubsection{Zero Ablations: Abstract Pattern Recognition Tasks}
\begin{table*}[ht!]
\centering
\footnotesize

\begin{tabularx}{\textwidth}{ccccccccc}
\toprule
\multicolumn{9}{c}{\textbf{Llama-3-8B}}\\
\midrule
Task & Shot & Full & \multicolumn{1}{c}{1\% ind.\vspace{-0.25pt}} & \multicolumn{1}{c}{1\% ind.\vspace{-0.25pt}} & \multicolumn{1}{c}{1\% rnd.\vspace{-0.25pt}} & \multicolumn{1}{c}{3\% ind.\vspace{-0.25pt}} & \multicolumn{1}{c}{3\% ind.\vspace{-0.25pt}} & \multicolumn{1}{c}{3\% rnd.\vspace{-0.25pt}}\\
 &  & model & \multicolumn{1}{c}{heads} & \multicolumn{1}{c}{pattern} & \multicolumn{1}{c}{heads} & \multicolumn{1}{c}{heads} & \multicolumn{1}{c}{pattern} & \multicolumn{1}{c}{heads} \\
\midrule
Repetition & 5 & 67.3 & 60.9 (-6.4) & 62.5 (-4.8) & 68.5 (+1.2) & 51.3 (-16.0) & 53.2 (-14.1) & 68.5 (+1.2)\\
           & 10 & 91.3 & 59.5 (-31.8) & 65.3 (-26.0) & 90.3 (-1.0) & 54.3 (-37.0) & 58.7 (-32.6) & 91.2 (-0.1)\\
\midrule
Recursion & 5 & 67.8 & 62.1 (-5.7) & 61.5 (-6.3) & 69.8 (+2.0) & 54.9 (-12.9) & 55.1 (-12.7) & 70.9 (+3.1)\\
          & 10 & 91.5 & 66.1 (-25.4) & 68.7 (-22.8) & 91.5 & 58.1 (-33.4) & 58.9 (-32.6) & 92.4 (+0.9)\\
\midrule
Centre-embedding & 5 & 58.8 & 54.0 (-4.8) & 55.0 (-3.8) & 59.2 (+0.4) & 48.3 (-10.5) & 50.5 (-8.3) & 59.7 (+0.9)\\
                 & 10 & 80.4 & 53.1 (-27.3) & 56.5 (-23.9) & 81.6 (+1.2) & 50.3 (-30.1) & 52.3 (-28.1) & 80.3 (-0.1)\\
\midrule
%Binary F\_a\_f\_p 
WordSeq 1 (binary)& 5 & 83.1 & 72.1 (-11.0) & 71.8 (-11.3) & 80.6 (-2.5) & 54.7 (-28.4) & 56.9 (-26.2) & 81.0 (-3.1)\\
                 & 10 & 99.4 & 96.4 (-3.0) & 97.3 (-2.1) & 98.4 (-1.0) & 79.2 (-20.2) & 82.2 (-17.2) & 97.9 (-1.5)\\
\midrule
%Binary V\_v\_b\_i
WordSeq 2 (binary)& 5 & 77.9 & 66.0 (-11.9) & 65.2 (-12.7) & 74.4 (-3.5) & 53.5 (-24.4) & 55.7 (-22.2) & 75.1 (-2.8)\\
                 & 10 & 99.4 & 95.3 (-4.1) & 96.4 (-3.0) & 97.3 (-2.1) & 77.2 (-22.2) & 81.1 (-18.3) & 97.6 (-1.8)\\
\midrule
%4-way F\_a\_f\_p 
WordSeq 1 (4-way)& 20 & 78.3 & 59.7 (-18.6) & 59.8 (-18.5) & 71.9 (-6.4) & 44.8 (-33.5) & 45.2 (-33.1) & 72.0 (-6.3)\\
\midrule
%4-way V\_v\_b\_i
WordSeq 2 (4-way)& 20 & 81.3 & 58.6 (-22.7) & 59.8 (-21.5) & 71.1 (-10.2) & 46.5 (-34.8) & 47.5 (-33.8) & 71.5 (-9.8)\\
\toprule
\end{tabularx}
\footnotesize
\textbf{InternLM2-20B}
\begin{tabularx}{\textwidth}{cccccccccc}
\midrule
Task & Shot & Full & \multicolumn{1}{c}{1\% ind.\vspace{-0.25pt}} & \multicolumn{1}{c}{1\% ind.\vspace{-0.25pt}} & \multicolumn{1}{c}{1\% rnd.\vspace{-0.25pt}} & \multicolumn{1}{c}{3\% ind.\vspace{-0.25pt}} & \multicolumn{1}{c}{3\% ind.\vspace{-0.25pt}} & \multicolumn{1}{c}{3\% rnd.\vspace{-0.25pt}}\\
 &  & model & \multicolumn{1}{c}{heads} & \multicolumn{1}{c}{pattern} & \multicolumn{1}{c}{heads} & \multicolumn{1}{c}{heads} & \multicolumn{1}{c}{pattern} & \multicolumn{1}{c}{heads} \\
\midrule
Repetition & 5 & 68.7 & 63.3 (-5.4) & 62.0 (-6.7) & 68.2 (-0.5) & 60.8 (-7.9) & 58.5 (-10.2) & 64.0 (-4.7)\\
& 10 & 88.1 & 73.2 (-14.9) & 72.1 (-16.0) & 87.8 (-0.3) & 68.7 (-19.4) & 61.2 (-26.9) & 84.1 (-4.0)\\
\midrule
Recursion & 5 & 68.1 & 61.4 (-6.7) & 59.9 (-8.2) & 68.2 (+0.1) & 59.6 (-8.5) & 55.3 (-12.8) & 64.4 (-3.7)\\
& 10 & 88.5 & 70.6 (-17.9) & 69.9 (-18.6) & 87.7 (-0.8) & 66.1 (-22.4) & 57.1 (-31.4) & 84.4 (-4.1)\\
\midrule
Centre-embedding & 5 & 59.5 & 56.3 (-3.2) & 56.1 (-3.4) & 58.9 (-0.6) & 55.1 (-4.4) & 54.1 (-5.4) & 56.3 (-3.2)\\
& 10 & 75.3 & 60.3 (-15.0) & 58.5 (-16.8) & 74.9 (-0.4) & 56.9 (-18.4) & 54.3 (-21.0) & 71.2 (-4.1)\\
\midrule
WordSeq 1 (binary) & 5 & 76.8 & 67.4 (-9.4) & 64.9 (-11.9) & 76.7 (-0.1) & 59.9 (-16.9) & 56.8 (-20.0) & 76.9 (+0.1)\\
& 10 & 95.6 & 91.7 (-3.9) & 88.2 (-7.4) & 94.8 (-0.8) & 85.7 (-9.9) & 83.3 (-12.3) & 95.6 \\
\midrule
WordSeq 2 (binary) & 5 & 83.3 & 74.8 (-8.5) & 70.1 (-13.2) & 82.5 (-0.8) & 68.6 (-14.7) & 64.1 (-19.2) & 79.9 (-3.4)\\
& 10 & 99.0 & 98.9 (-0.1) & 96.8 (-2.2) & 96.2 (-2.8) & 94.1 (-4.9) & 86.7 (-12.3) & 98.6 (-0.4)\\
\midrule
WordSeq 1 (4-way)& 20 & 77.3 & 68.8 (-8.5) & 65.4 (-11.9) & 76.0 (-1.3) & 37.3 (-40.0) & 37.4 (-39.9) & 75.5 (-1.8)\\
\midrule
WordSeq 2 (4-way) & 20 & 80.4 & 78.4 (-2.0) & 75.5 (-4.9) & 79.3 (-1.1) & 52.9 (-27.5) & 47.5 (-32.9) & 77.3 (-3.1)\\

\bottomrule
\end{tabularx}
\caption{Llama-3-8B (top) and InternLM2-20B (bottom) ablation experiments on the abstract pattern recognition tasks. For both models, zero-shot performance of the full model is \textasciitilde$50\%$ in all tasks. Columns labelled "1\% ind. heads" and "3\% ind. heads" show results from fully ablating 1\% and 3\% of heads with the highest prefix scores, respectively. "1\% ind. pattern" and "3\% ind. pattern" columns depict outcomes from blocking induction attention patterns in 1\% and 3\% of these heads (Sec. \ref{sec:Knockout}). Columns "1\% rnd. heads" and "3\% rnd. heads" illustrate the effects of randomly ablating 1\% and 3\% of all heads in the model. Performance differences due to the ablation, compared to the full model, are indicated in parentheses.}
\label{tab:Abstr_Results}
\end{table*}
\newpage

\subsubsection{Zero Ablations: Full NLP ablation experiments}
\label{sec:NLP_Tables}

\begin{table*}[ht!]
    \centering
    \scriptsize
    \begin{tabularx}{0.55\textwidth}{ccccccc}
    \toprule
        \multicolumn{7}{c}{\textbf{Llama-3-8B}}\\
        \midrule
        Task & Shot & Full & \multicolumn{1}{c}{1\% ind.\vspace{-0.25pt}} & \multicolumn{1}{c}{1\% rnd.\vspace{-0.25pt}} & \multicolumn{1}{c}{3\% ind.\vspace{-0.25pt}} & \multicolumn{1}{c}{3\% rnd.\vspace{-0.25pt}}\\
 &  & model & \multicolumn{1}{c}{heads} & \multicolumn{1}{c}{heads} & \multicolumn{1}{c}{heads} & \multicolumn{1}{c}{heads} \\
    \midrule
    BoolQ & 0 & 77.8 & 76.2 (-1.6) & 72.7 (-5.1) & 75.2 (-2.6) & 71.9 (-5.9)\\
    & 5 & 79.8 & 77.1 (-2.7) & 76.4 (-3.4) & 73.0 (-6.8) & 77.2 (-2.6)\\
    & 10 & 79.6 & 77.7 (-1.9) & 77.1 (-2.5) & 70.7 (-8.9) & 77.3 (-2.3)\\
    \textit{SUL} & 10 & 46.3 & - & - & - & - \\
    \midrule
    ETHOS & 0 & 70.0 & 72.0 (+2.0) & 68.6 (-1.4) & 70.0  & 62.0 (-8.0)\\
    & 5 & 82.5 & 82.5  & 79.5 (-3.0) & 81.2 (-1.3) & 75.3 (-7.2)\\
    & 10 & 80.2 & 80.2 & 80.9 (+0.7) & 81.1 (-1.1) & 77.5 (-2.7)\\
    \textit{SUL} & 10 & 70.3 & 52.8 (-17.5) & 66.8 (-3.5) & 57.9 (-12.4) & 63.8 (-6.5)\\
    \midrule
    RTE & 0 & 71.1 & 67.5 (-3.6) & 70.3 (-0.8) & 67.9 (-3.2) & 67.5 (-3.6)\\
    & 5 & 79.5 & 75.2 (-4.3) & 79.0 (-0.5) & 71.6 (-7.9) & 78.1 (-1.4)\\
    & 10 & 79.4 & 74.2 (-5.2) & 78.7 (-0.7)& 68.6 (-10.8)& 78.4 (-1.0)\\
    \textit{SUL} & 10 & 52.4 & - & - & - & -\\
    \midrule
    SST-2 & 0 & 92.4 & 92.1 (-0.3) & 92.7 (+0.3) & 92.2 (-0.2) & 91.9 (-0.5)\\
    & 5 & 93.9 & 92.4 (-1.5) & 94.3 (+0.4) & 93.0 (-0.9) & 93.9 \\
    & 10 & 94.6 & 92.9 (-1.7) & 94.7 (+0.1) & 93.2 (-1.4) & 94.5 (-0.1)\\
    \textit{SUL} & 10 & 94.0 & 72.8 (-21.2) & 93.6 (-0.4) & 71.8 (-22.2) & 92.6 (-1.4)\\
    \midrule
    SUBJ & 0 & 52.6 & 53.0 (+0.4) & 54.0 (+1.4) & 52.4 (-0.2) & 53.2 (+0.6)\\
    & 5 & 58.1 & 53.5 (-4.6) & 58.8 (+0.7) & 56.3 (-1.8) & 56.5 (-1.6)\\
    & 10 & 74.7 & 62.4 (-12.3) & 74.6 (-0.1) & 62.6 (-12.1) & 71.9 (-2.8)\\
    \textit{SUL} & 10 & 69.8 & 49.9 (-19.9) & 67.0 (-2.8) & 53.1 (-16.7) & 70.4 (+0.6)\\
    \midrule
    WIC & 0 & 51.4 & 51.4  & 51.0 (-0.4) & 50.9 (-0.5) & 51.7 (+0.3)\\
    & 5 & 59.6 & 58.2 (-1.4) & 59.0 (-0.6) & 56.3 (-3.3) & 56.8 (-2.8)\\
    & 10 & 61.6 & 56.8 (-4.8) & 59.6 (-2.0) & 56.2 (-5.4) & 57.0 (-4.6)\\
    \textit{SUL}& 10 & 52.6 & - & - & - & - \\
\end{tabularx}
    \centering
    \scriptsize
    \begin{tabularx}{0.55\textwidth}{ccccccc}
    \toprule
        \multicolumn{7}{c}{\textbf{InternLM2-20B}}\\
        \midrule
        Task & Shot & Full & \multicolumn{1}{c}{1\% ind.\vspace{-0.25pt}} & \multicolumn{1}{c}{1\% rnd.\vspace{-0.25pt}} & \multicolumn{1}{c}{3\% ind.\vspace{-0.25pt}} & \multicolumn{1}{c}{3\% rnd.\vspace{-0.25pt}}\\
 &  & model & \multicolumn{1}{c}{heads} & \multicolumn{1}{c}{heads} & \multicolumn{1}{c}{heads} & \multicolumn{1}{c}{heads} \\
    \midrule
    BoolQ & 0 & 88.6 & 88.6 & 88.5 (-0.1) & 87.7 (-0.9) & 86.9 (-1.7)\\
    & 5 & 88.5 & 88.5 & 88.6 (+0.1) & 88.2 (-0.3) & 87.5 (-1.0)\\
    & 10 & 88.3 & 88.3 & 88.2 (-0.1) & 87.9 (-0.4) & 87.1 (-1.3)\\
    \textit{SUL} & 10 & 80.7 & 72.4 (-8.3) & 80.9 (+0.2) & 64.9 (-15.8) & 77.0 (-3.7)\\
    \midrule
    ETHOS & 0 & 82.0 & 82.5 (+0.5) & 82.5 (+0.5) & 82.0  & 81.5 (-0.5)\\
    & 5 & 84.7 & 81.3 (-3.4) & 84.2 (-0.5) & 81.8 (-2.9) & 82.6 (-2.1)\\
    & 10 & 82.6 & 81.0 (-1.6) & 82.9 (+0.3) & 82.1 (-0.5) & 82.8 (+0.2)\\
    \textit{SUL} & 10 & 82.5 & 74.8 (-7.7) & 81.3 (-1.2) & 68.2 (-14.3) & 74.4 (-8.1)\\
    \midrule
    RTE & 0 & 81.2 & 80.1 (-1.1) & 80.6 (-0.6) & 80.5 (-0.7) & 79.9 (-1.3)\\
    & 5 & 84.7 & 82.3 (-2.4) & 84.4 (-0.3) & 81.4 (-3.3) & 84.4 (-0.3)\\
    & 10 & 87.2 & 84.0 (-3.2) & 86.8 (-0.4) & 82.3 (-4.9) & 86.4 (-0.8) \\
    \textit{SUL} & 10 & 80.1 & 62.9 (-17.2) & 81.4 (+1.3) & 56.4 (-23.7) & 76.9 (-3.2)\\
    \midrule
    SST-2 & 0 & 96.0 & 96.0 & 95.7 (-0.3) & 95.6 (-0.4) & 95.7 (-0.3)\\
    & 5 & 95.1 & 95.1 & 95.0 (-0.1) & 94.2 (-0.9) & 94.9 (-0.2)\\
    & 10 & 95.3 & 95.1 (-0.2) & 95.2 (-0.1) & 94.3 (-1.0) & 95.2 (-0.1)\\
    \textit{SUL} & 10 & 94.1 & 93.6 (-0.5) & 94.1 & 92.2 (-1.9) & 93.6 (-0.5)\\
    \midrule
    SUBJ & 0 & 61.1 & 61.7 (+0.6) & 61.6 (+0.5) & 62.6 (+1.5) & 59.9 (-1.2)\\
    & 5 & 75.4 & 69.6 (-5.8) & 75.0 (-0.4) & 70.0 (-5.4) & 73.4 (-2.0)\\
    & 10 & 81.4 & 75.1 (-6.3) & 81.4 & 71.3 (-10.1) & 78.8 (-2.6)\\
    \textit{SUL} & 10 & 64.9 & 53.4 (-11.5) & 65.1 (+0.2) & 52.2 (-12.7) & 61.3 (-3.6)\\
    \midrule
    WIC & 0 & 60.8 & 60.7 (-0.2) & 60.8  & 59.9 (-0.9) & 60.1 (-0.7)\\
    & 5 & 68.6 & 67.2 (-1.4) & 68.7 (+0.1) & 61.9 (-6.7) & 68.9 (+0.3)\\
    & 10 & 68.9 & 67.0 (-1.9) & 68.6 (-0.3) & 63.5 (-5.4) & 68.3 (-0.6)\\
    \textit{SUL}& 10 & 57.2 & 54.8 (-2.4) & 57.7 (+0.5) & 56.6 (-0.6) & 54.9 (-2.3)\\
    \midrule
    WSC & 0 & 70.2 & 67.3 (-2.9) & 69.6 (-0.6) & 68.3 (-1.9) & 70.8 (+0.6)\\
    & 5 & 73.1 & 68.0 (-5.1) & 73.0 (-0.1) & 69.6 (-3.5) & 73.1 \\
    & 10 & 75.3 & 69.9 (-5.4) & 75.2 (-0.1) & 68.6 (-1.3) & 71.9 (-3.4)\\
    \textit{SUL} & 10 & 43.3 & - & - & - & -\\
    \bottomrule
\end{tabularx}
\caption{LLama-3-8B and InternLM2-20B ablation experiments on the NLP tasks. Columns labelled "1\% ind. heads" and "3\% ind. heads" show the results from fully ablating 1\% and 3\% of heads with the highest prefix scores, respectively. Columns "1\% rnd. heads" and "3\% rnd. heads" illustrate the effects of randomly ablating 1\% and 3\% of all heads in the model. The "SUL" row denotes settings using semantically unrelated labels. Performance differences due to the ablation, compared to the full model, are indicated in parentheses.}
\label{tab:NLP_Table_Zero}
\end{table*}

\newpage
\subsubsection{Zero Ablations: NLP Figures}
\label{sec:NLP_Heatmaps_Zero}
\begin{figure}[h!]
    \centering
    \includegraphics[scale=0.4]{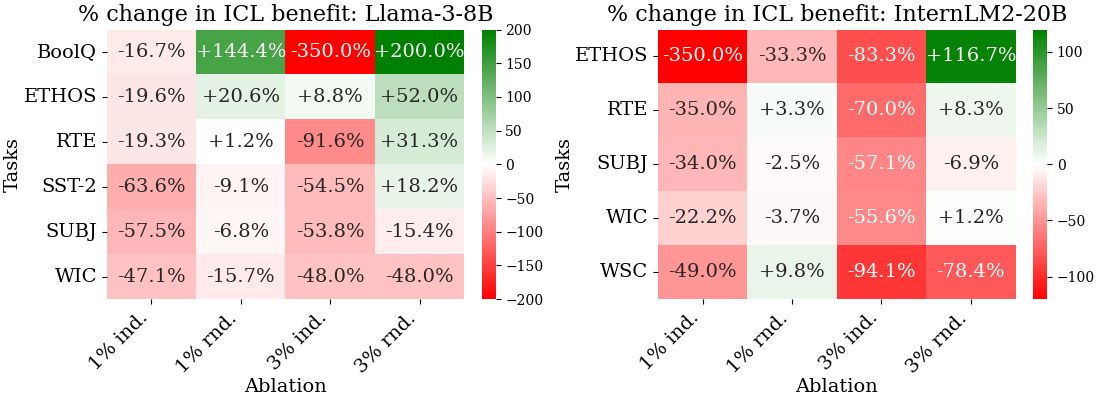}
    \caption{Change in ICL benefit for Llama-3-8B (left) and InternLM2-20B (right), due to head ablations when compared to that of the full model. "1\% ind." and "3\% ind." denote ablating the top respective percentage of induction heads. "1\% rnd." and "3\% rnd." denote randomly ablating the respective percentage of all heads in the model.}
    \label{fig:NLP_Zero}
\end{figure}
\begin{figure}[h!]
    \centering
    \footnotesize
    \textbf{SUL Ablation Experiments: Llama-3-8B}
    \includegraphics[scale=0.4]{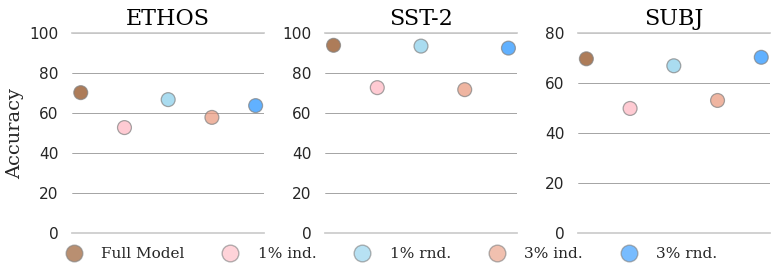}
    \newline
    \newline
    \footnotesize
    \textbf{SUL Ablation Experiments: InternLM2-20B}\\
    \includegraphics[scale=0.4]{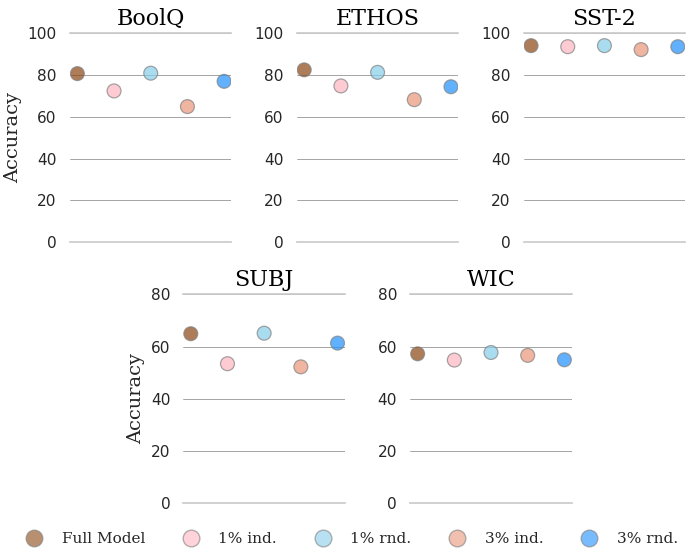}
    \caption{SUL ablation experiments for Llama-3-8B (top) and InternLM2-20B (bottom) on NLP tasks. "1\% ind." and "3\% ind." denote ablating the top respective percentage of induction heads. "1\% rnd." and "3\% rnd." denote randomly ablating the respective percentage of all heads in the model.}
    \label{fig:SUL_Zero}
\end{figure}
\newpage

\subsection{Attention Patterns}
\label{sec:att_pattern}
\subsubsection{WordSeq1 (Binary)}

\begin{figure*}[h!]
\centering
\begin{minipage}{0.48\textwidth}
\centering
  \includegraphics[scale=0.35]{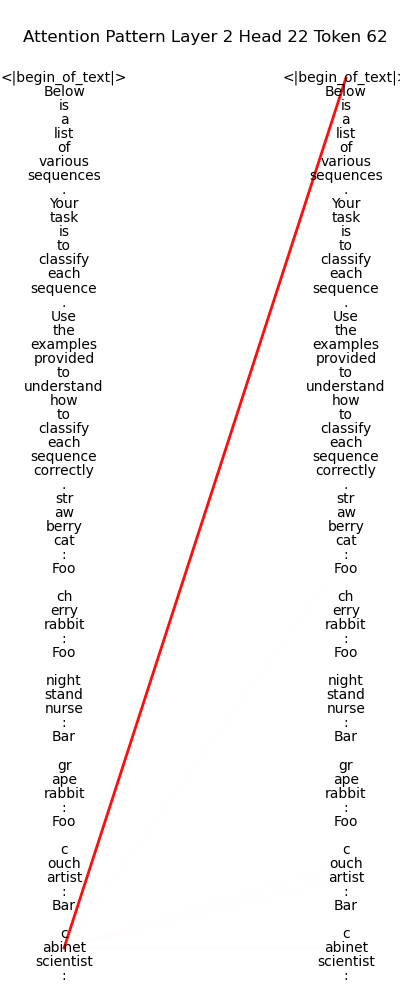}
  \caption{Attention pattern of token 62 in head 22 of layer 2 in Llama-3-8B. Example 1/5.}
\end{minipage}\hfill
\begin{minipage}{0.48\textwidth}
\centering
  \includegraphics[scale=0.35]{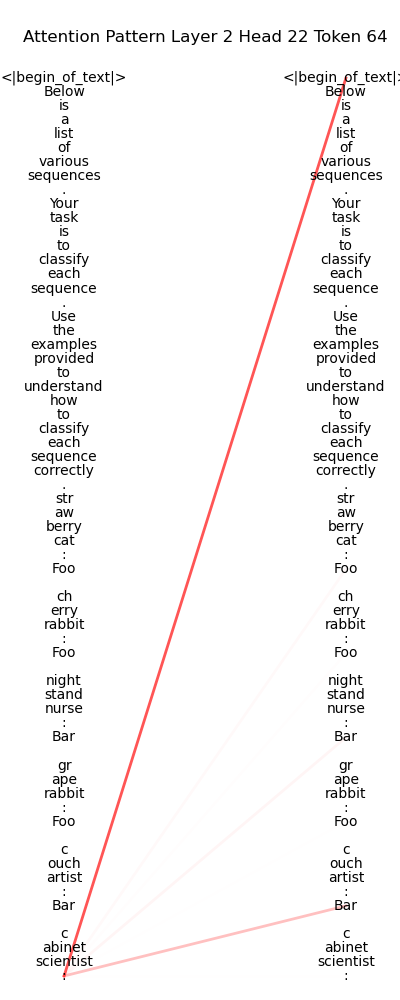}
  \caption{Attention pattern of token 64 in head 22 of layer 2 in Llama-3-8B. Plot 1/5.}
\end{minipage}
\end{figure*}

\begin{figure*}[h!]
\centering
\begin{minipage}{0.48\textwidth}
\centering
  \includegraphics[scale=0.35]{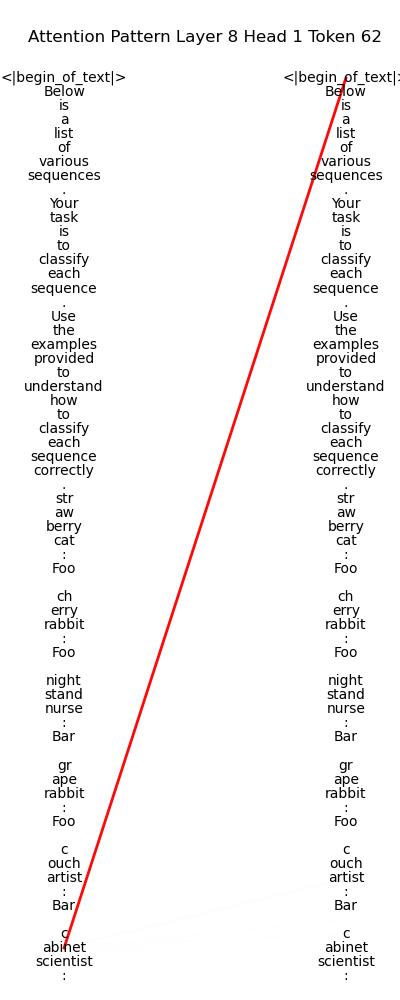}
  \caption{Attention pattern of token 62 in head 1 of layer 8 in Llama-3-8B. Example 1/5.}
\end{minipage}\hfill
\begin{minipage}{0.48\textwidth}
\centering
  \includegraphics[scale=0.35]{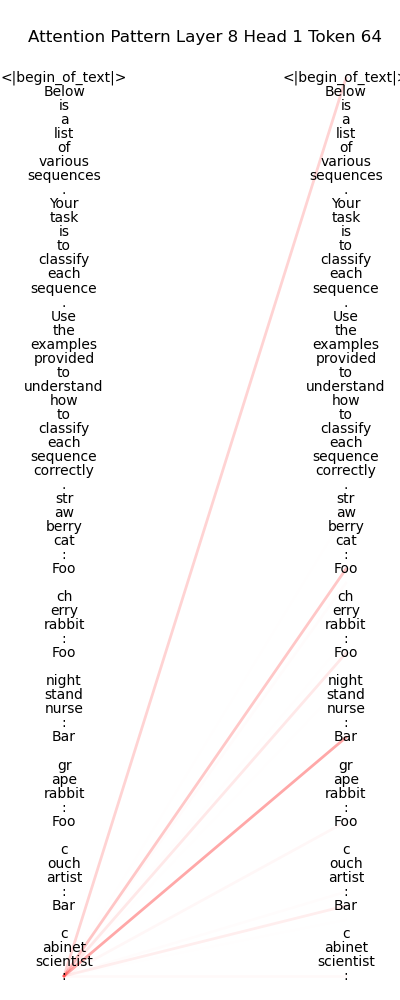}
  \caption{Attention pattern of token 64 in head 1 of layer 8 in Llama-3-8B. Plot 1/5.}
\end{minipage}
\end{figure*}

\begin{figure*}[h!]
\centering
\begin{minipage}{0.48\textwidth}
\centering
  \includegraphics[scale=0.35]{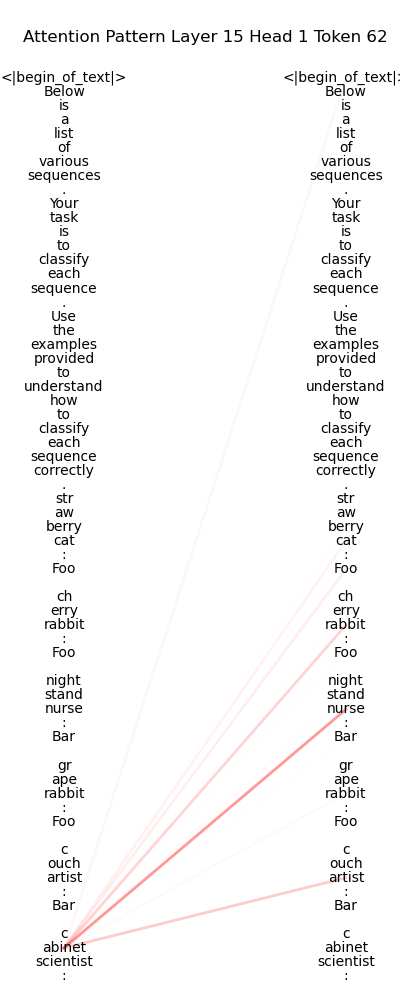}
  \caption{Attention pattern of token 62 in head 1 of layer 15 in Llama-3-8B. Example 1/5.}
\end{minipage}\hfill
\begin{minipage}{0.48\textwidth}
\centering
  \includegraphics[scale=0.35]{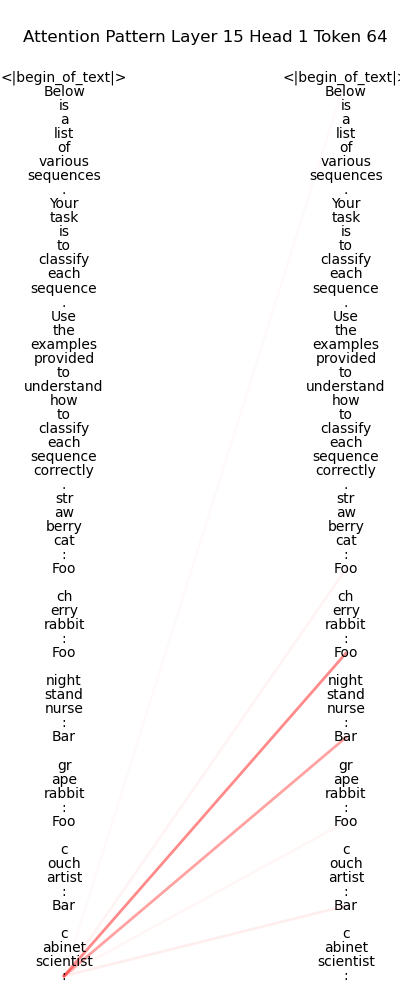}
  \caption{Attention pattern of token 64 in head 1 of layer 15 in Llama-3-8B. Plot 1/5.}
\end{minipage}
\end{figure*}

\begin{figure*}[h!]
\centering
\begin{minipage}{0.48\textwidth}
\centering
  \includegraphics[scale=0.35]{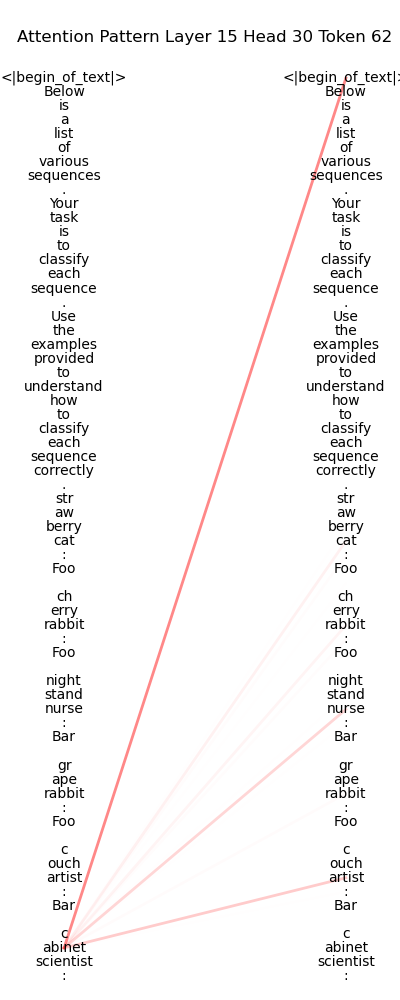}
  \caption{Attention pattern of token 62 in head 30 of layer 15 in Llama-3-8B. Example 1/5.}
\end{minipage}\hfill
\begin{minipage}{0.48\textwidth}
\centering
  \includegraphics[scale=0.35]{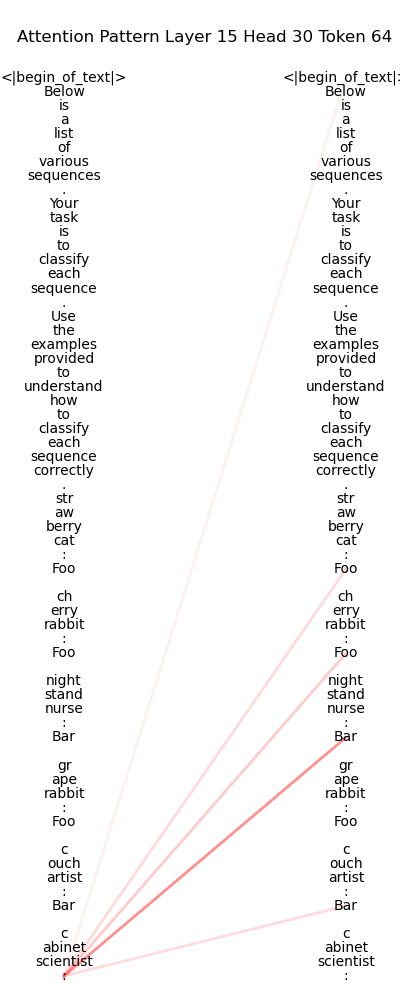}
  \caption{Attention pattern of token 64 in head 30 of layer 15 in Llama-3-8B. Plot 1/5.}
\end{minipage}
\end{figure*}

\begin{figure*}[h!]
\centering
\begin{minipage}{0.48\textwidth}
\centering
  \includegraphics[scale=0.35]{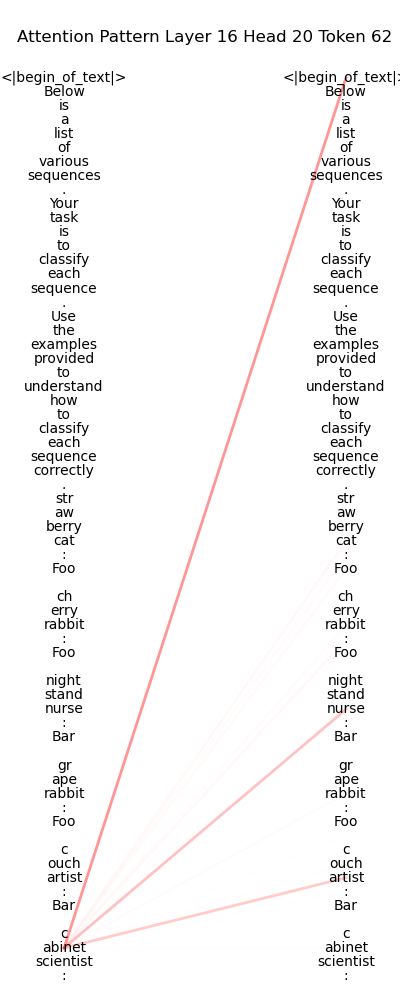}
  \caption{Attention pattern of token 62 in head 20 of layer 16 in Llama-3-8B. Example 1/5.}
\end{minipage}\hfill
\begin{minipage}{0.48\textwidth}
\centering
  \includegraphics[scale=0.35]{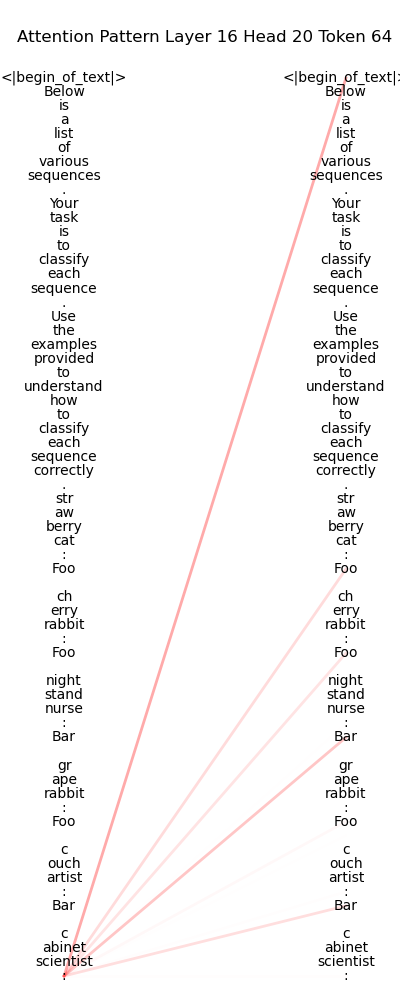}
  \caption{Attention pattern of token 64 in head 20 of layer 16 in Llama-3-8B. Plot 1/5.}
\end{minipage}
\end{figure*}

\begin{figure*}[h!]
\centering
\begin{minipage}{0.48\textwidth}
\centering
  \includegraphics[scale=0.35]{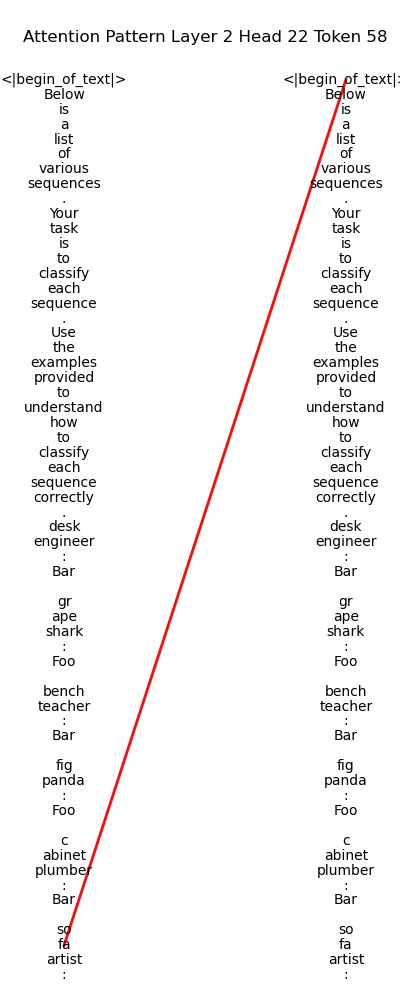}
  \caption{Attention pattern of token 58 in head 22 of layer 2 in Llama-3-8B. Example 2/5.}
\end{minipage}\hfill
\begin{minipage}{0.48\textwidth}
\centering
  \includegraphics[scale=0.35]{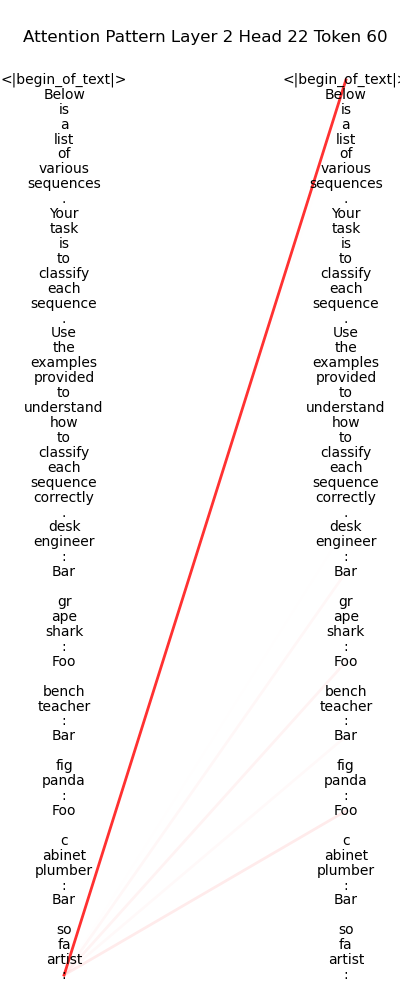}
  \caption{Attention pattern of token 60 in head 22 of layer 2 in Llama-3-8B. Plot 2/5.}
\end{minipage}
\end{figure*}

\begin{figure*}[h!]
\centering
\begin{minipage}{0.48\textwidth}
\centering
  \includegraphics[scale=0.35]{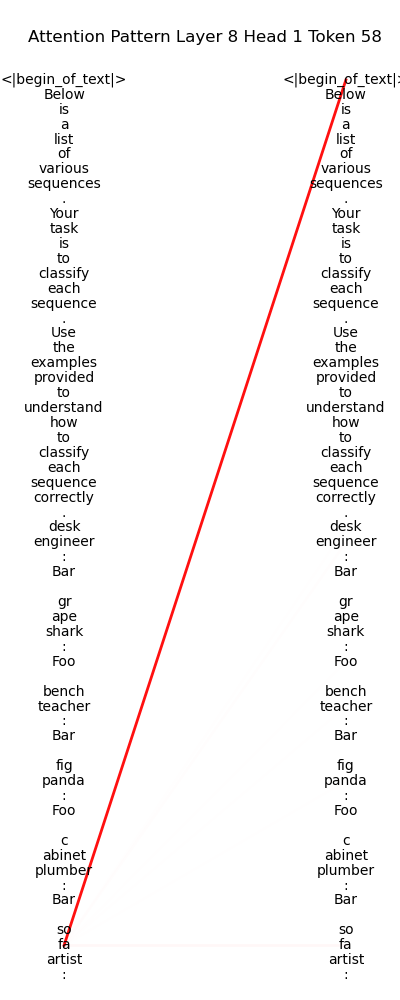}
  \caption{Attention pattern of token 58 in head 1 of layer 8 in Llama-3-8B. Example 2/5.}
\end{minipage}\hfill
\begin{minipage}{0.48\textwidth}
\centering
  \includegraphics[scale=0.35]{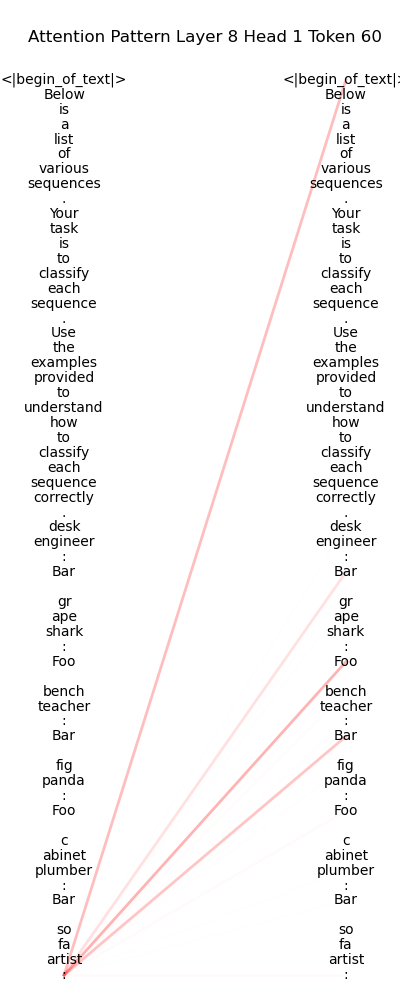}
  \caption{Attention pattern of token 60 in head 1 of layer 8 in Llama-3-8B. Plot 2/5.}
\end{minipage}
\end{figure*}

\begin{figure*}[h!]
\centering
\begin{minipage}{0.48\textwidth}
\centering
  \includegraphics[scale=0.35]{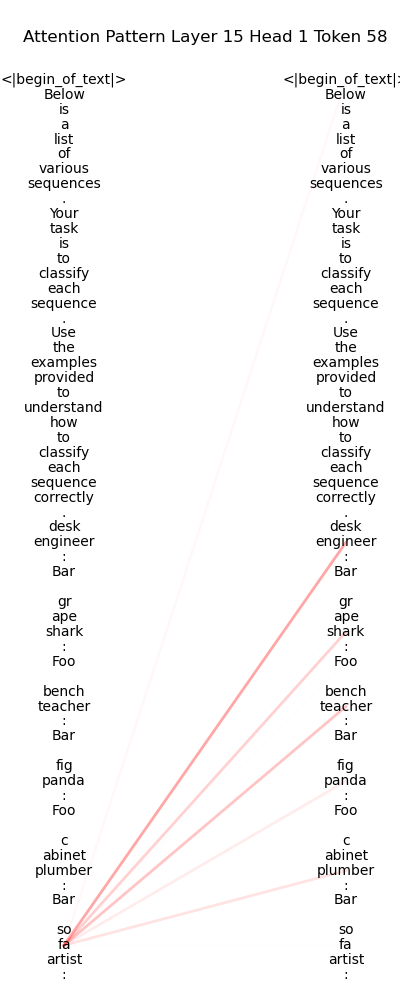}
  \caption{Attention pattern of token 58 in head 1 of layer 15 in Llama-3-8B. Example 2/5.}
\end{minipage}\hfill
\begin{minipage}{0.48\textwidth}
\centering
  \includegraphics[scale=0.35]{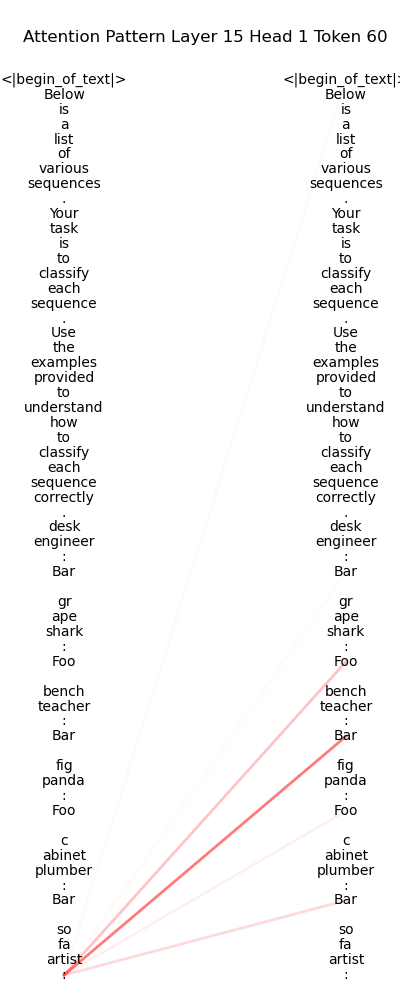}
  \caption{Attention pattern of token 60 in head 1 of layer 15 in Llama-3-8B. Plot 2/5.}
\end{minipage}
\end{figure*}

\begin{figure*}[h!]
\centering
\begin{minipage}{0.48\textwidth}
\centering
  \includegraphics[scale=0.35]{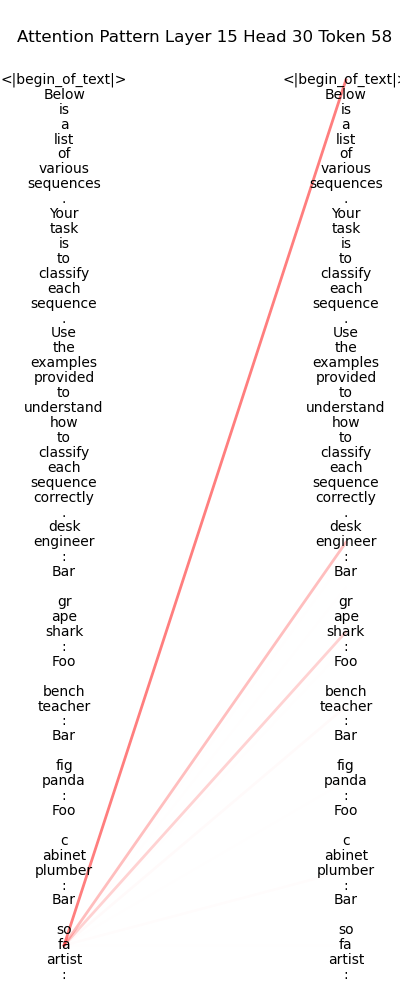}
  \caption{Attention pattern of token 58 in head 30 of layer 15 in Llama-3-8B. Example 2/5.}
\end{minipage}\hfill
\begin{minipage}{0.48\textwidth}
\centering
  \includegraphics[scale=0.35]{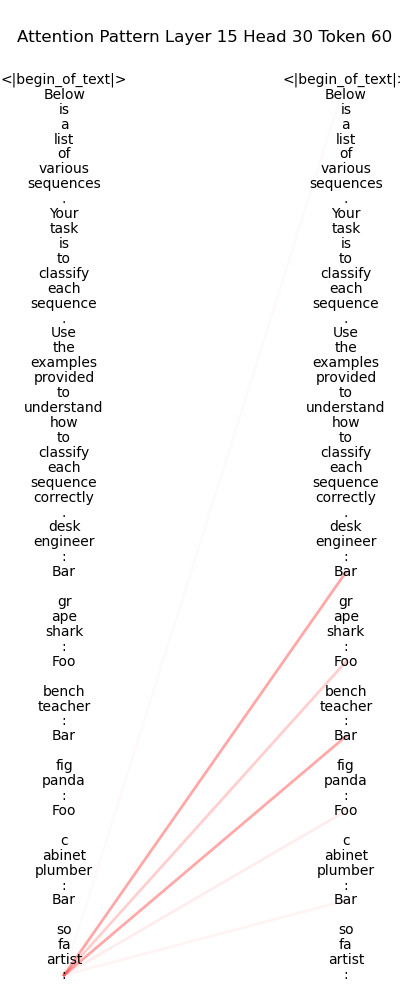}
  \caption{Attention pattern of token 60 in head 30 of layer 15 in Llama-3-8B. Plot 2/5.}
\end{minipage}
\end{figure*}

\begin{figure*}[h!]
\centering
\begin{minipage}{0.48\textwidth}
\centering
  \includegraphics[scale=0.35]{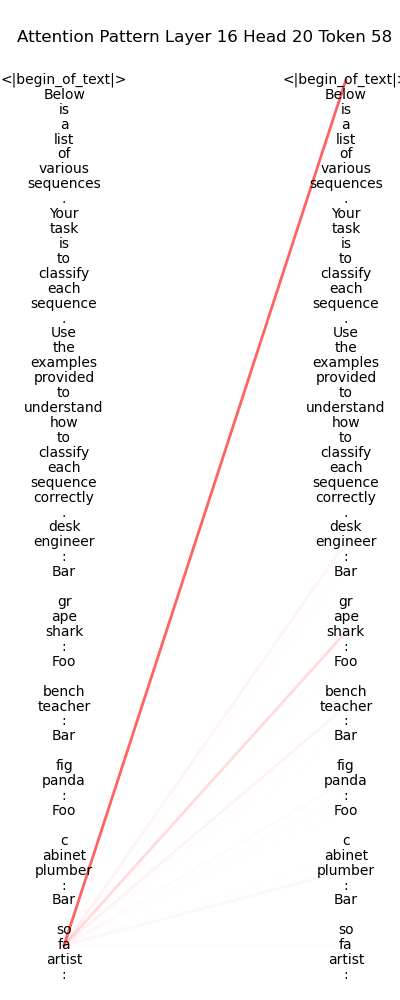}
  \caption{Attention pattern of token 58 in head 20 of layer 16 in Llama-3-8B. Example 2/5.}
\end{minipage}\hfill
\begin{minipage}{0.48\textwidth}
\centering
  \includegraphics[scale=0.35]{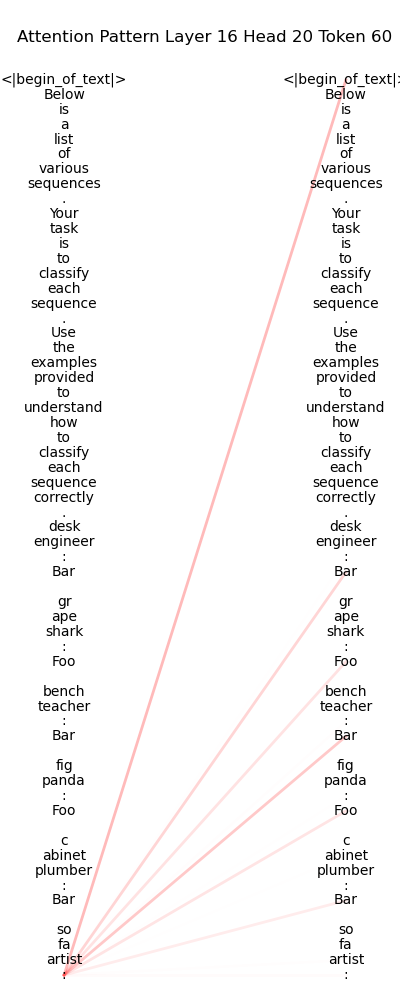}
  \caption{Attention pattern of token 60 in head 20 of layer 16 in Llama-3-8B. Plot 2/5.}
\end{minipage}
\end{figure*}

\begin{figure*}[h!]
\centering
\begin{minipage}{0.48\textwidth}
\centering
  \includegraphics[scale=0.35]{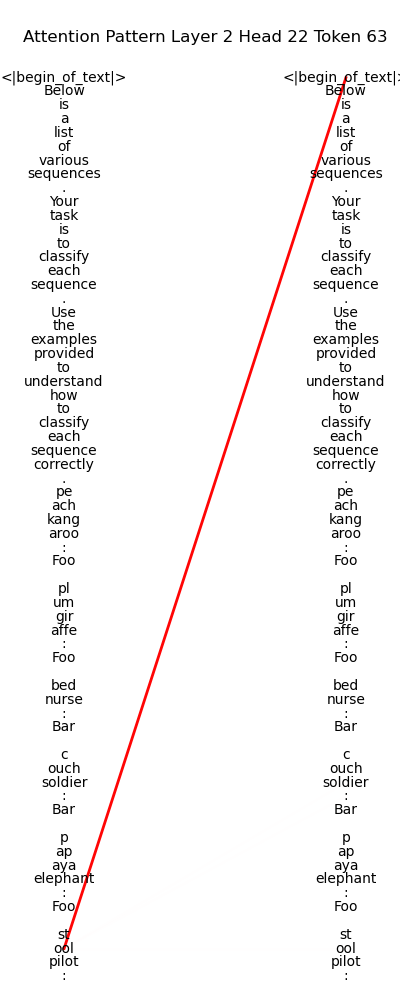}
  \caption{Attention pattern of token 63 in head 22 of layer 2 in Llama-3-8B. Example 3/5.}
\end{minipage}\hfill
\begin{minipage}{0.48\textwidth}
\centering
  \includegraphics[scale=0.35]{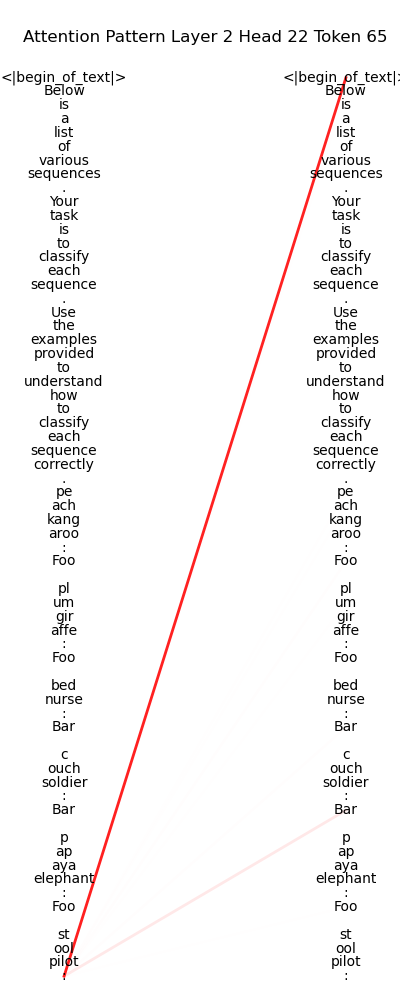}
  \caption{Attention pattern of token 65 in head 22 of layer 2 in Llama-3-8B. Plot 3/5.}
\end{minipage}
\end{figure*}

\begin{figure*}[h!]
\centering
\begin{minipage}{0.48\textwidth}
\centering
  \includegraphics[scale=0.35]{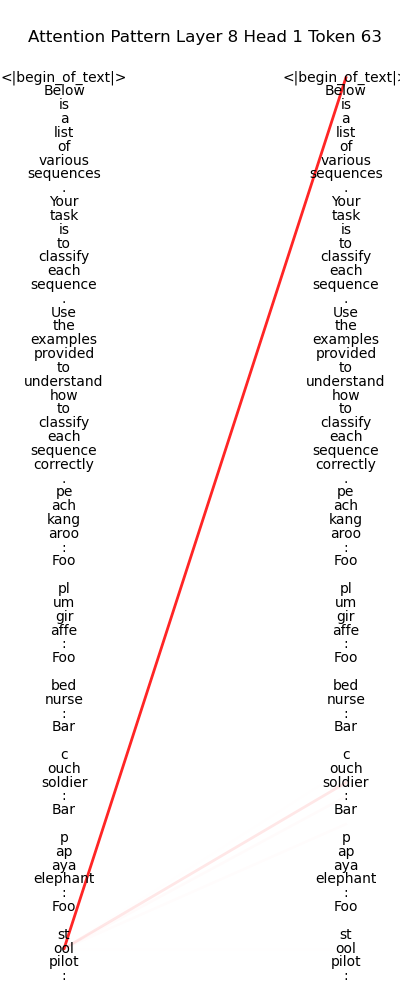}
  \caption{Attention pattern of token 63 in head 1 of layer 8 in Llama-3-8B. Example 3/5.}
\end{minipage}\hfill
\begin{minipage}{0.48\textwidth}
\centering
  \includegraphics[scale=0.35]{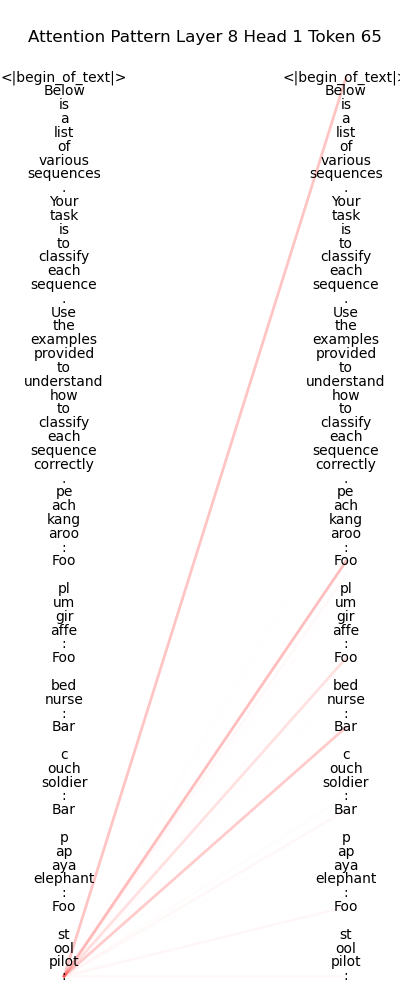}
  \caption{Attention pattern of token 65 in head 1 of layer 8 in Llama-3-8B. Plot 3/5.}
\end{minipage}
\end{figure*}

\begin{figure*}[h!]
\centering
\begin{minipage}{0.48\textwidth}
\centering
  \includegraphics[scale=0.35]{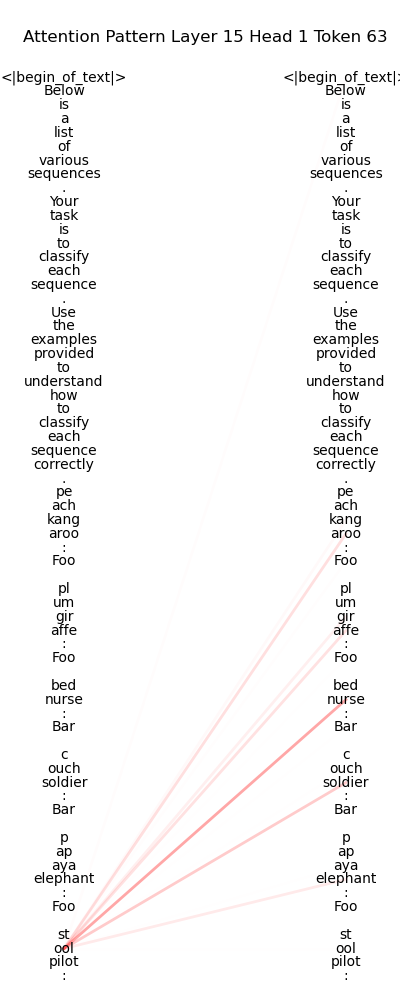}
  \caption{Attention pattern of token 63 in head 1 of layer 15 in Llama-3-8B. Example 3/5.}
\end{minipage}\hfill
\begin{minipage}{0.48\textwidth}
\centering
  \includegraphics[scale=0.35]{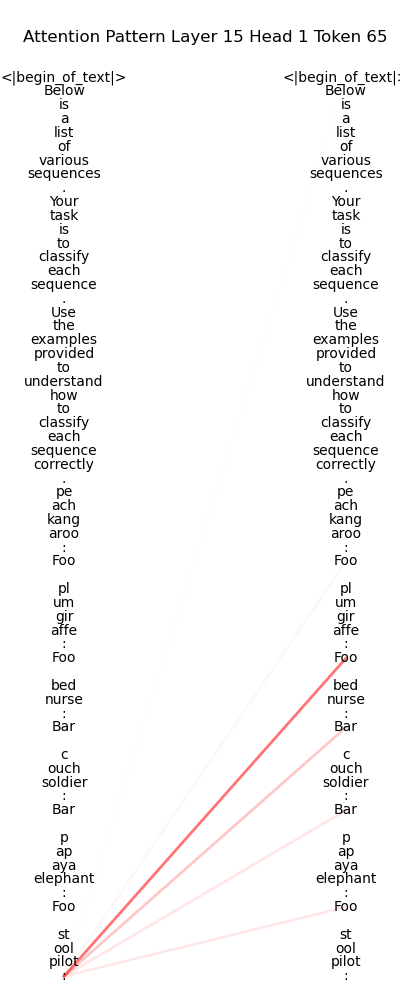}
  \caption{Attention pattern of token 65 in head 1 of layer 15 in Llama-3-8B. Plot 3/5.}
\end{minipage}
\end{figure*}

\begin{figure*}[h!]
\centering
\begin{minipage}{0.48\textwidth}
\centering
  \includegraphics[scale=0.35]{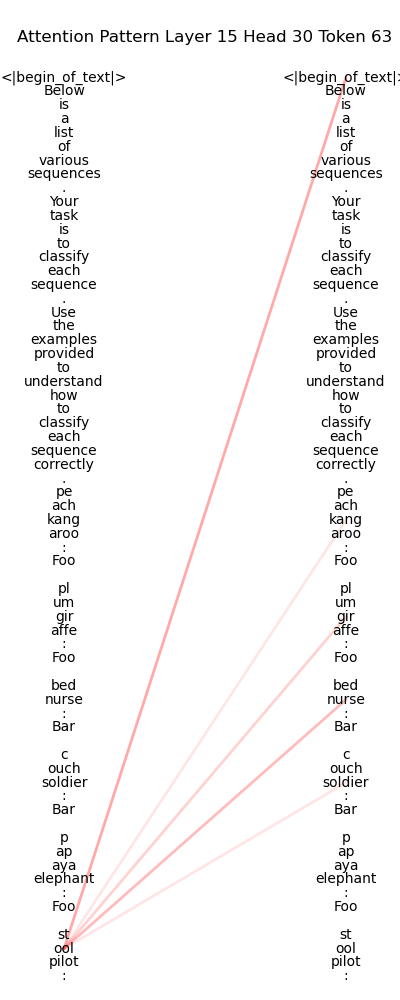}
  \caption{Attention pattern of token 63 in head 30 of layer 15 in Llama-3-8B. Example 3/5.}
\end{minipage}\hfill
\begin{minipage}{0.48\textwidth}
\centering
  \includegraphics[scale=0.35]{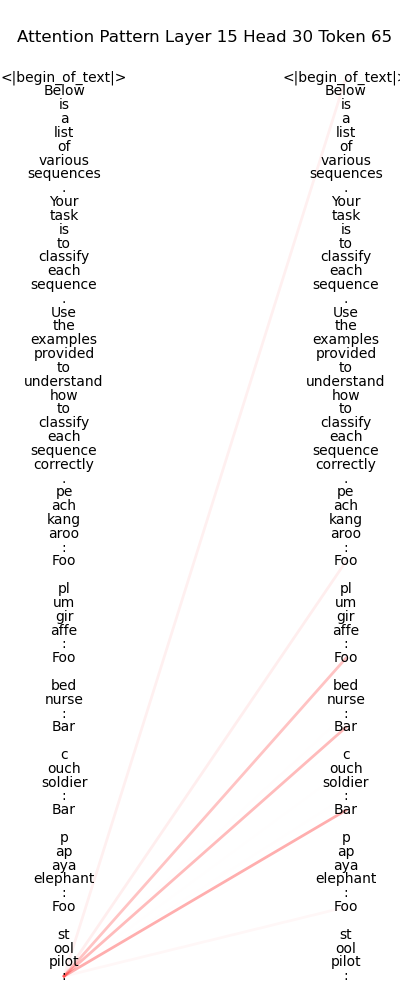}
  \caption{Attention pattern of token 65 in head 30 of layer 15 in Llama-3-8B. Plot 3/5.}
\end{minipage}
\end{figure*}

\begin{figure*}[h!]
\centering
\begin{minipage}{0.48\textwidth}
\centering
  \includegraphics[scale=0.35]{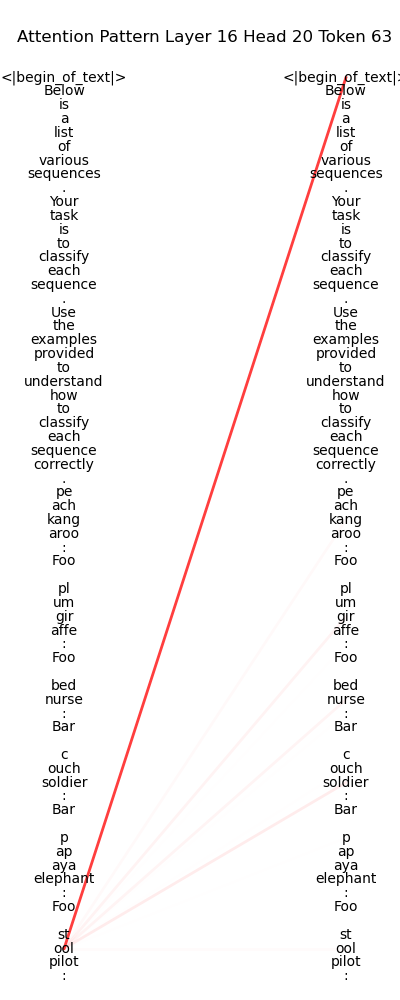}
  \caption{Attention pattern of token 63 in head 20 of layer 16 in Llama-3-8B. Example 3/5.}
\end{minipage}\hfill
\begin{minipage}{0.48\textwidth}
\centering
  \includegraphics[scale=0.35]{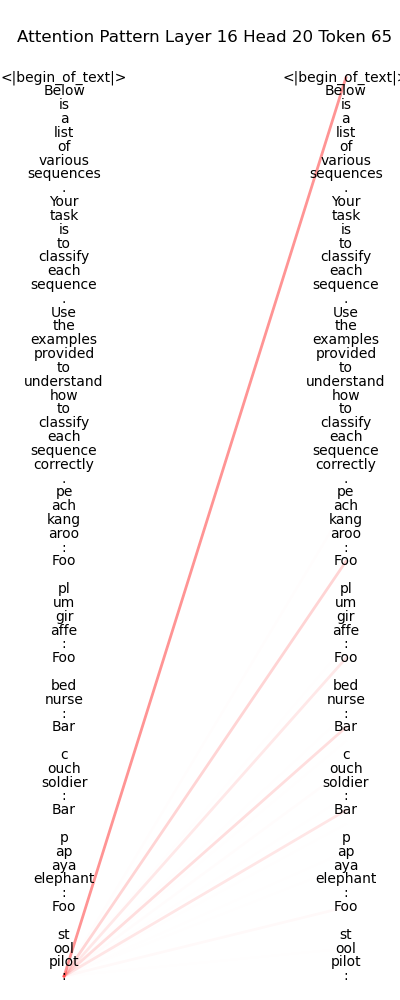}
  \caption{Attention pattern of token 65 in head 20 of layer 16 in Llama-3-8B. Plot 3/5.}
\end{minipage}
\end{figure*}

\begin{figure*}[h!]
\centering
\begin{minipage}{0.48\textwidth}
\centering
  \includegraphics[scale=0.35]{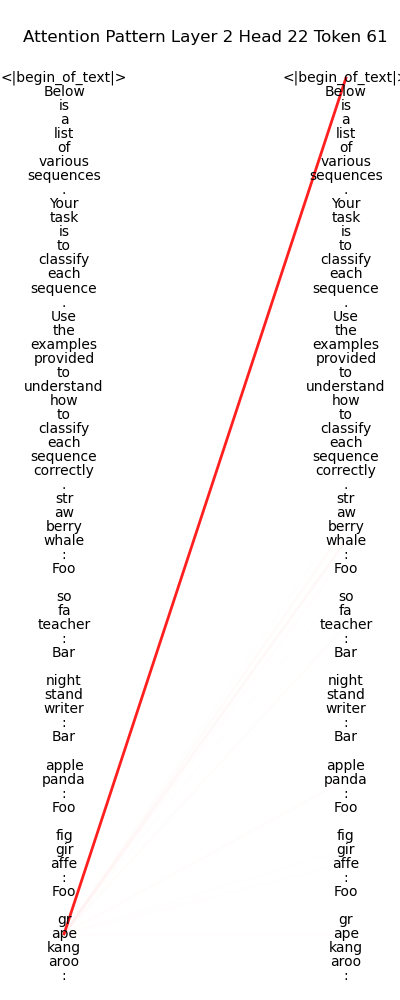}
  \caption{Attention pattern of token 61 in head 22 of layer 2 in Llama-3-8B. Example 4/5.}
\end{minipage}\hfill
\begin{minipage}{0.48\textwidth}
\centering
  \includegraphics[scale=0.35]{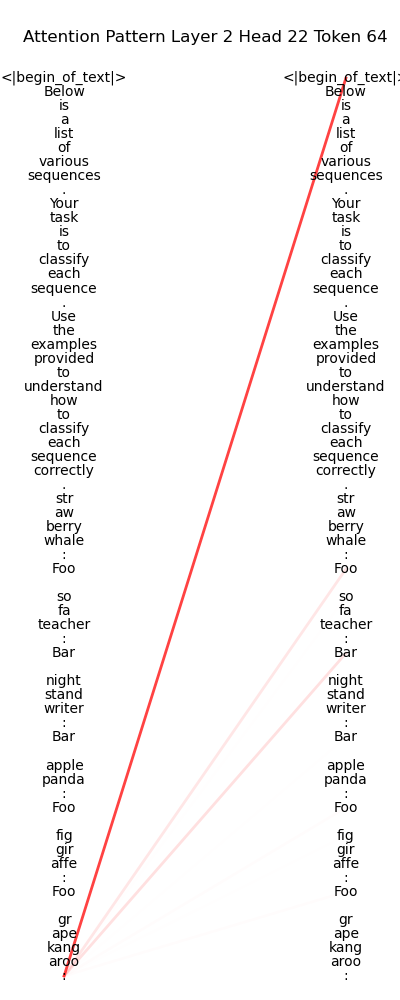}
  \caption{Attention pattern of token 64 in head 22 of layer 2 in Llama-3-8B. Plot 4/5.}
\end{minipage}
\end{figure*}

\begin{figure*}[h!]
\centering
\begin{minipage}{0.48\textwidth}
\centering
  \includegraphics[scale=0.35]{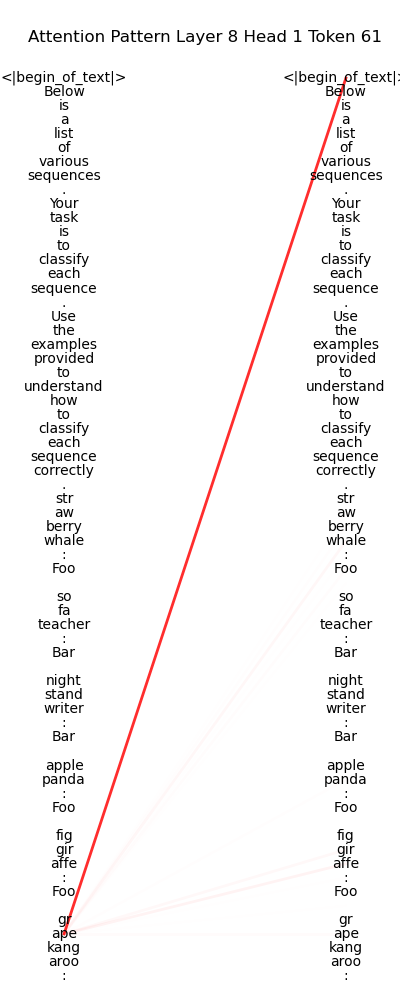}
  \caption{Attention pattern of token 61 in head 1 of layer 8 in Llama-3-8B. Example 4/5.}
\end{minipage}\hfill
\begin{minipage}{0.48\textwidth}
\centering
  \includegraphics[scale=0.35]{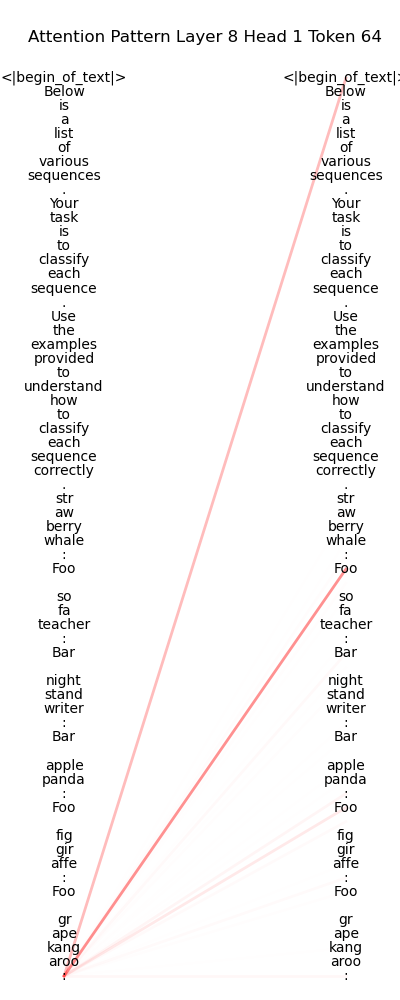}
  \caption{Attention pattern of token 64 in head 1 of layer 8 in Llama-3-8B. Plot 4/5.}
\end{minipage}
\end{figure*}

\begin{figure*}[h!]
\centering
\begin{minipage}{0.48\textwidth}
\centering
  \includegraphics[scale=0.35]{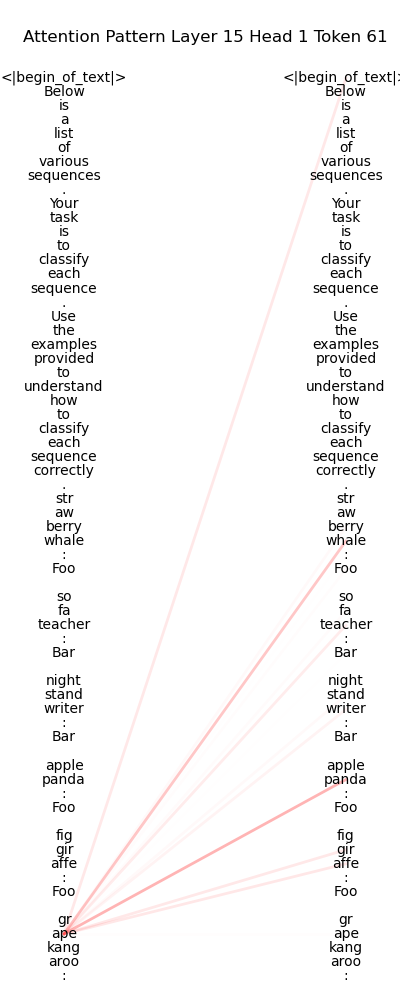}
  \caption{Attention pattern of token 61 in head 1 of layer 15 in Llama-3-8B. Example 4/5.}
\end{minipage}\hfill
\begin{minipage}{0.48\textwidth}
\centering
  \includegraphics[scale=0.35]{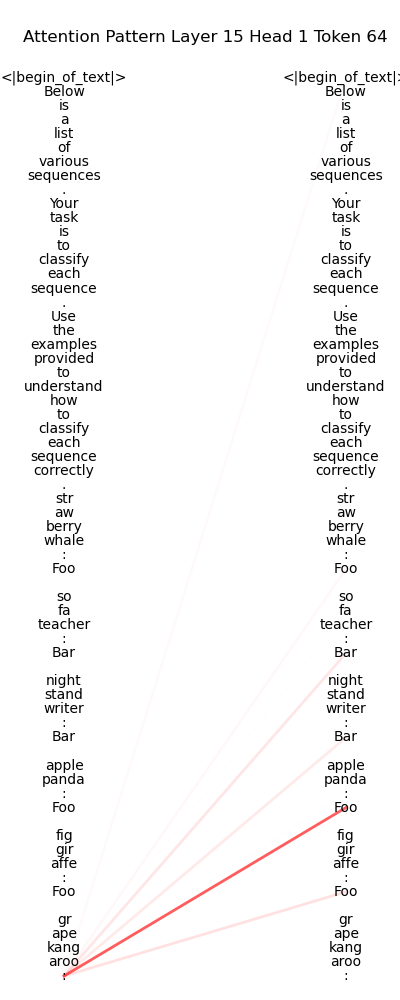}
  \caption{Attention pattern of token 64 in head 1 of layer 15 in Llama-3-8B. Plot 4/5.}
\end{minipage}
\end{figure*}

\begin{figure*}[h!]
\centering
\begin{minipage}{0.48\textwidth}
\centering
  \includegraphics[scale=0.35]{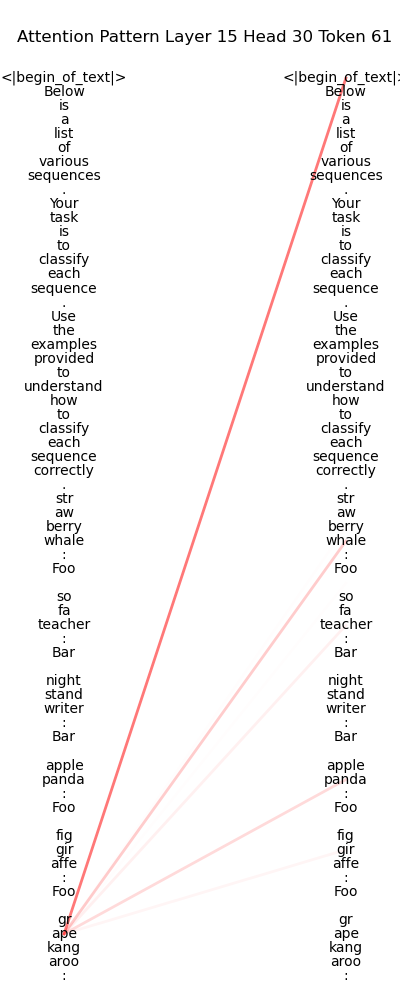}
  \caption{Attention pattern of token 61 in head 30 of layer 15 in Llama-3-8B. Example 4/5.}
\end{minipage}\hfill
\begin{minipage}{0.48\textwidth}
\centering
  \includegraphics[scale=0.35]{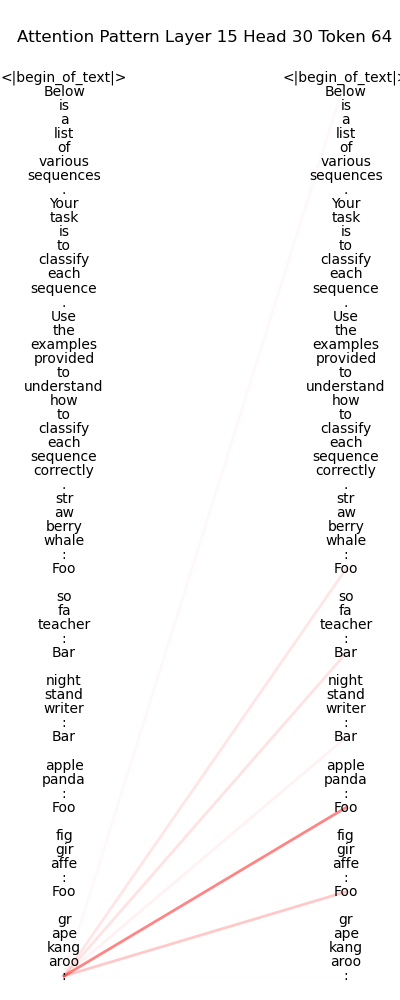}
  \caption{Attention pattern of token 64 in head 30 of layer 15 in Llama-3-8B. Plot 4/5.}
\end{minipage}
\end{figure*}

\begin{figure*}[h!]
\centering
\begin{minipage}{0.48\textwidth}
\centering
  \includegraphics[scale=0.35]{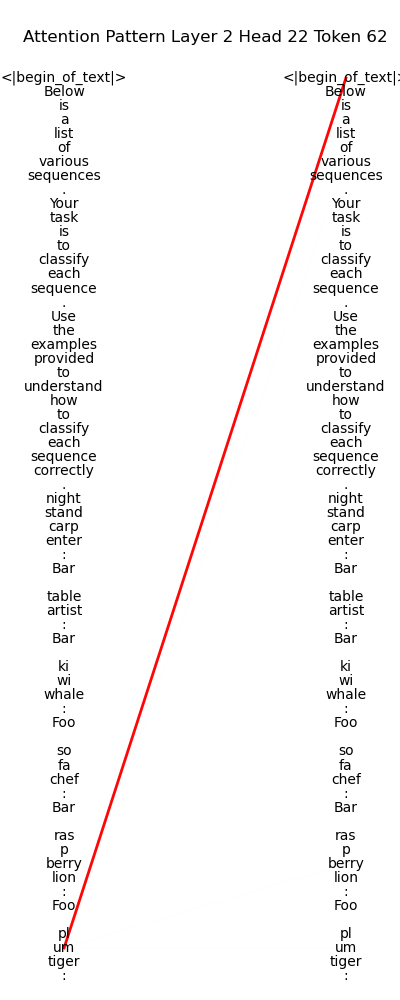}
  \caption{Attention pattern of token 62 in head 22 of layer 2 in Llama-3-8B. Example 5/5.}
\end{minipage}\hfill
\begin{minipage}{0.48\textwidth}
\centering
  \includegraphics[scale=0.35]{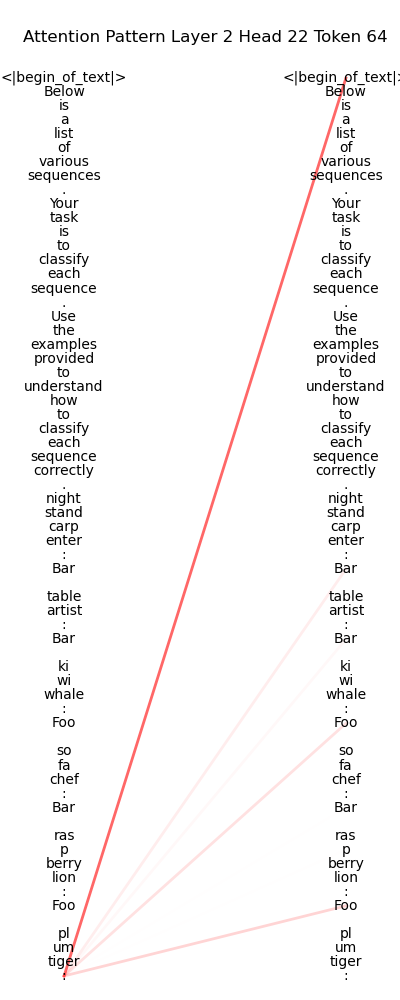}
  \caption{Attention pattern of token 64 in head 22 of layer 2 in Llama-3-8B. Plot 5/5.}
\end{minipage}
\end{figure*}

\begin{figure*}[h!]
\centering
\begin{minipage}{0.48\textwidth}
\centering
  \includegraphics[scale=0.35]{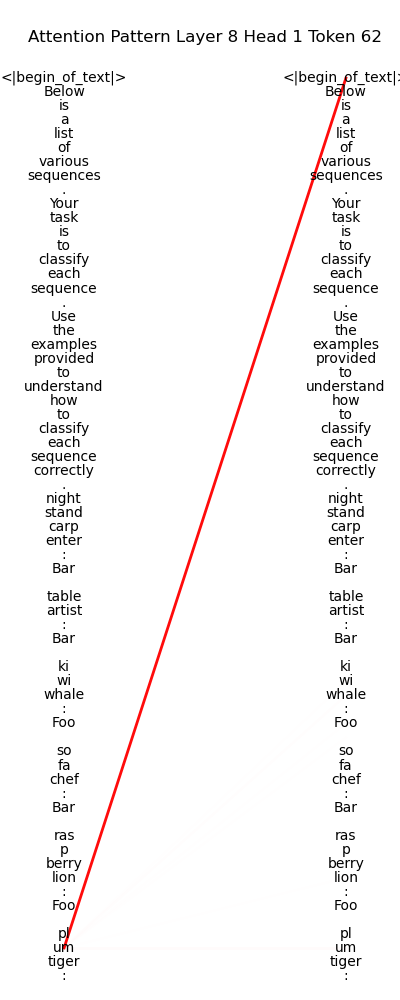}
  \caption{Attention pattern of token 62 in head 1 of layer 8 in Llama-3-8B. Example 5/5.}
\end{minipage}\hfill
\begin{minipage}{0.48\textwidth}
\centering
  \includegraphics[scale=0.35]{images/Appendix/binary/plot_5_layer_8_head_1_token_62.png}
  \caption{Attention pattern of token 62 in head 1 of layer 8 in Llama-3-8B. Plot 5/5.}
\end{minipage}
\end{figure*}

\begin{figure*}[h!]
\centering
\begin{minipage}{0.48\textwidth}
\centering
  \includegraphics[scale=0.35]{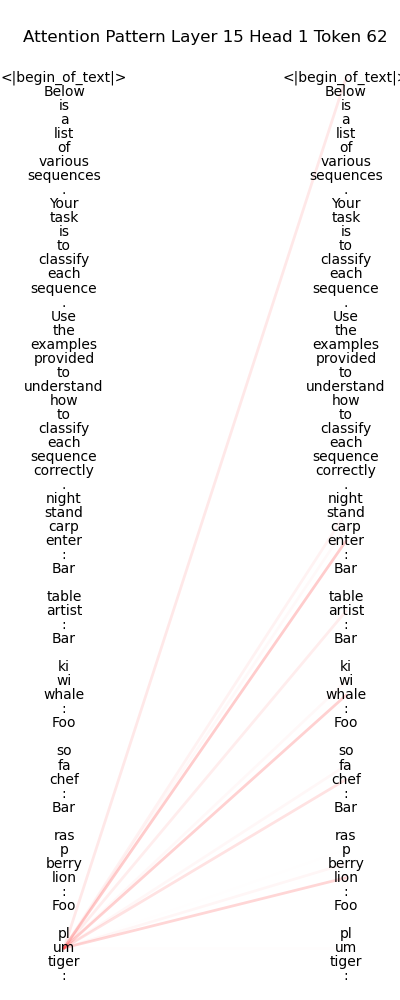}
  \caption{Attention pattern of token 62 in head 1 of layer 15 in Llama-3-8B. Example 5/5.}
\end{minipage}\hfill
\begin{minipage}{0.48\textwidth}
\centering
  \includegraphics[scale=0.35]{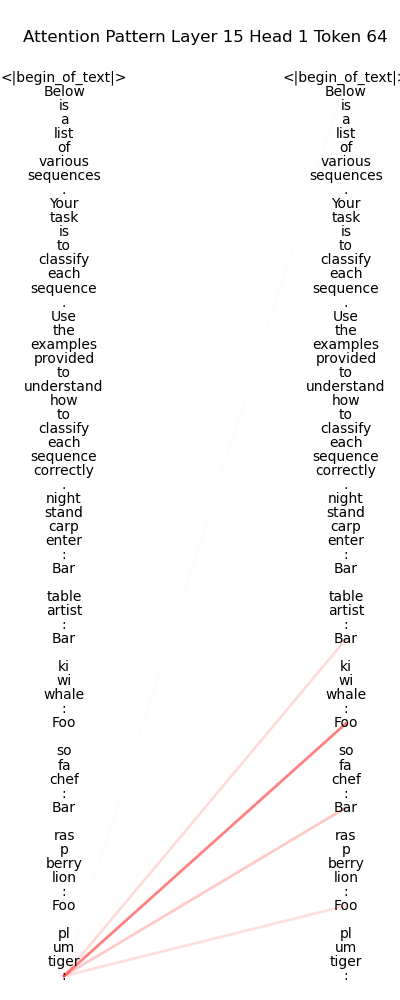}
  \caption{Attention pattern of token 64 in head 1 of layer 15 in Llama-3-8B. Plot 5/5.}
\end{minipage}
\end{figure*}

\begin{figure*}[h!]
\centering
\begin{minipage}{0.48\textwidth}
\centering
  \includegraphics[scale=0.35]{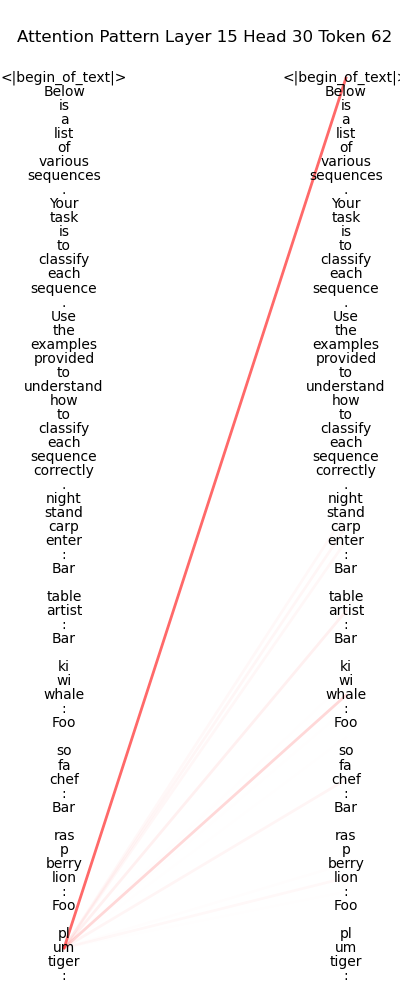}
  \caption{Attention pattern of token 62 in head 30 of layer 15 in Llama-3-8B. Example 5/5.}
\end{minipage}\hfill
\begin{minipage}{0.48\textwidth}
\centering
  \includegraphics[scale=0.35]{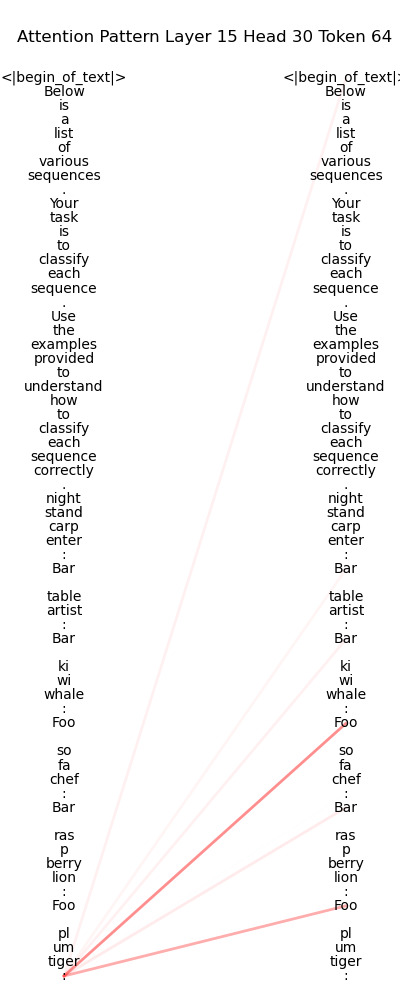}
  \caption{Attention pattern of token 64 in head 30 of layer 15 in Llama-3-8B. Plot 5/5.}
\end{minipage}
\end{figure*}

\begin{figure*}[h!]
\centering
\begin{minipage}{0.48\textwidth}
\centering
  \includegraphics[scale=0.35]{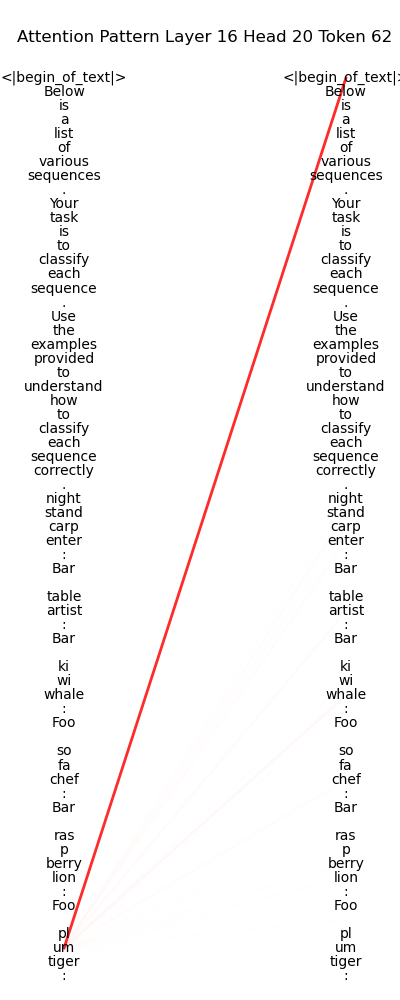}
  \caption{Attention pattern of token 62 in head 20 of layer 16 in Llama-3-8B. Example 5/5.}
\end{minipage}\hfill
\begin{minipage}{0.48\textwidth}
\centering
  \includegraphics[scale=0.35]{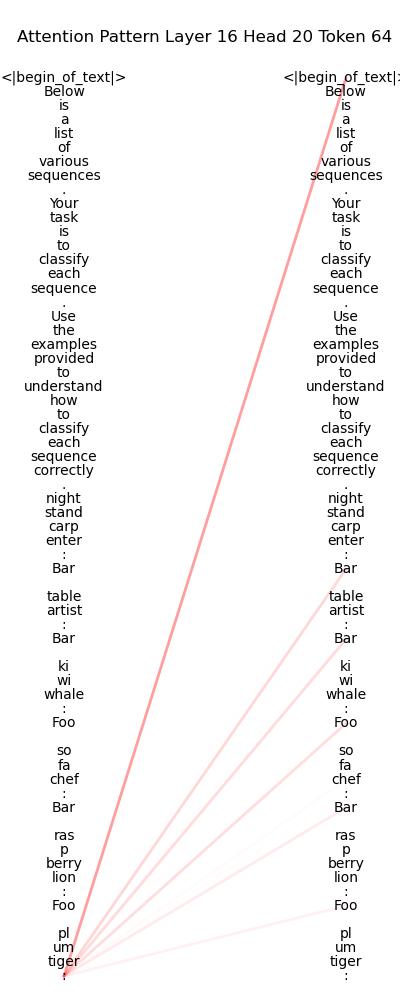}
  \caption{Attention pattern of token 64 in head 20 of layer 16 in Llama-3-8B. Plot 5/5.}
\end{minipage}
\end{figure*}

\clearpage
\subsubsection{WordSeq 1 (4-way)}

\begin{figure*}[h!]
\centering
\begin{minipage}{0.48\textwidth}
\centering
  \includegraphics[scale=0.35]{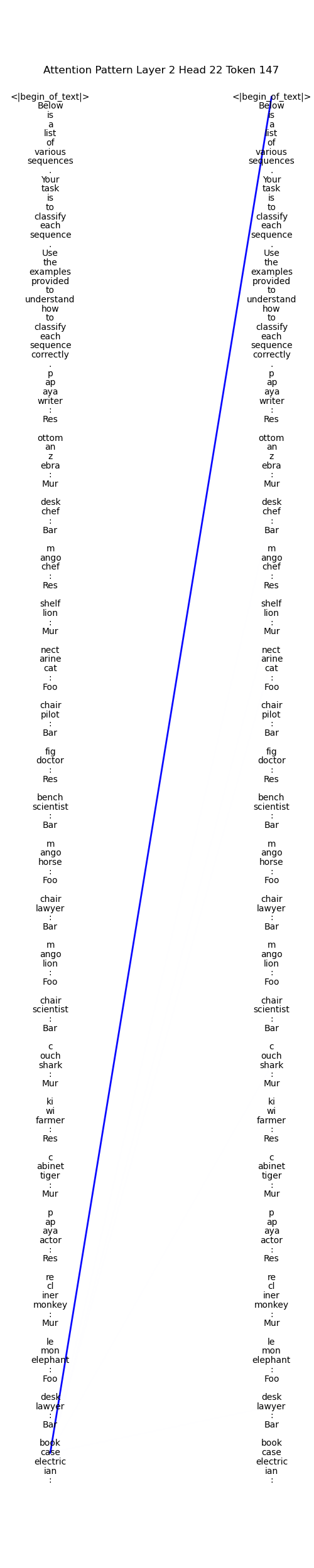}
  \caption{Attention pattern of token 147 in head 22 of layer 2 in Llama-3-8B. Example 1/5.}
\end{minipage}\hfill
\begin{minipage}{0.48\textwidth}
\centering
  \includegraphics[scale=0.35]{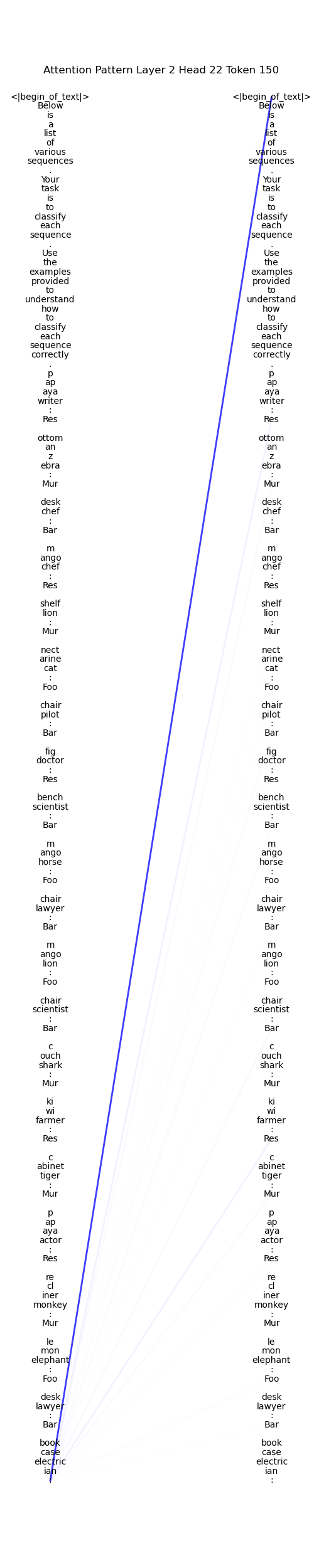}
  \caption{Attention pattern of token 150 in head 22 of layer 2 in Llama-3-8B. Plot 1/5.}
\end{minipage}
\end{figure*}
\newpage

\begin{figure*}[h!]
\centering
\begin{minipage}{0.48\textwidth}
\centering
  \includegraphics[scale=0.35]{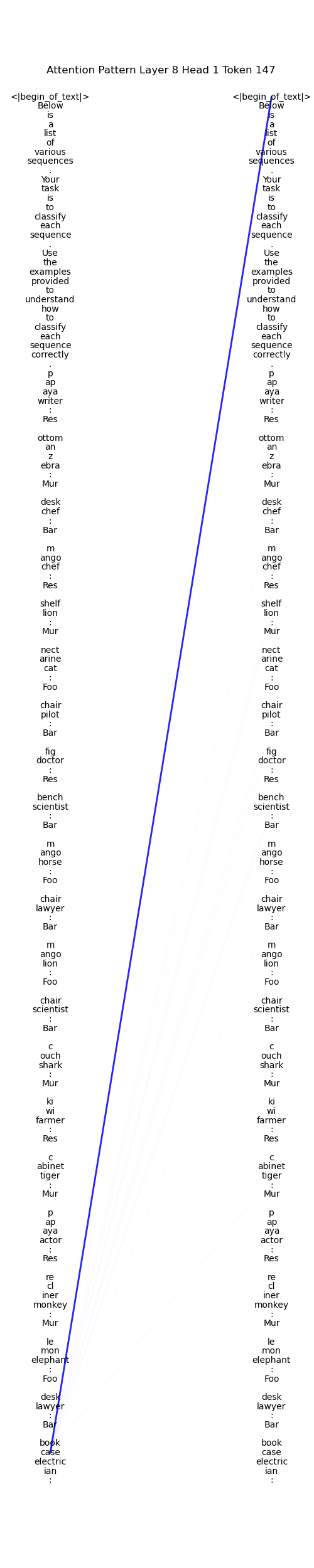}
  \caption{Attention pattern of token 147 in head 1 of layer 8 in Llama-3-8B. Example 1/5.}
\end{minipage}\hfill
\begin{minipage}{0.48\textwidth}
\centering
  \includegraphics[scale=0.35]{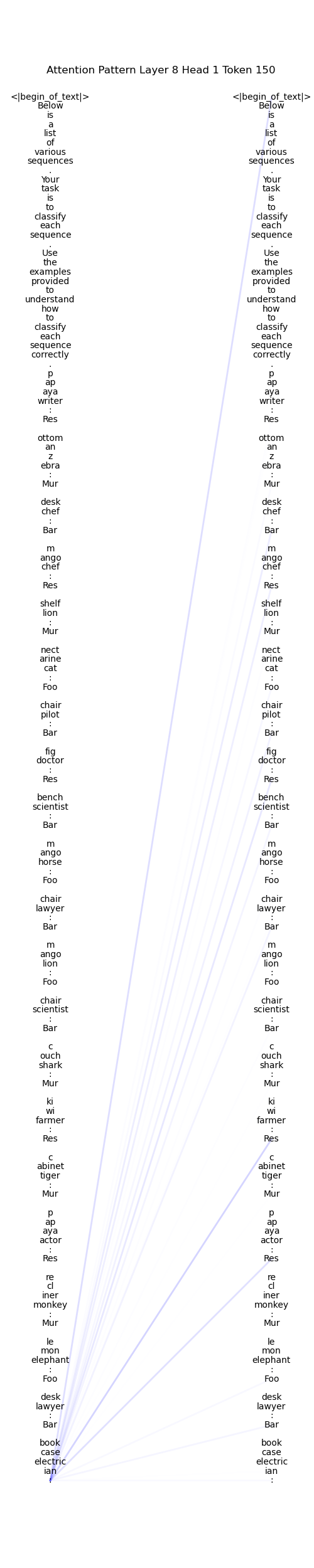}
  \caption{Attention pattern of token 150 in head 1 of layer 8 in Llama-3-8B. Plot 1/5.}
\end{minipage}
\end{figure*}
\newpage

\begin{figure*}[h!]
\centering
\begin{minipage}{0.48\textwidth}
\centering
  \includegraphics[scale=0.35]{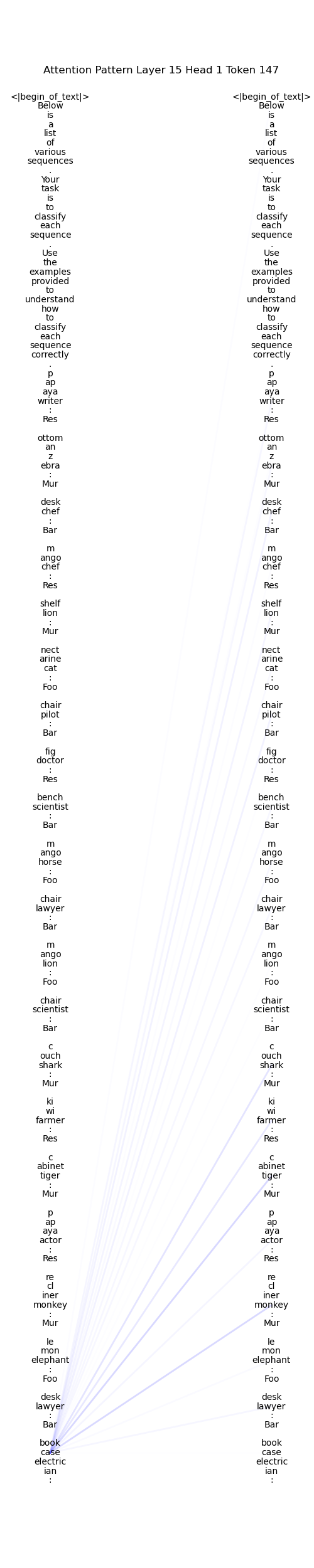}
  \caption{Attention pattern of token 147 in head 1 of layer 15 in Llama-3-8B. Example 1/5.}
\end{minipage}\hfill
\begin{minipage}{0.48\textwidth}
\centering
  \includegraphics[scale=0.35]{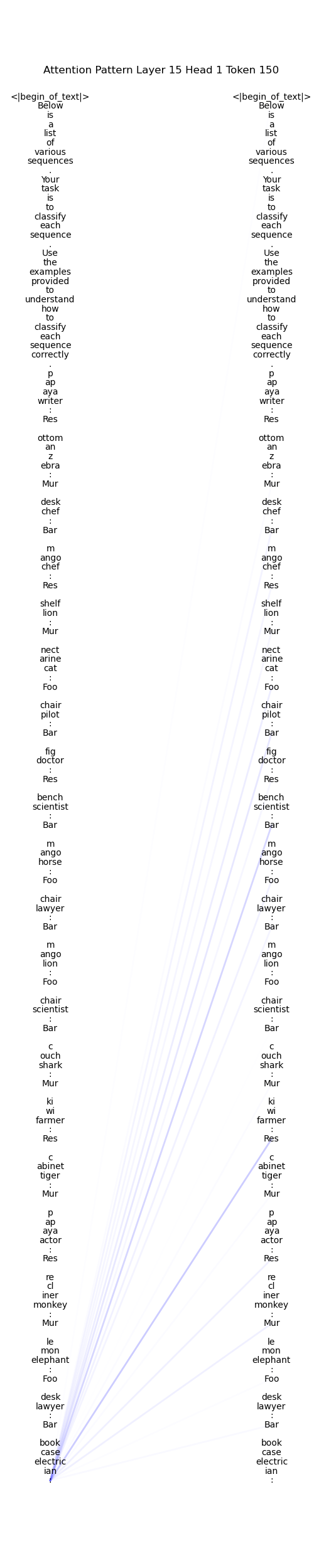}
  \caption{Attention pattern of token 150 in head 1 of layer 15 in Llama-3-8B. Plot 1/5.}
\end{minipage}
\end{figure*}
\newpage

\begin{figure*}[h!]
\centering
\begin{minipage}{0.48\textwidth}
\centering
  \includegraphics[scale=0.35]{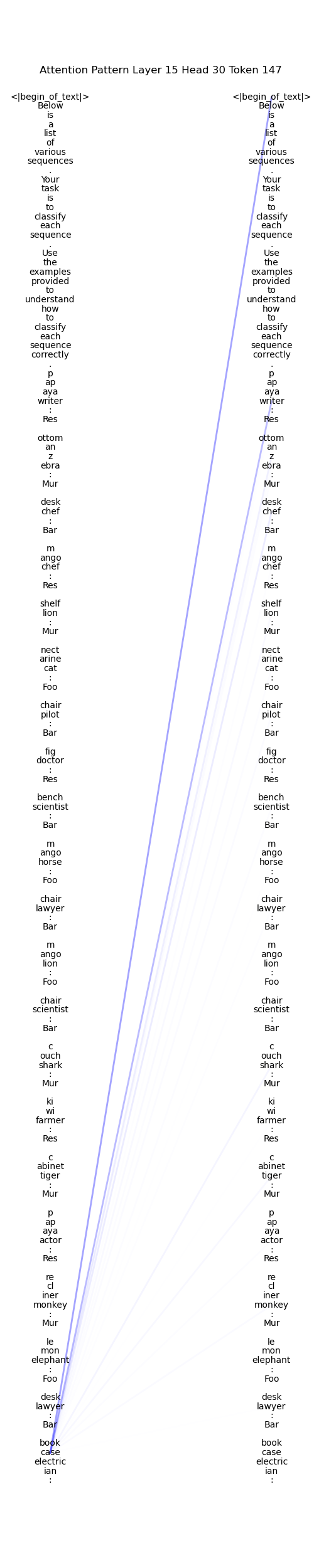}
  \caption{Attention pattern of token 147 in head 30 of layer 15 in Llama-3-8B. Example 1/5.}
\end{minipage}\hfill
\begin{minipage}{0.48\textwidth}
\centering
  \includegraphics[scale=0.35]{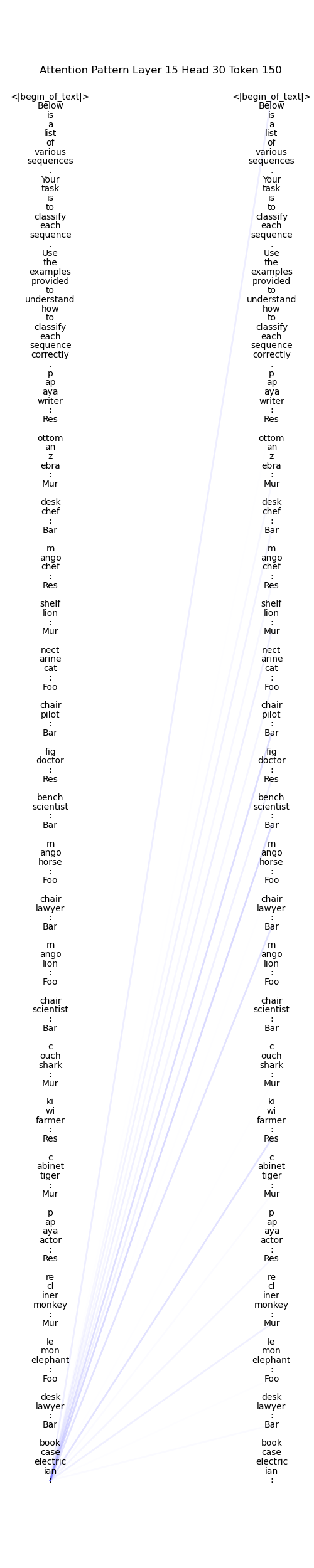}
  \caption{Attention pattern of token 150 in head 30 of layer 15 in Llama-3-8B. Plot 1/5.}
\end{minipage}
\end{figure*}
\newpage

\begin{figure*}[h!]
\centering
\begin{minipage}{0.48\textwidth}
\centering
  \includegraphics[scale=0.35]{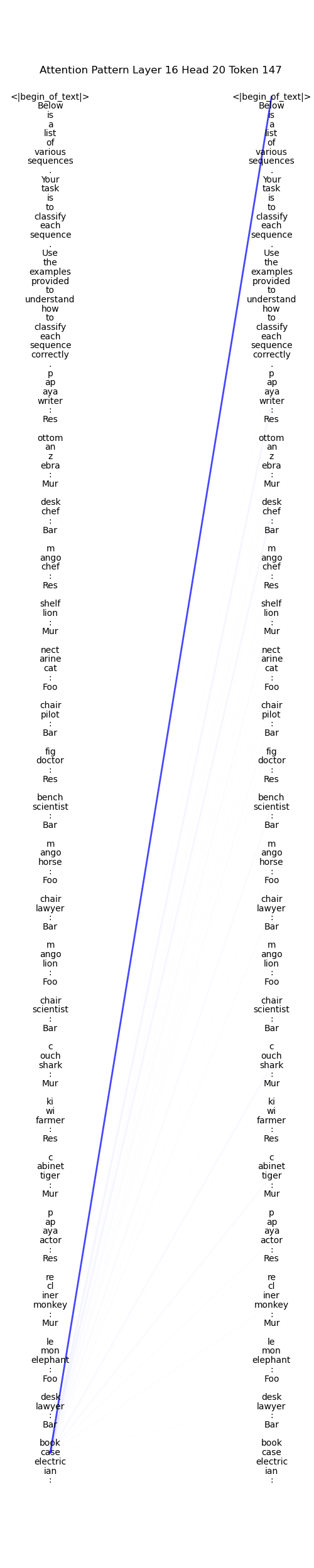}
  \caption{Attention pattern of token 147 in head 20 of layer 16 in Llama-3-8B. Example 1/5.}
\end{minipage}\hfill
\begin{minipage}{0.48\textwidth}
\centering
  \includegraphics[scale=0.35]{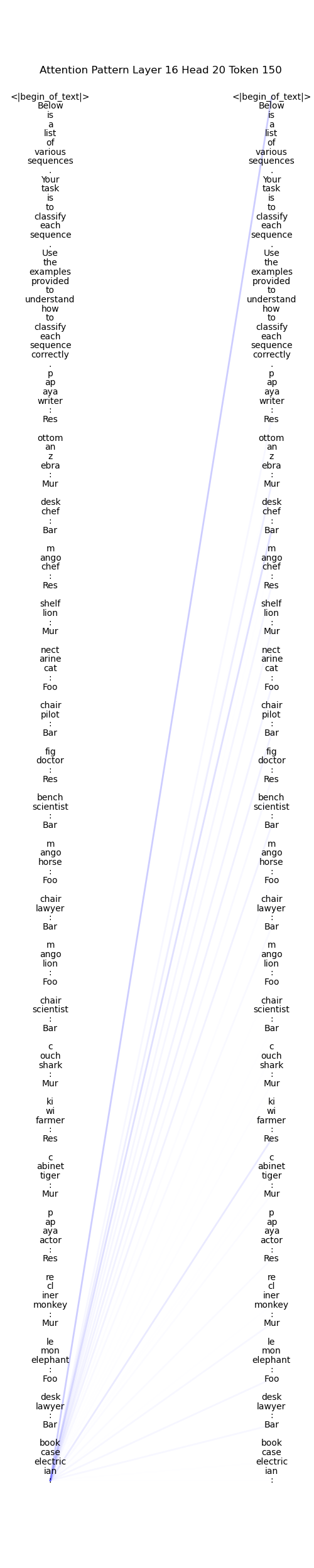}
  \caption{Attention pattern of token 150 in head 20 of layer 16 in Llama-3-8B. Plot 1/5.}
\end{minipage}
\end{figure*}
\newpage

\begin{figure*}[h!]
\centering
\begin{minipage}{0.48\textwidth}
\centering
  \includegraphics[scale=0.35]{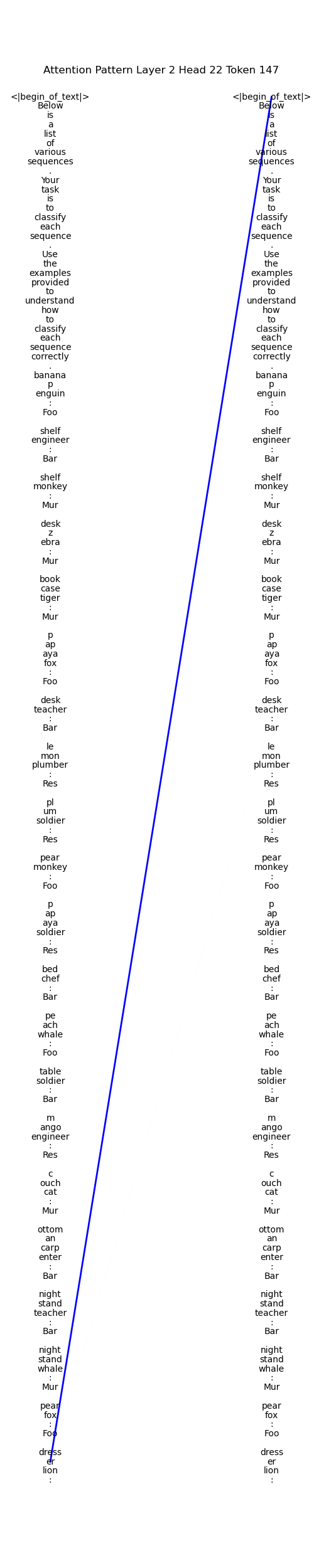}
  \caption{Attention pattern of token 147 in head 22 of layer 2 in Llama-3-8B. Example 2/5.}
\end{minipage}\hfill
\begin{minipage}{0.48\textwidth}
\centering
  \includegraphics[scale=0.35]{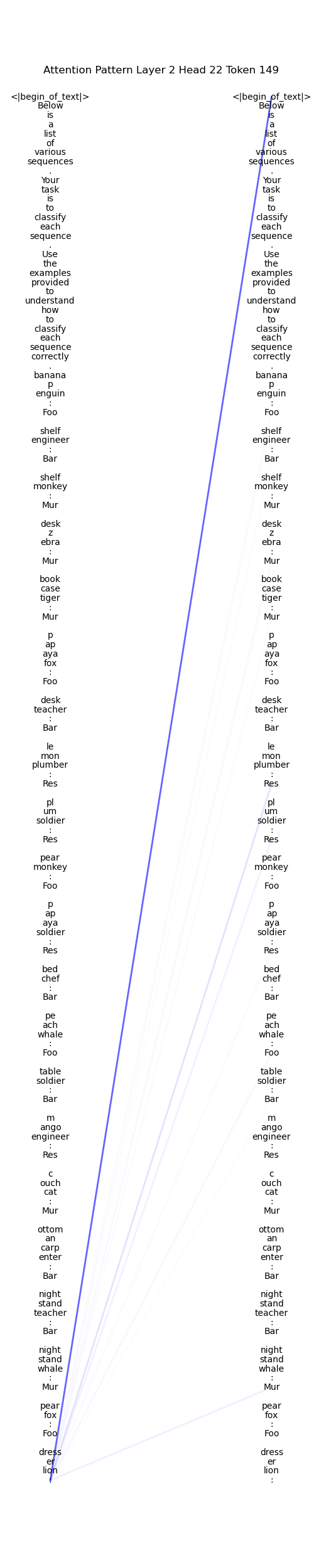}
  \caption{Attention pattern of token 149 in head 22 of layer 2 in Llama-3-8B. Plot 2/5.}
\end{minipage}
\end{figure*}
\newpage

\begin{figure*}[h!]
\centering
\begin{minipage}{0.48\textwidth}
\centering
  \includegraphics[scale=0.35]{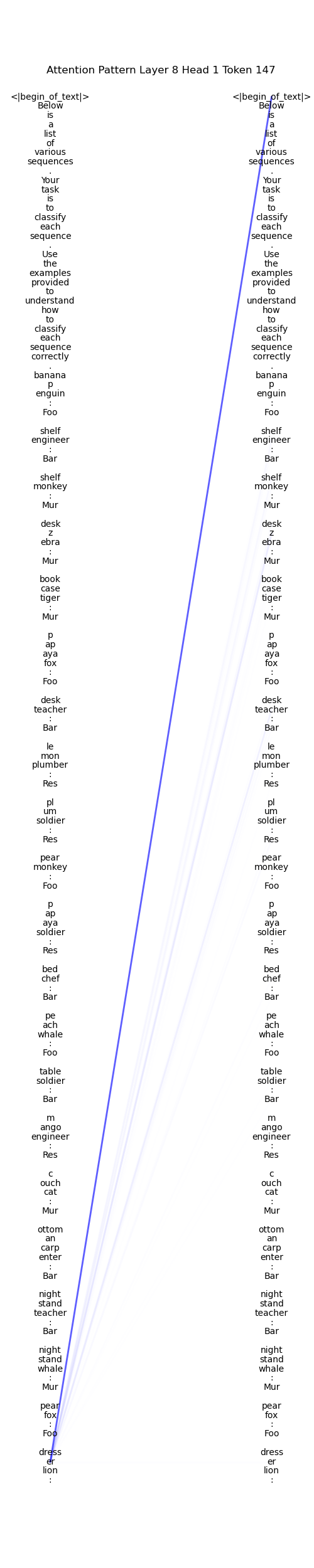}
  \caption{Attention pattern of token 147 in head 1 of layer 8 in Llama-3-8B. Example 2/5.}
\end{minipage}\hfill
\begin{minipage}{0.48\textwidth}
\centering
  \includegraphics[scale=0.35]{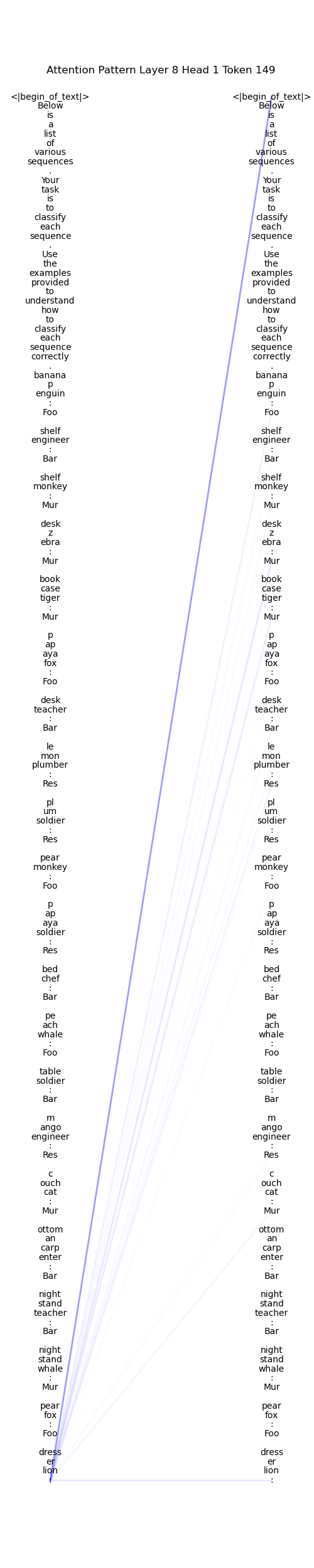}
  \caption{Attention pattern of token 149 in head 1 of layer 8 in Llama-3-8B. Plot 2/5.}
\end{minipage}
\end{figure*}
\newpage

\begin{figure*}[h!]
\centering
\begin{minipage}{0.48\textwidth}
\centering
  \includegraphics[scale=0.35]{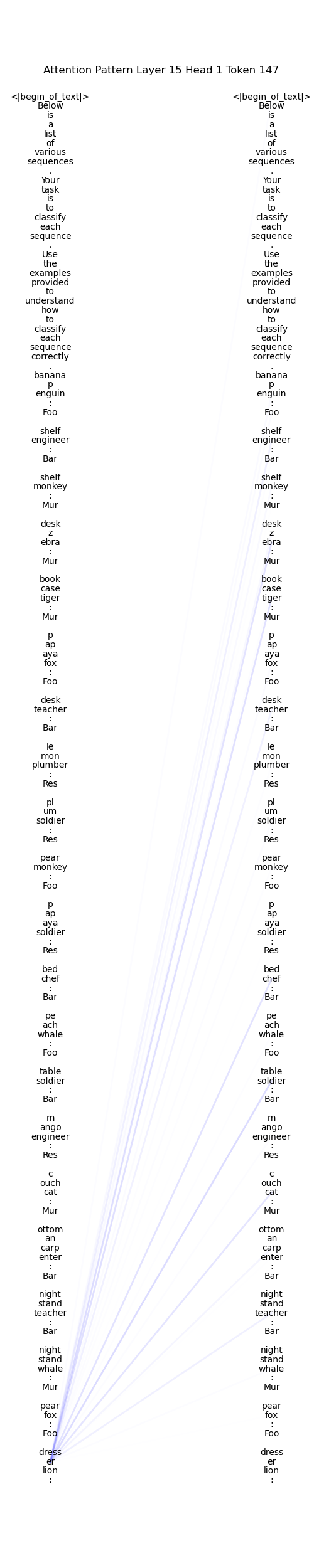}
  \caption{Attention pattern of token 147 in head 1 of layer 15 in Llama-3-8B. Example 2/5.}
\end{minipage}\hfill
\begin{minipage}{0.48\textwidth}
\centering
  \includegraphics[scale=0.35]{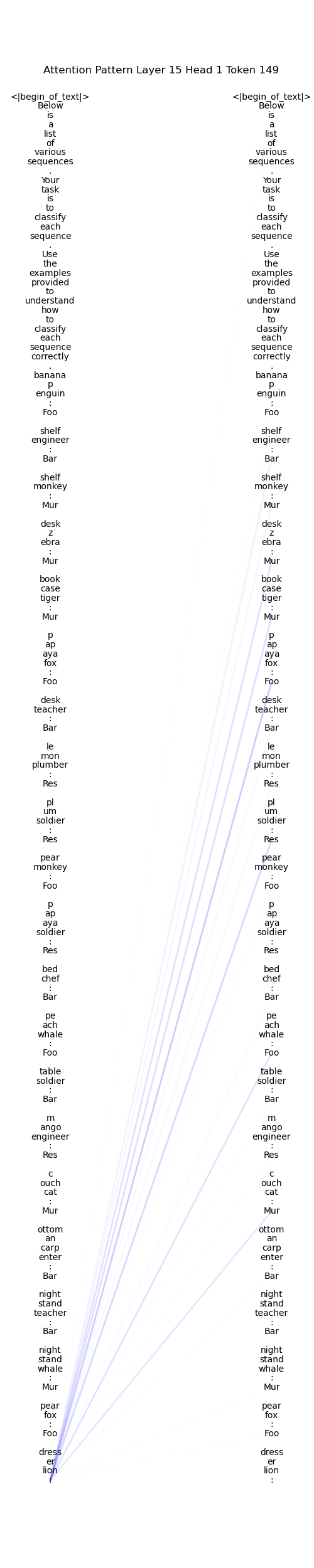}
  \caption{Attention pattern of token 149 in head 1 of layer 15 in Llama-3-8B. Plot 2/5.}
\end{minipage}
\end{figure*}
\newpage

\begin{figure*}[h!]
\centering
\begin{minipage}{0.48\textwidth}
\centering
  \includegraphics[scale=0.35]{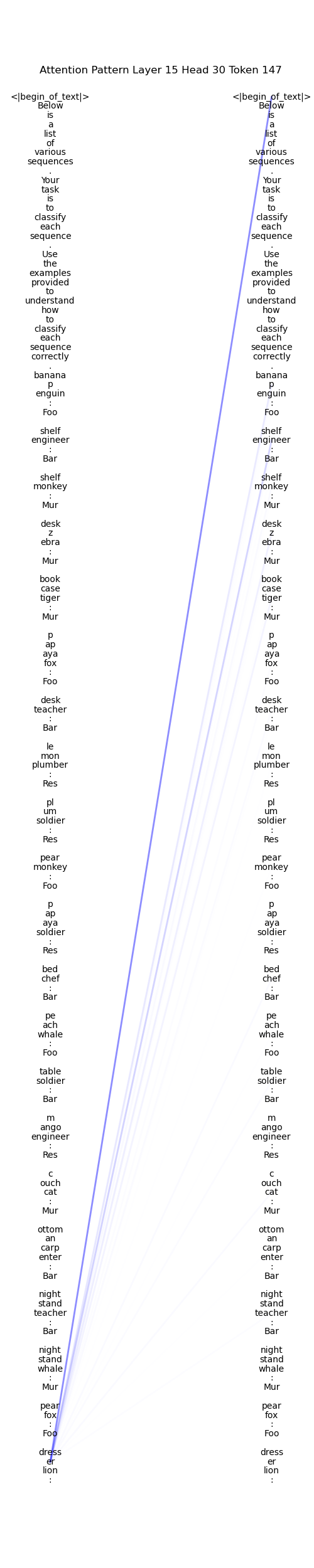}
  \caption{Attention pattern of token 147 in head 30 of layer 15 in Llama-3-8B. Example 2/5.}
\end{minipage}\hfill
\begin{minipage}{0.48\textwidth}
\centering
  \includegraphics[scale=0.35]{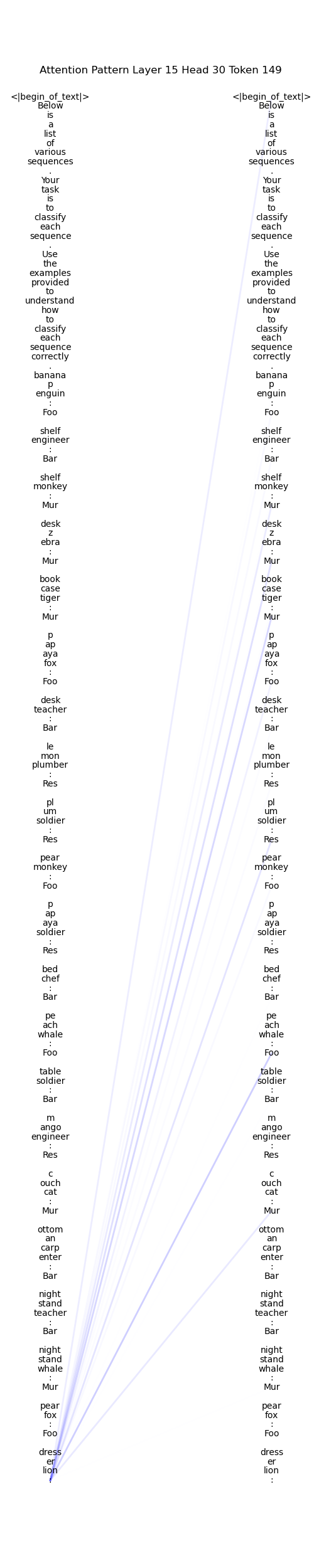}
  \caption{Attention pattern of token 149 in head 30 of layer 15 in Llama-3-8B. Plot 2/5.}
\end{minipage}
\end{figure*}
\newpage

\begin{figure*}[h!]
\centering
\begin{minipage}{0.48\textwidth}
\centering
  \includegraphics[scale=0.35]{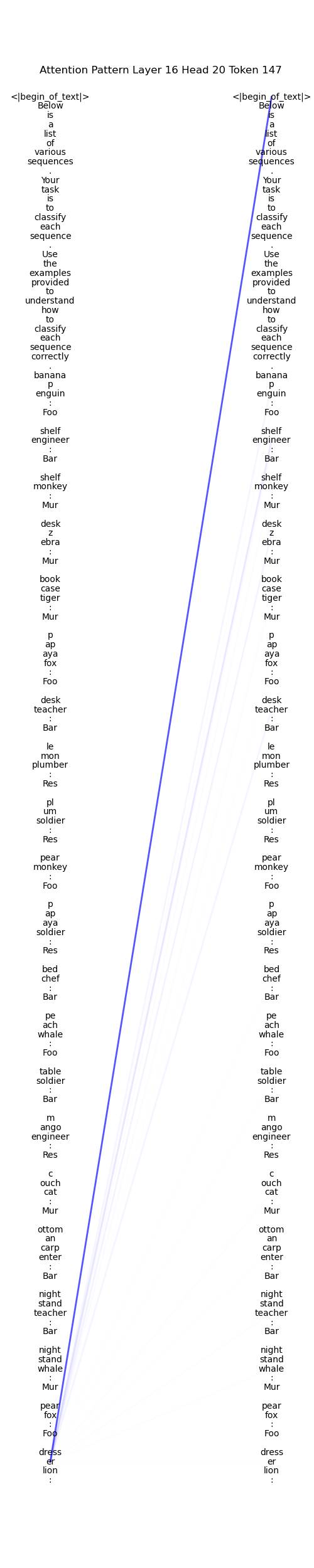}
  \caption{Attention pattern of token 147 in head 20 of layer 16 in Llama-3-8B. Example 2/5.}
\end{minipage}\hfill
\begin{minipage}{0.48\textwidth}
\centering
  \includegraphics[scale=0.35]{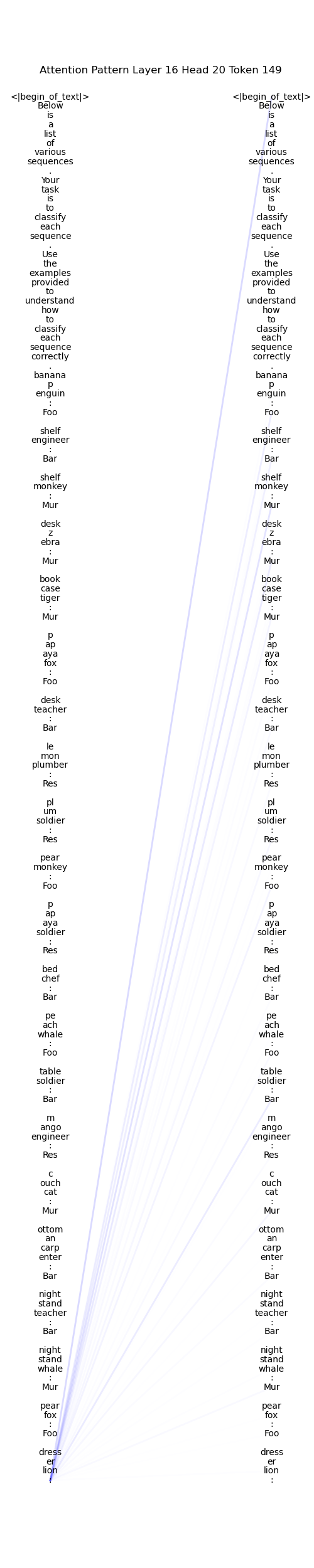}
  \caption{Attention pattern of token 149 in head 20 of layer 16 in Llama-3-8B. Plot 2/5.}
\end{minipage}
\end{figure*}
\newpage

\begin{figure*}[h!]
\centering
\begin{minipage}{0.48\textwidth}
\centering
  \includegraphics[scale=0.35]{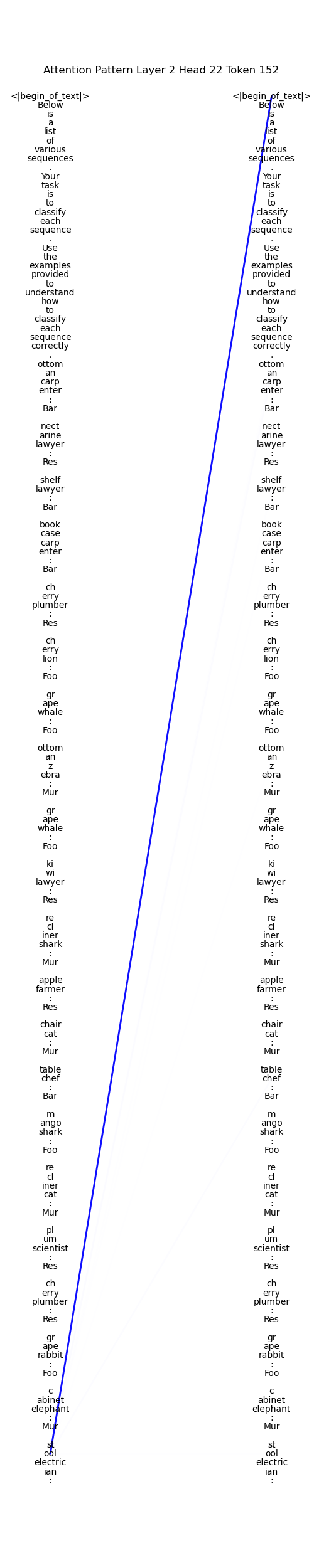}
  \caption{Attention pattern of token 152 in head 22 of layer 2 in Llama-3-8B. Example 3/5.}
\end{minipage}\hfill
\begin{minipage}{0.48\textwidth}
\centering
  \includegraphics[scale=0.35]{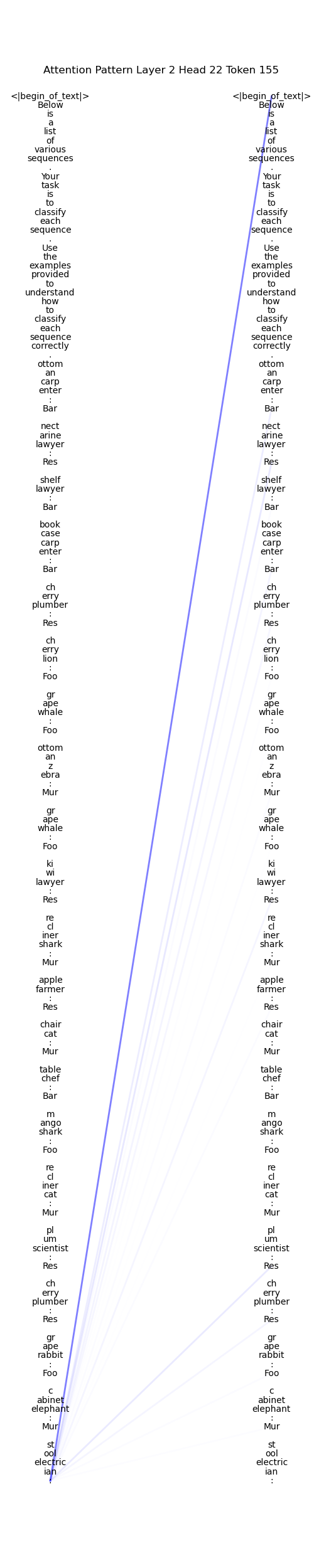}
  \caption{Attention pattern of token 155 in head 22 of layer 2 in Llama-3-8B. Plot 3/5.}
\end{minipage}
\end{figure*}
\newpage

\begin{figure*}[h!]
\centering
\begin{minipage}{0.48\textwidth}
\centering
  \includegraphics[scale=0.35]{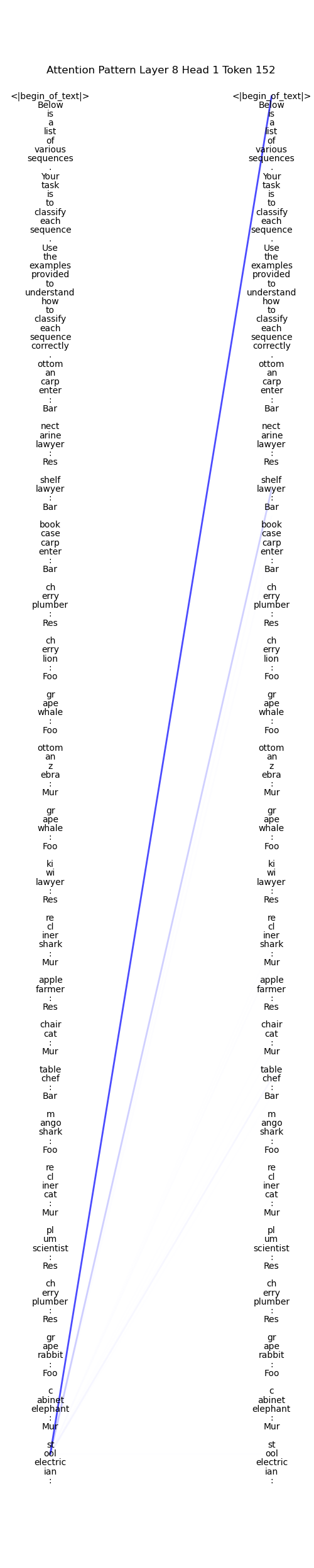}
  \caption{Attention pattern of token 152 in head 1 of layer 8 in Llama-3-8B. Example 3/5.}
\end{minipage}\hfill
\begin{minipage}{0.48\textwidth}
\centering
  \includegraphics[scale=0.35]{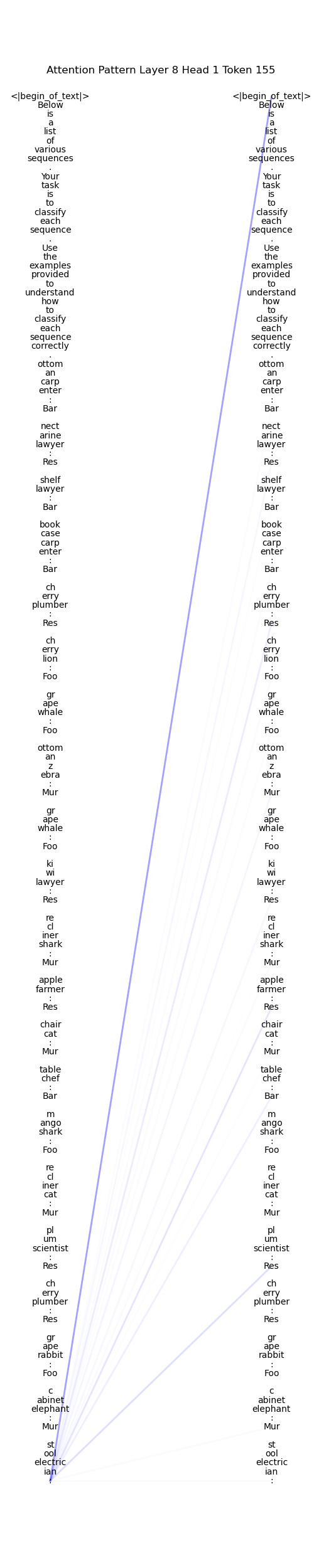}
  \caption{Attention pattern of token 155 in head 1 of layer 8 in Llama-3-8B. Plot 3/5.}
\end{minipage}
\end{figure*}
\newpage

\begin{figure*}[h!]
\centering
\begin{minipage}{0.48\textwidth}
\centering
  \includegraphics[scale=0.35]{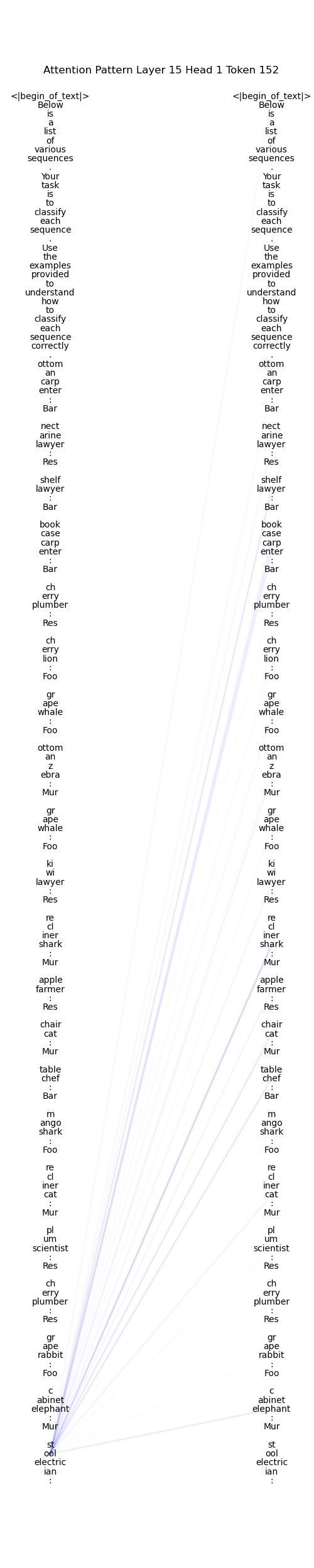}
  \caption{Attention pattern of token 152 in head 1 of layer 15 in Llama-3-8B. Example 3/5.}
\end{minipage}\hfill
\begin{minipage}{0.48\textwidth}
\centering
  \includegraphics[scale=0.35]{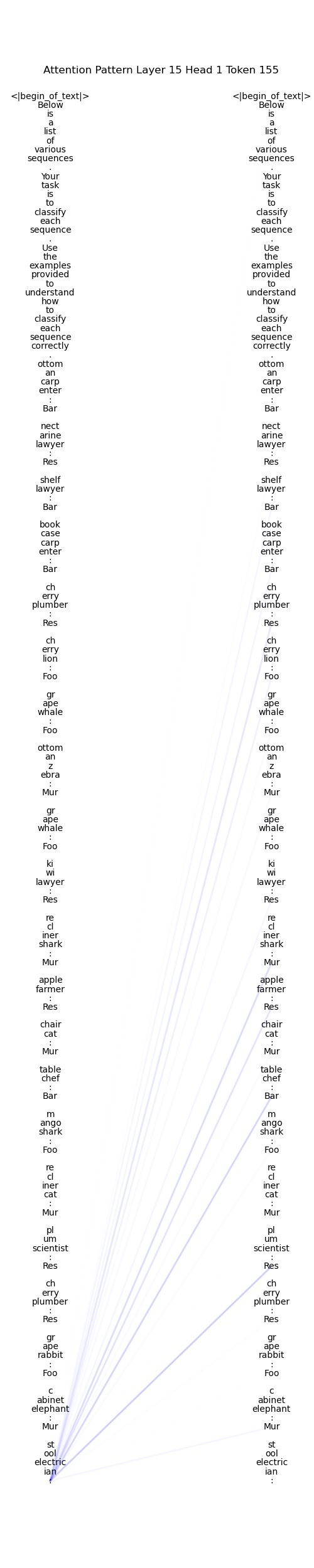}
  \caption{Attention pattern of token 155 in head 1 of layer 15 in Llama-3-8B. Plot 3/5.}
\end{minipage}
\end{figure*}
\newpage

\begin{figure*}[h!]
\centering
\begin{minipage}{0.48\textwidth}
\centering
  \includegraphics[scale=0.35]{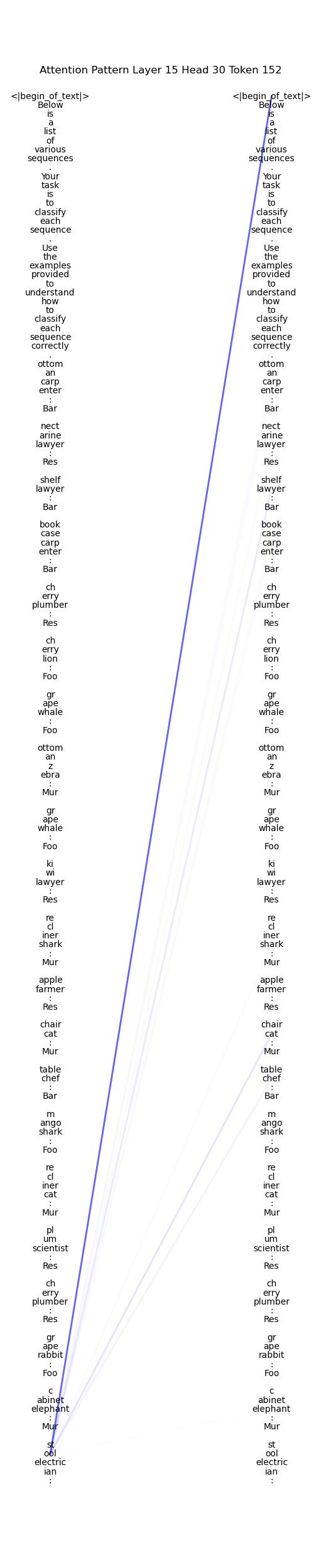}
  \caption{Attention pattern of token 152 in head 30 of layer 15 in Llama-3-8B. Example 3/5.}
\end{minipage}\hfill
\begin{minipage}{0.48\textwidth}
\centering
  \includegraphics[scale=0.35]{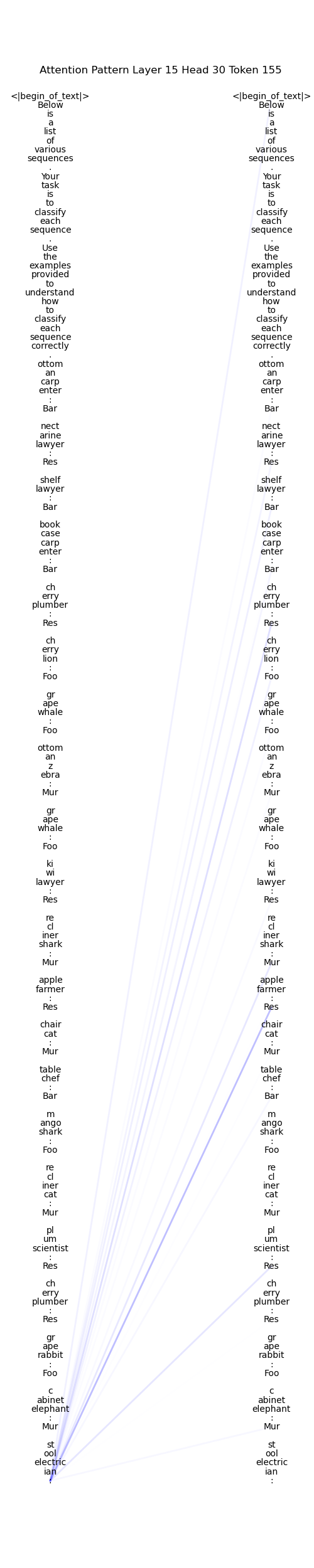}
  \caption{Attention pattern of token 155 in head 30 of layer 15 in Llama-3-8B. Plot 3/5.}
\end{minipage}
\end{figure*}
\newpage

\begin{figure*}[h!]
\centering
\begin{minipage}{0.48\textwidth}
\centering
  \includegraphics[scale=0.35]{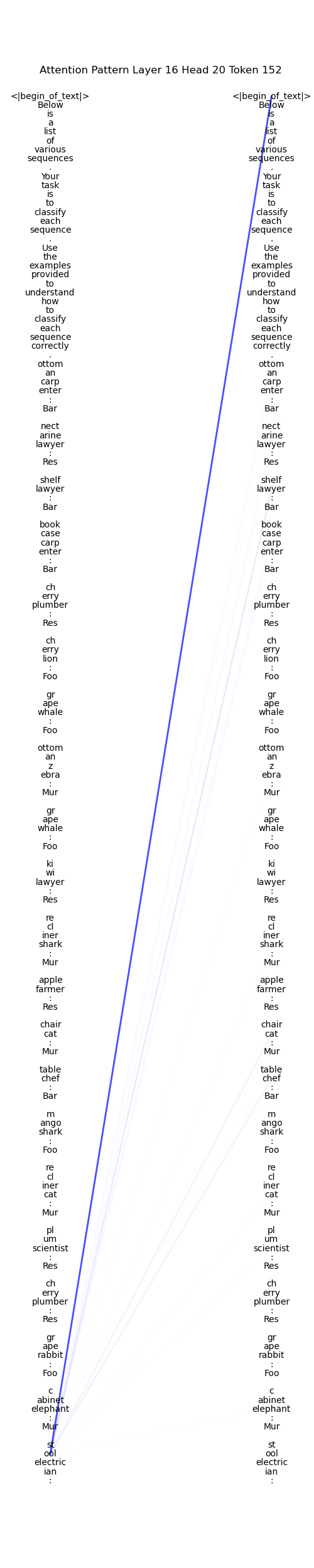}
  \caption{Attention pattern of token 152 in head 20 of layer 16 in Llama-3-8B. Example 3/5.}
\end{minipage}\hfill
\begin{minipage}{0.48\textwidth}
\centering
  \includegraphics[scale=0.35]{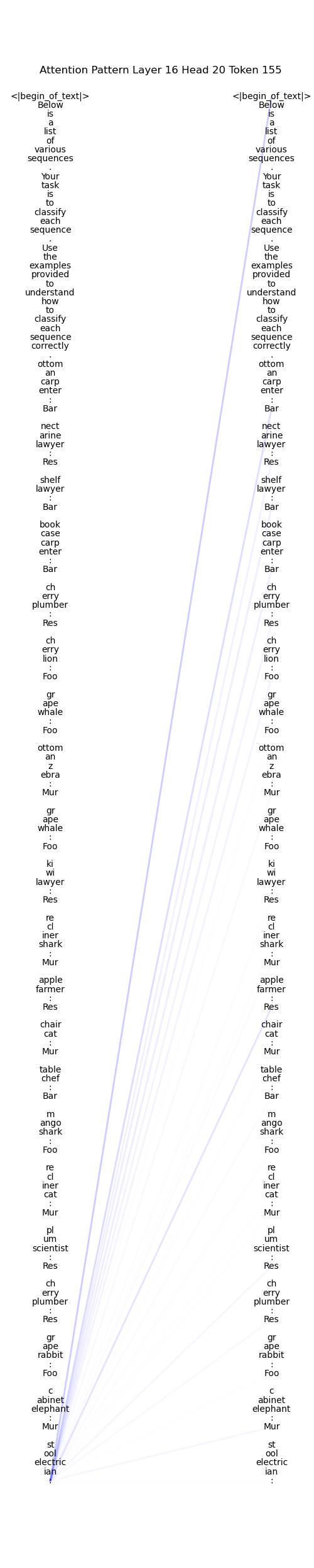}
  \caption{Attention pattern of token 155 in head 20 of layer 16 in Llama-3-8B. Plot 3/5.}
\end{minipage}
\end{figure*}
\newpage

\begin{figure*}[h!]
\centering
\begin{minipage}{0.48\textwidth}
\centering
  \includegraphics[scale=0.35]{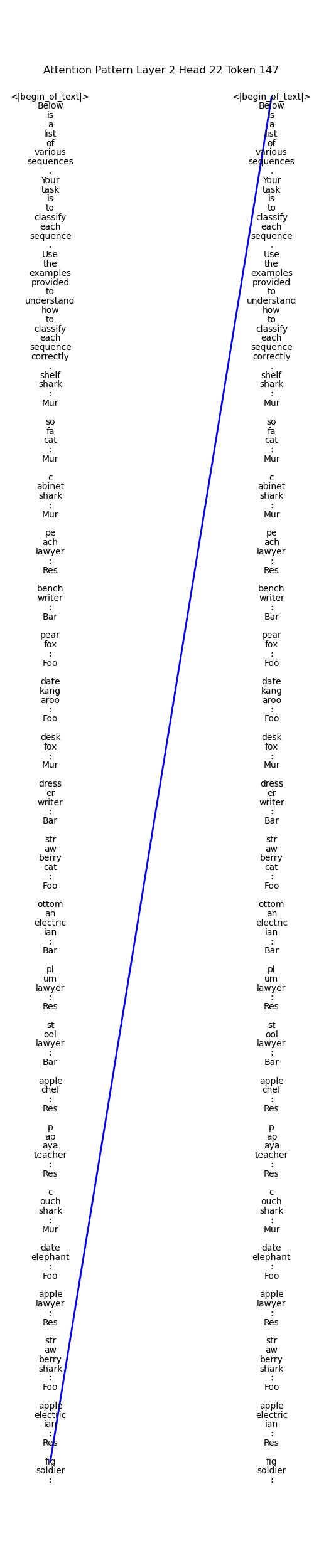}
  \caption{Attention pattern of token 147 in head 22 of layer 2 in Llama-3-8B. Example 4/5.}
\end{minipage}\hfill
\begin{minipage}{0.48\textwidth}
\centering
  \includegraphics[scale=0.35]{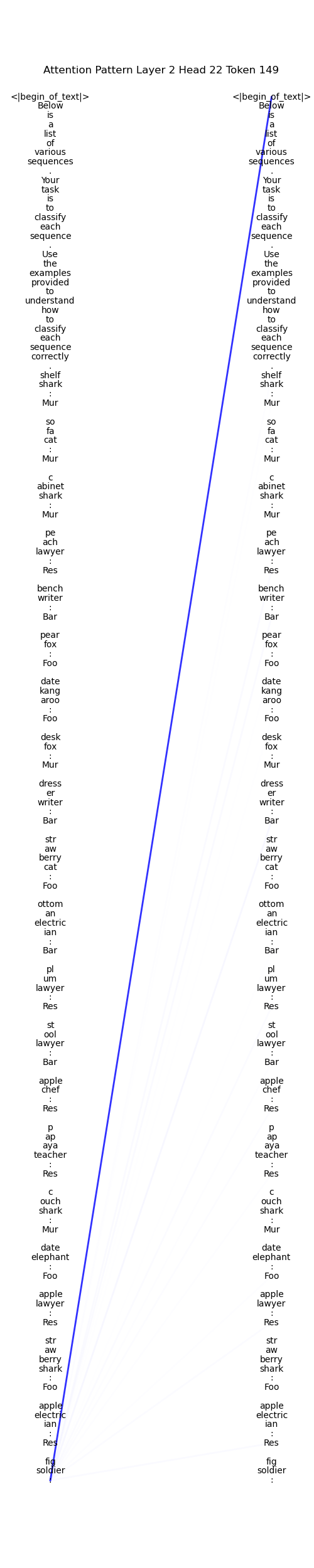}
  \caption{Attention pattern of token 149 in head 22 of layer 2 in Llama-3-8B. Plot 4/5.}
\end{minipage}
\end{figure*}
\newpage

\begin{figure*}[h!]
\centering
\begin{minipage}{0.48\textwidth}
\centering
  \includegraphics[scale=0.35]{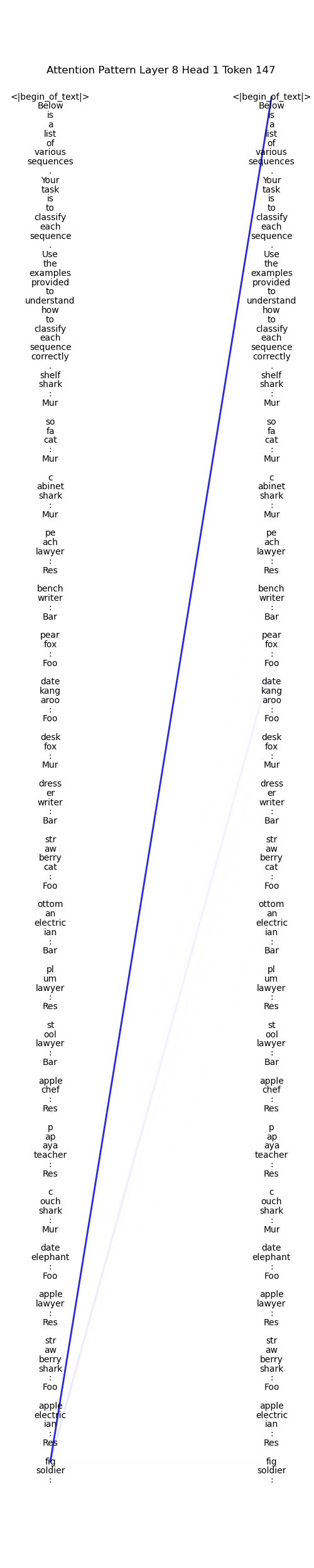}
  \caption{Attention pattern of token 147 in head 1 of layer 8 in Llama-3-8B. Example 4/5.}
\end{minipage}\hfill
\begin{minipage}{0.48\textwidth}
\centering
  \includegraphics[scale=0.35]{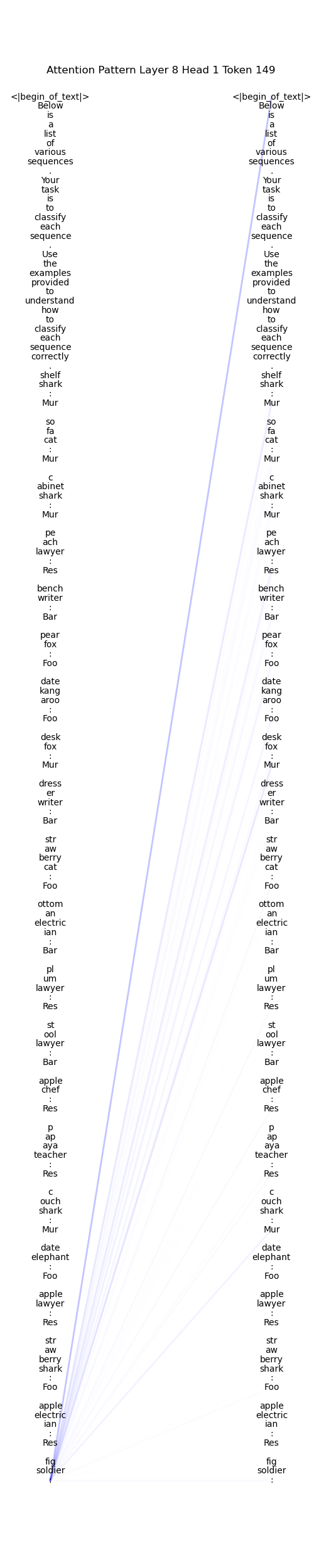}
  \caption{Attention pattern of token 149 in head 1 of layer 8 in Llama-3-8B. Plot 4/5.}
\end{minipage}
\end{figure*}
\newpage

\begin{figure*}[h!]
\centering
\begin{minipage}{0.48\textwidth}
\centering
  \includegraphics[scale=0.35]{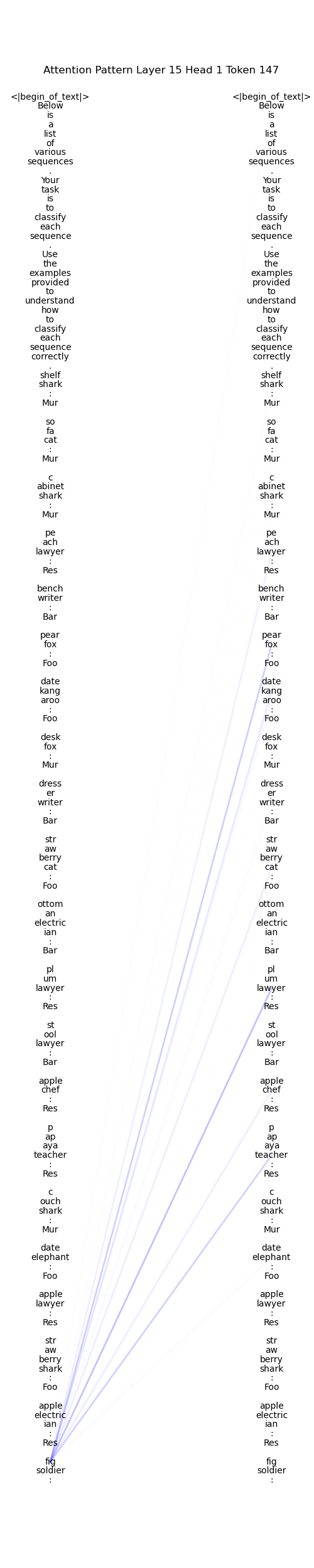}
  \caption{Attention pattern of token 147 in head 1 of layer 15 in Llama-3-8B. Example 4/5.}
\end{minipage}\hfill
\begin{minipage}{0.48\textwidth}
\centering
  \includegraphics[scale=0.35]{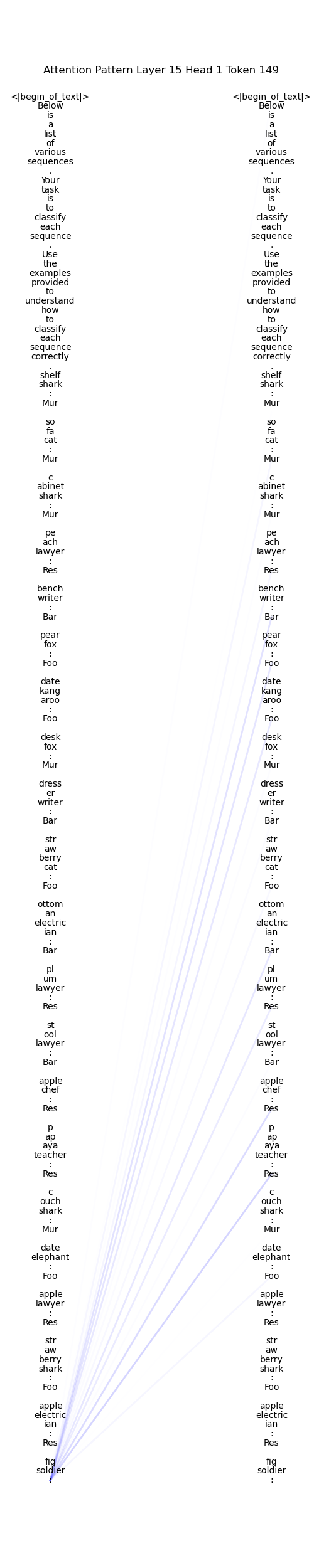}
  \caption{Attention pattern of token 149 in head 1 of layer 15 in Llama-3-8B. Plot 4/5.}
\end{minipage}
\end{figure*}
\newpage

\begin{figure*}[h!]
\centering
\begin{minipage}{0.48\textwidth}
\centering
  \includegraphics[scale=0.35]{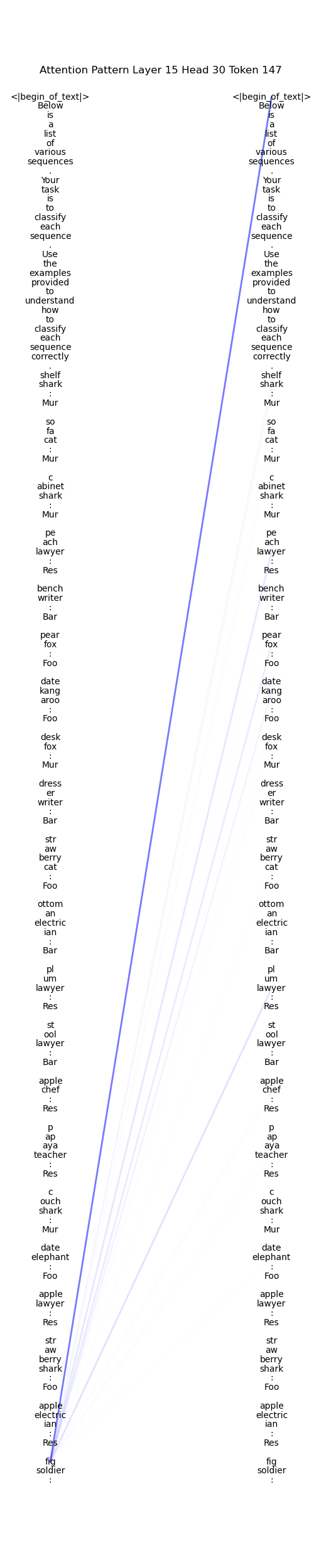}
  \caption{Attention pattern of token 147 in head 30 of layer 15 in Llama-3-8B. Example 4/5.}
\end{minipage}\hfill
\begin{minipage}{0.48\textwidth}
\centering
  \includegraphics[scale=0.35]{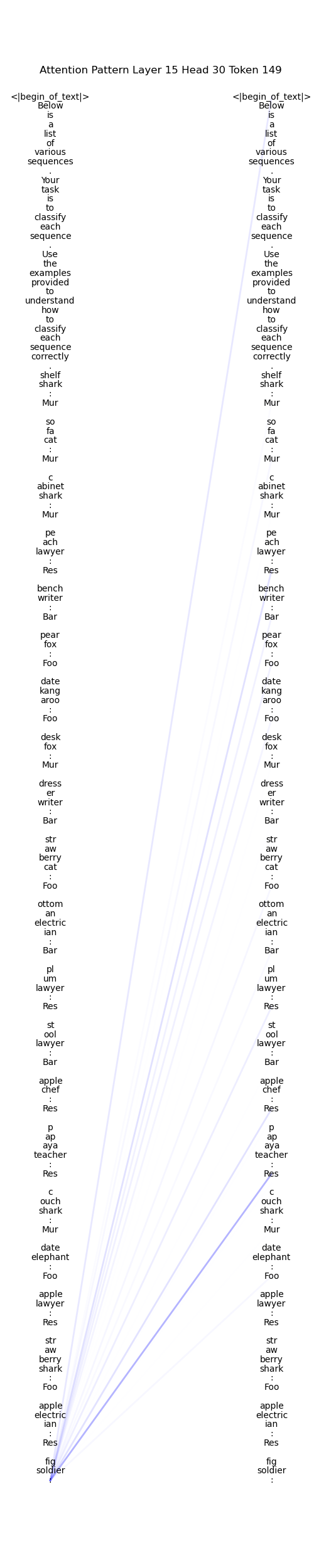}
  \caption{Attention pattern of token 149 in head 30 of layer 15 in Llama-3-8B. Plot 4/5.}
\end{minipage}
\end{figure*}
\newpage

\begin{figure*}[h!]
\centering
\begin{minipage}{0.48\textwidth}
\centering
  \includegraphics[scale=0.35]{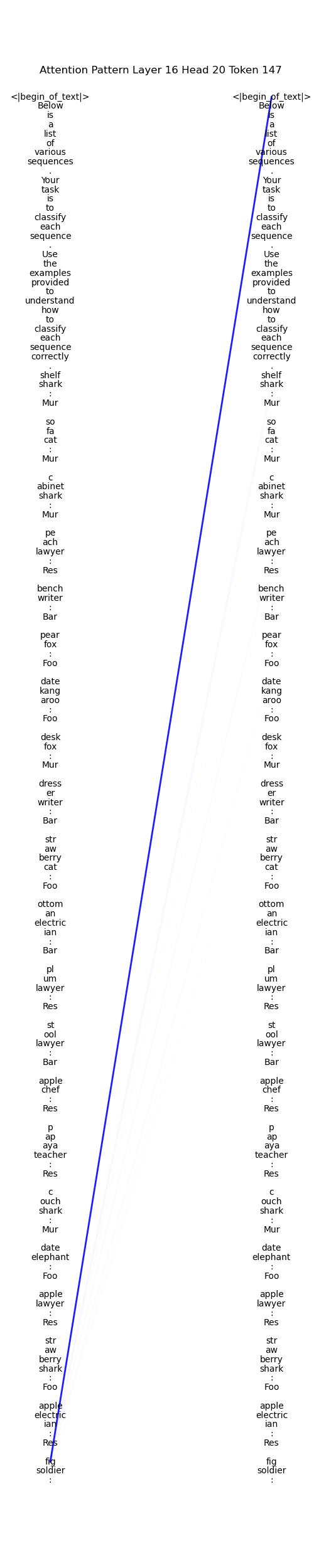}
  \caption{Attention pattern of token 147 in head 20 of layer 16 in Llama-3-8B. Example 4/5.}
\end{minipage}\hfill
\begin{minipage}{0.48\textwidth}
\centering
  \includegraphics[scale=0.35]{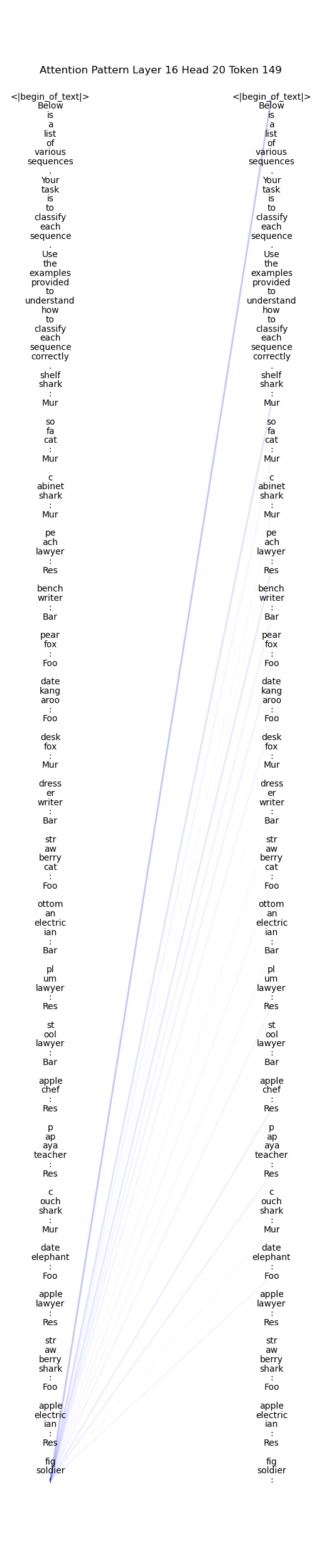}
  \caption{Attention pattern of token 149 in head 20 of layer 16 in Llama-3-8B. Plot 4/5.}
\end{minipage}
\end{figure*}
\newpage

\begin{figure*}[h!]
\centering
\begin{minipage}{0.48\textwidth}
\centering
  \includegraphics[scale=0.35]{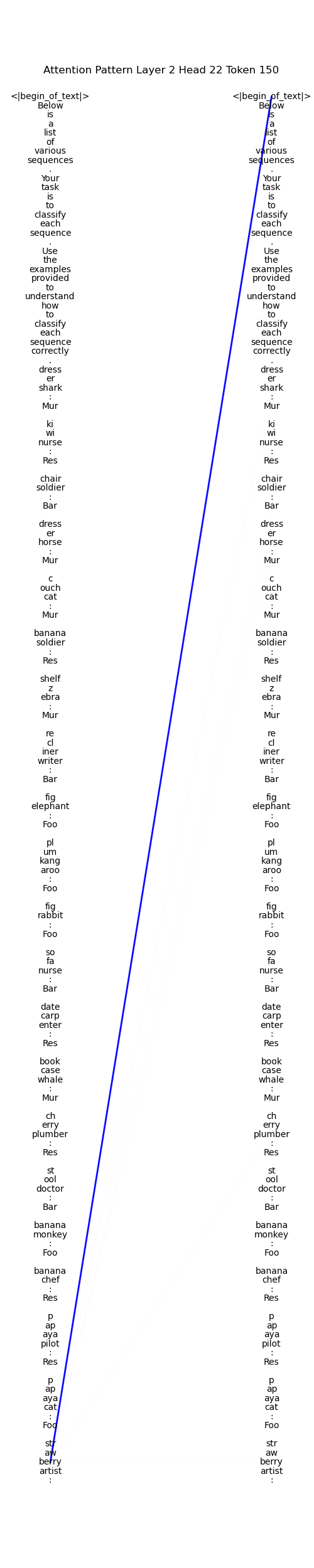}
  \caption{Attention pattern of token 150 in head 22 of layer 2 in Llama-3-8B. Example 5/5.}
\end{minipage}\hfill
\begin{minipage}{0.48\textwidth}
\centering
  \includegraphics[scale=0.35]{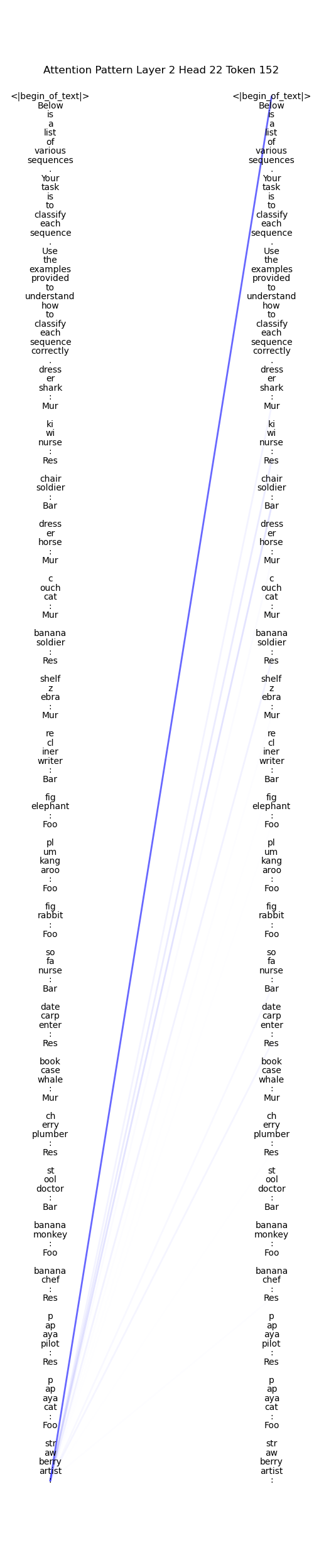}
  \caption{Attention pattern of token 152 in head 22 of layer 2 in Llama-3-8B. Plot 5/5.}
\end{minipage}
\end{figure*}
\newpage

\begin{figure*}[h!]
\centering
\begin{minipage}{0.48\textwidth}
\centering
  \includegraphics[scale=0.35]{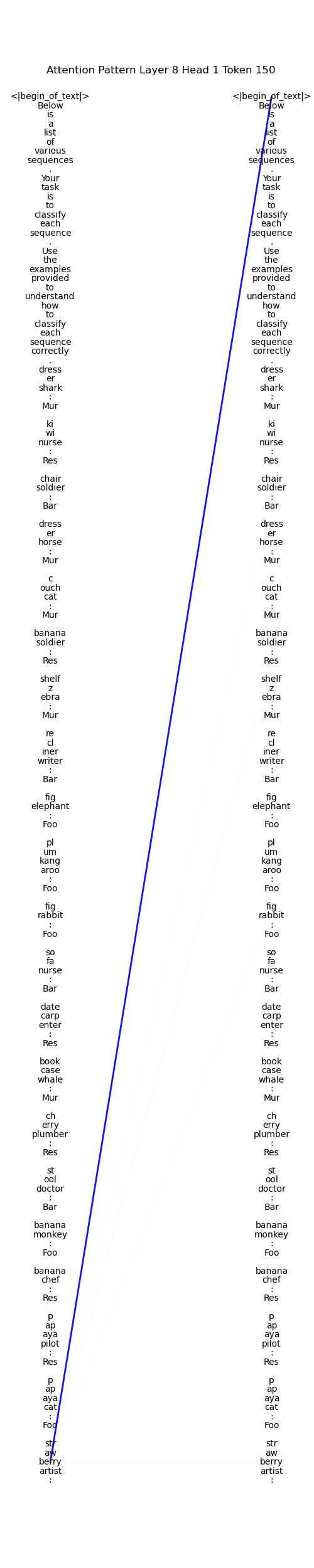}
  \caption{Attention pattern of token 150 in head 1 of layer 8 in Llama-3-8B. Example 5/5.}
\end{minipage}\hfill
\begin{minipage}{0.48\textwidth}
\centering
  \includegraphics[scale=0.35]{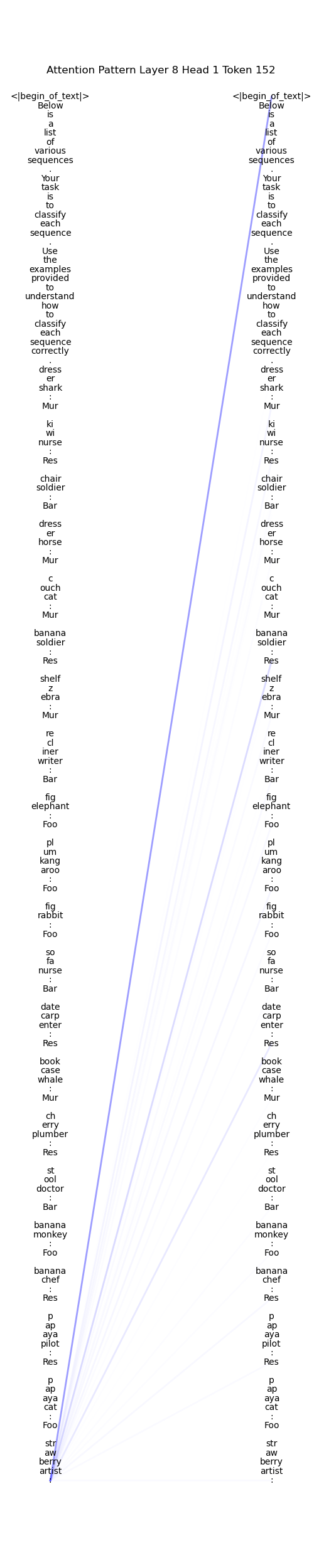}
  \caption{Attention pattern of token 152 in head 1 of layer 8 in Llama-3-8B. Plot 5/5.}
\end{minipage}
\end{figure*}
\newpage

\begin{figure*}[h!]
\centering
\begin{minipage}{0.48\textwidth}
\centering
  \includegraphics[scale=0.35]{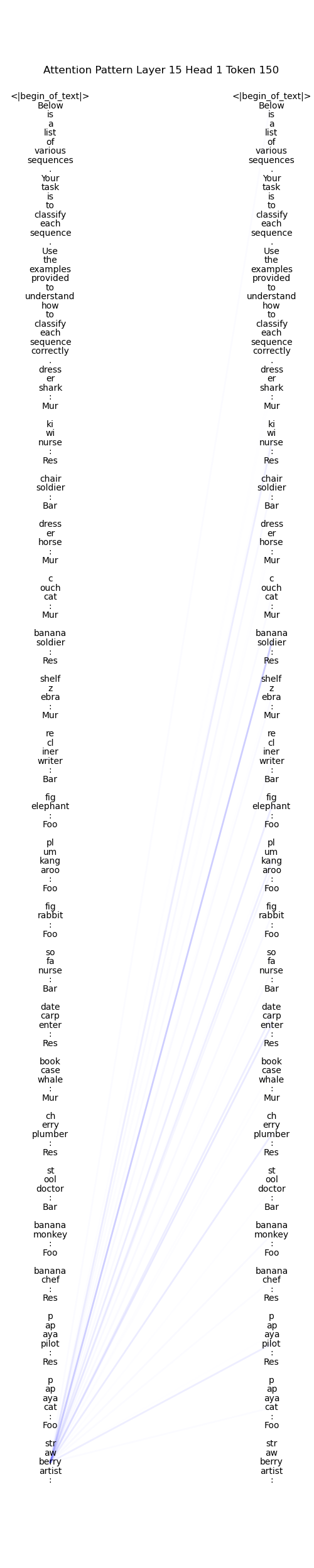}
  \caption{Attention pattern of token 150 in head 1 of layer 15 in Llama-3-8B. Example 5/5.}
\end{minipage}\hfill
\begin{minipage}{0.48\textwidth}
\centering
  \includegraphics[scale=0.35]{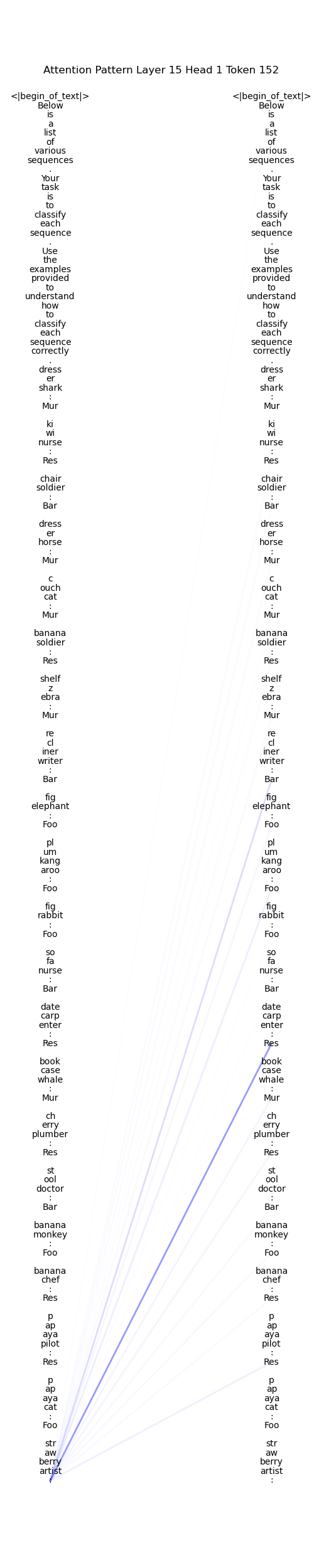}
  \caption{Attention pattern of token 152 in head 1 of layer 15 in Llama-3-8B. Plot 5/5.}
\end{minipage}
\end{figure*}
\newpage

\begin{figure*}[h!]
\centering
\begin{minipage}{0.48\textwidth}
\centering
  \includegraphics[scale=0.35]{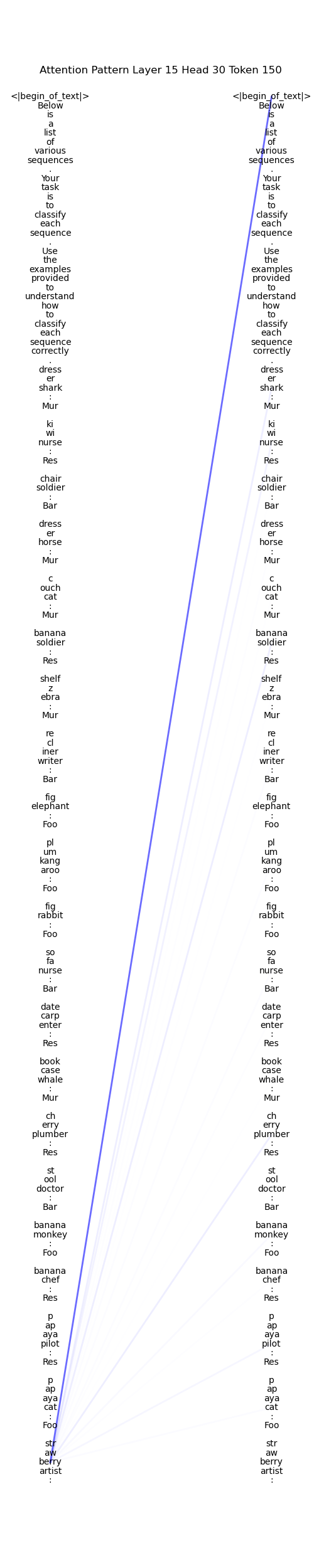}
  \caption{Attention pattern of token 150 in head 30 of layer 15 in Llama-3-8B. Example 5/5.}
\end{minipage}\hfill
\begin{minipage}{0.48\textwidth}
\centering
  \includegraphics[scale=0.35]{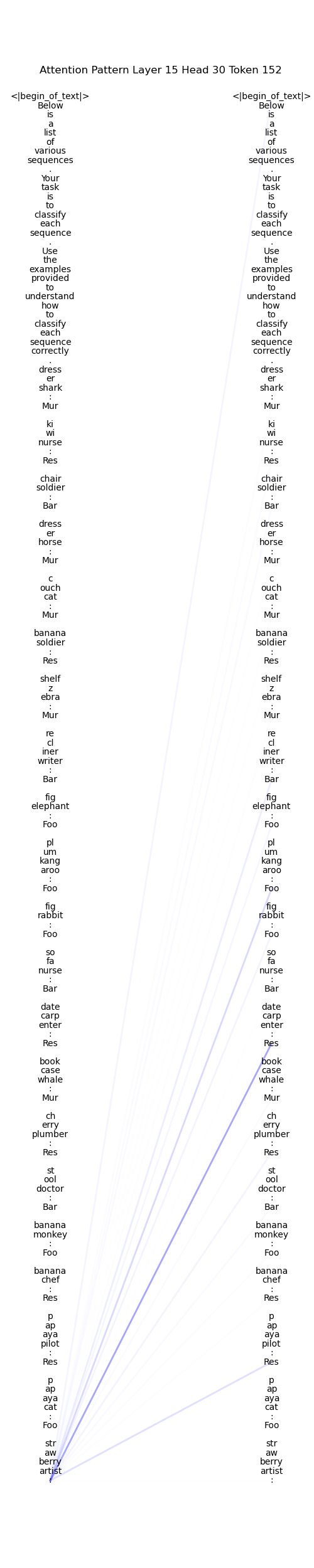}
  \caption{Attention pattern of token 152 in head 30 of layer 15 in Llama-3-8B. Plot 5/5.}
\end{minipage}
\end{figure*}
\newpage

\begin{figure*}[h!]
\centering
\begin{minipage}{0.48\textwidth}
\centering
  \includegraphics[scale=0.35]{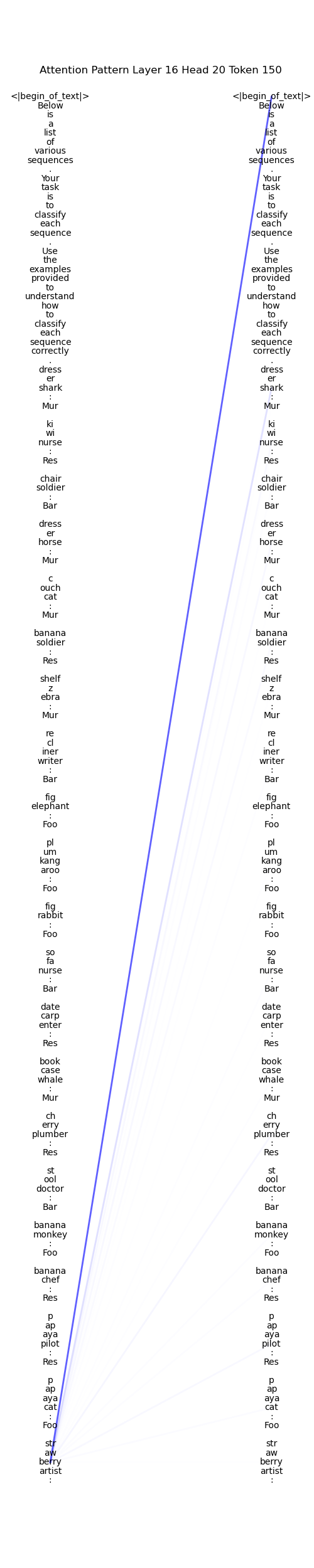}
  \caption{Attention pattern of token 150 in head 20 of layer 16 in Llama-3-8B. Example 5/5.}
\end{minipage}\hfill
\begin{minipage}{0.48\textwidth}
\centering
  \includegraphics[scale=0.35]{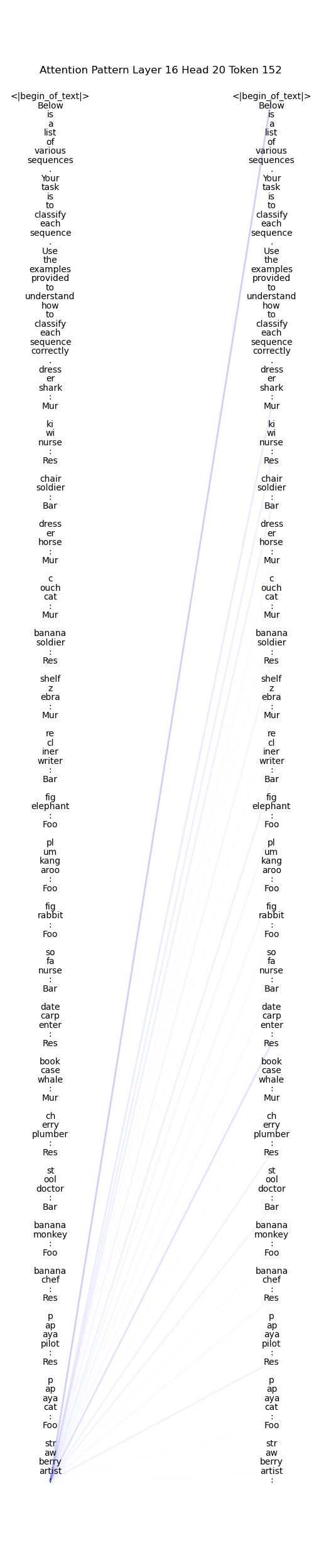}
  \caption{Attention pattern of token 152 in head 20 of layer 16 in Llama-3-8B. Plot 5/5.}
\end{minipage}
\end{figure*}
\newpage

\end{document}